\documentclass[AMA,Times1COL]{WileyNJDv5}
\articletype{Original article}

\usepackage{tikz}
\usepackage{import}
\usetikzlibrary{quotes,arrows.meta}
\usetikzlibrary{positioning}

\def\edgecolor{rgb:blue,4;red,1;green,4;black,3}
\newcommand{\midarrow}{\tikz \draw[-Stealth,line width =0.8mm,draw=\edgecolor] (-0.3,0) -- ++(0.3,0);}

\usepackage{Ball}
\usepackage{Box}
\usepackage{RightBandedBox}

\usepackage[percent]{overpic}
\usepackage{comment}
\usepackage{cleveref}
\usepackage{subcaption}
\usepackage{enumitem}
\usepackage{booktabs}
\usepackage{makecell}

\usetikzlibrary{positioning, calc, decorations.pathreplacing, arrows.meta}

\usetikzlibrary{positioning}
\usetikzlibrary{3d} 
\def\ConvColor{rgb:yellow,5;red,2.5;white,5}
\def\ConvReluColor{rgb:yellow,5;red,5;white,5}
\def\PoolColor{rgb:red,1;black,0.3}
\def\DcnvColor{rgb:blue,5;green,2.5;white,5}
\def\SoftmaxColor{rgb:magenta,5;black,7}
\def\SumColor{rgb:blue,5;green,15}

\newcommand{\sM}{\begin{array}{ccccccccc}}
\newcommand{\eM}{\end{array}}

\newcommand{\pd}[2]{\displaystyle\frac{\displaystyle\partial #1}{\displaystyle\partial #2}}

\newcommand{\z}[1]{{\tilde{#1}}}

\newcommand{\lb}{\left(}
\newcommand{\rb}{\right)}

\newcommand{\la}{\langle}
\newcommand{\ra}{\rangle}

\newcommand{\sv}{\lb\begin{array}{ccccccccccccccccc}}
\newcommand{\sV}{\begin{bmatrix}}
\newcommand{\eV}{\end{bmatrix}}
\newcommand{\ev}{\end{array}\rb}

\newcommand{\fempty}[1]{{}}

\renewcommand{\r}[2]{{#2_{\langle #1 \rangle}}}
\newcommand{\sty}[1]{{\boldsymbol{#1}}}

\newcommand{\styy}[1]{{\mathbb{#1}}}

\newcommand{\fe}{\sty{ e}}

\newcommand{\fx}{\sty{ x}}

\newcommand{\fA}{\sty{ A}}
\newcommand{\fB}{\sty{ B}}

\newcommand{\fI}{\sty{ I}}

\newcommand{\fQ}{\sty{ Q}}

\newcommand{\fU}{\sty{ U}}
\newcommand{\fV}{\sty{ V}}

\newcommand{\fX}{\sty{ X}}
\newcommand{\fY}{\sty{ Y}}
\newcommand{\fZ}{\sty{ Z}}

\newcommand{\fzero}{\sty{ 0}}

\newcommand{\ffC}{\styy{ C}}

\newcommand{\ffI}{\styy{ I}}

\newcommand{\ffP}{\styy{ P}}

\newcommand{\ffR}{\styy{ R}}
\newcommand{\ffS}{\styy{ S}}

\newcommand{\ffY}{\styy{ Y}}

\newcommand{\cA}{{\cal A}}
\newcommand{\cB}{{\cal B}}
\newcommand{\cC}{{\cal C}}

\newcommand{\cE}{{\cal E}}
\newcommand{\cF}{{\cal F}}

\newcommand{\cI}{{\cal I}}
\newcommand{\cJ}{{\cal J}}
\newcommand{\cK}{{\cal K}}
\newcommand{\cL}{{\cal L}}
\newcommand{\cM}{{\cal M}}

\newcommand{\cO}{{\cal O}}
\newcommand{\cP}{{\cal P}}

\newcommand{\cR}{{\cal R}}
\newcommand{\cS}{{\cal S}}
\newcommand{\cT}{{\cal T}}
\newcommand{\cU}{{\cal U}}

\renewcommand{\thefootnote}{\fnsymbol{footnote}}

\newcommand{\WT}[1]{\widetilde{#1}}
\newcommand{\WH}[1]{\widehat{#1}}
\newcommand{\ol}[1]{\overline{#1}}

\renewcommand{\ul}[1]{\underline{#1}} 
\newcommand{\ulWH}[1]{\ul{\widehat{#1}}}
\newcommand{\ulWT}[1]{\ul{\widetilde{#1}}}

\newcommand{\ull}[1]{\ul{\ul{#1}}}
\newcommand{\ullWH}[1]{\ull{\widehat{#1}}}
\newcommand{\ullWT}[1]{\ull{\widetilde{#1}}}
\newcommand{\ullWTH}[1]{\WH{\ullWT{#1}}}
\newcommand{\ulWTH}[1]{\WH{\ulWT{#1}}}

\newcommand\python[1]{\colorbox{lightgray!30}{\texttt{#1}}}

\newcommand{\Voigt}[1]{{#1}_\mathrm{V}}
\newcommand{\Reuss}[1]{{#1}_\mathrm{R}}

\newcommand{\uCvoigt}{\Voigt{\ol{\ull{C}}}}
\newcommand{\uCreuss}{\Reuss{\ol{\ull{C}}}}

\newcommand{\VRNet}{{Voigt-Reuss net}}
\newcommand{\levelset}{\psi}
\newcommand{\levelsetWT}{\WT{\psi}}
\DeclareMathOperator*{\argmin}{arg\,min}

\definecolor{uniSlightblue}{HTML}{00BEFF}
\definecolor{uniSblue}{HTML}{004191}
\definecolor{uniSlblue}{HTML}{00BEFF}
\colorlet{uniSlblue40}{uniSlblue!40!white}
\definecolor{uniSgray}{RGB}{62, 68, 76}
\colorlet{uniSgray20}{uniSgray!20!white}

\begin{document}

\title{Robust inverse material design with physical guarantees\\using the Voigt-Reuss Net}

\author[1]{Sanath Keshav}
\author[1]{Felix Fritzen}

\authormark{KESHAV \& FRITZEN}
\titlemark{Robust inverse material design with physical guarantees using the Voigt-Reuss Net}

\address[1]{\orgdiv{Institute of Applied Mechanics, Data Analytics in Engineering}, \orgname{University of Stuttgart}, \orgaddress{\state{Universitätsstr.32, 70569 Stuttgart, Baden-Württemberg}, \country{Germany}}}

\corres{Corresponding author: Felix Fritzen \email{fritzen@mib.uni-stuttgart.de}}

\presentaddress{Universitätsstr. 32, 70569 Stuttgart, Germany}

\abstract[Abstract]{

We apply the \VRNet{}, a spectrally normalized neural surrogate introduced in~\cite{Keshav2025}, for forward and inverse mechanical homogenization with a key guarantee that all predicted effective stiffness tensors satisfy Voigt-Reuss bounds in the Löwner sense during training, inference, and gradient-driven optimization. The approach operates in a bounded spectral space by reparameterizing each effective tensor relative to its Voigt and Reuss bounds, ensuring that all outputs reside within a unit-cube domain and are mapped back via a deterministic inverse transform to physically admissible tensors.

For 3D elasticity, a fully connected \VRNet{} is trained on $\sim1.18$ million high-fidelity homogenization labels of stochastic biphasic microstructures. The model ingests 236 image-derived morphological descriptors and phase parameters that encode bulk and shear moduli, enabling a single surrogate to represent material combinations spanning 4 orders of magnitude. Due to the rotational invariance of the descriptor set, the surrogate recovers the isotropic projection of the effective stiffness ($R^2\ge 0.998$ for isotropy-related components). However, anisotropy-revealing entries remain unlearnable from the available features.  At the tensor level, the relative Frobenius error exhibits a median of approximately 1.8\% (mean approximately 3.6\%) and approaches the irreducible isotropic-projection floor, far outperforming all alternative surrogates considered.

In 2D plane-strain elasticity, spectral normalization is integrated with a differentiable microstructure renderer and a convolutional regressor, yielding a surrogate that maps generator parameters to effective stiffness tensors for highly anisotropic and high-contrast composites. \VRNet{} is compared against vanilla and Cholesky regressors trained with identical architectures, data, and training procedures. The unconstrained surrogates frequently violate Voigt/Reuss bounds and even positive definiteness, whereas \VRNet{} produces no violations by design and also enables robust inverse design. Utilizing the surrogate with batched first-order optimization, the approach can match prescribed target tensors and optimize tensor functionals, recovering classical optima while avoiding non-physical and spurious designs observed with unconstrained surrogates.
}

\keywords{inverse materials design, structure-property linkage, physics-constrained ML, microstructure homogenization, composite materials, spectral normalization, Voigt-Reuss net}

\jnlcitation{\cname{%
\author{Keshav S}, and
\author{Fritzen F}}.
\ctitle{TODO} \cjournal{\it XXX} \cvol{}.}

\maketitle

\renewcommand\thefootnote{}

\renewcommand\thefootnote{\fnsymbol{footnote}}
\setcounter{footnote}{1}

\section{Introduction}
\subsection{Motivation}
The mechanical response of heterogeneous materials is governed by a subtle interplay between phase properties, volume fractions, and microstructural geometry across multiple scales.
From fiber-reinforced polymers to architected lattices and metamaterials, designers are increasingly seeking to engineer microstructures to tailor effective stiffness, anisotropy, and load transfer pathways for specific applications.
Classical homogenization theory provides a rigorous framework for relating microstructure and local constitutive laws to effective properties, e.g., via asymptotic expansions, variational principles, and empirical estimates and bounds~\cite{Torquato2002, Milton2002, zohdi2005, nemat2013micromechanics}.
Yet, the corresponding \textit{G-closure}~\cite{Cherkaev2000, Allaire2002}, i.e., the set of all effective tensors realizable by mixing given phases in all admissible microgeometries, is typically high-dimensional, non-convex, and analytically intractable.

Computational homogenization and direct numerical simulation have become the workhorses to explore the material design space in practice.
Finite element (FE) and Fast Fourier Transform (FFT)-based schemes~\cite{moulinec1998,Leuschner2017,Schneider2021,combo2022} allow one to solve corrector problems on intricate unit cells and extract homogenized tensors.
These methods are robust and systematically improvable, but they are also computationally expensive:
A single 3D simulation on a $192^3$ voxel grid with strong phase contrast involves $\approx$21 million degrees of freedom. Therefore, it can already be costly, and design tasks that require exploring tens of thousands of candidate microstructures or solving nested optimization problems are often prohibitive if each evaluation relies on high-fidelity simulation. More strikingly, differentiating these results with respect to design parameters is challenging, if not impossible, in many situations.
These bottlenecks have motivated a surge of interest in surrogate models that learn microstructure property relationships from data and can be queried at negligible cost once trained.

\subsection{Machine learned effective property forecasting}

Machine-learning-based surrogates for homogenization are now widely explored.
Approaches range from regression on hand-crafted descriptors (e.g., morphological statistics, correlation measures, or spectral features)~\cite{lissner2019,Prifling2021,lissner2024} to deep convolutional networks ingesting microstructure images or volumes directly~\cite{Aldakheel2023,Nguyen2025-cn,Heider2025-il}.
They have been used to predict scalar effective properties~\cite{Prifling2021}, full elasticity tensors~\cite{EIDEL2023115741}, and even local fields~\cite{Bhattacharya2024,NGUYEN2026106418}.
In parallel, data-driven paradigms in computational mechanics~\cite{Farhat2022,Linden2023,Klein2025} have demonstrated that learned constitutive maps can replace or augment traditional phenomenological models, enabling faster simulations and design. Furthermore, these works have demonstrated the need to incorporate physical constraints into surrogate models to improve data efficiency, robustness, and out-of-sample prediction accuracy.
On the design side, several works now combine surrogates with generative models~\cite{Bastek2023-qt}, optimization~\cite{Rassloff2025, Otto2025-gt,Rosenkranz2025-ej}, learned inverse mappings~\cite{Kumar2020-vq}, or autoencoders~\cite{Zheng2023-gv} to tackle inverse problems, where one seeks microstructures or process parameters that achieve targeted effective responses.

Despite these advances, most existing surrogates for effective tensors suffer from a critical limitation: they treat the components of the effective tensor as generic regression targets, with physical constraints imposed (if at all) through post-processing or soft penalization.
In linear elasticity, the effective stiffness tensor must be symmetric, positive definite, and satisfy fundamental variational bounds, most notably the first-order Voigt and Reuss bounds, which constrain the admissible tensors in the Löwner sense~\cite{nemat2013micromechanics}.
Naive regression does not respect such structural constraints: predicted tensors can be indefinite, violate ellipticity, or lie outside the Voigt-Reuss envelope, especially when queried far from the training distribution.
Such violations are more than just aesthetic flaws: they can destabilize downstream FE simulations, break thermodynamic consistency, and render inverse-design loops unreliable.
In order to preserve symmetry and positive definiteness, learning a Cholesky decomposition~\cite{Xu2021} is a viable solution. Another option is to add penalty terms to the loss in the spirit of physics-informed neural networks (PINNs\cite{Karniadakis2021}) which, however, only yield a weak consideration of the constraints. None of these approaches guarantees adherence to both upper and lower Voigt-Reuss bounds, and often this misleads the optimization process.

At the same time, homogenization theory provides a much richer structure than simple SPD constraints.
For a given set of phase tensors and volume fractions, the effective tensor must lie in the intersection of the SPD cone with a set of variational bounds such as the classical Voigt and Reuss bounds~\cite{Voigt1889,Reuss1929,Wiener1912}. Under some assumptions, the Hashin-Shtrikman bounds~\cite{Hashin1963-am}, and more refined and problem-specific estimates~\cite{Castaneda1992,Talbot2004} can apply, as well.
In Löwner order, the basic Voigt-Reuss inequality states that the effective tensor is bounded above and below by those of the weighted arithmetic and harmonic averages, respectively.
Geometrically, this defines a compact spectral simplex in the space of SPD matrices where the difference between the Voigt and Reuss bounds is positive semi-definite and encodes the admissible variation of effective responses.
Any surrogate used for forward prediction or inverse design \textit{must} remain in the thereby defined feasible set. In particular, any configurations explored along optimization trajectories must reside inside this envelope.

Another source of difficulty is the intrinsic complexity of the G-closure, especially for high-contrast and anisotropic composites.
For strongly heterogeneous materials, small changes in geometry can cause large, non-smooth changes in effective response, particularly near percolation thresholds or when load-bearing paths reconfigure~\cite{Torquato2002,Boddapati2024-tb}.
Effective stiffness eigenvalues can exhibit jumps as microstructure parameters (e.g., volume fraction, connectivity, or orientation) cross critical values.
Mean-field approaches, such as, e.g., the Mori-Tanaka scheme \cite{nemat2013micromechanics} and self-consistent schemes, which implicitly assume dilute or weakly interacting inclusions, cannot capture these transitions.
Standard neural surrogates can approximate them, but when unconstrained, they may also generate unphysical responses, e.g., in poorly sampled regions of the design space.
Robust inverse design thus demands surrogates that are both expressive enough to capture non-smooth, geometry-driven behavior and structured enough to enforce physical admissibility by construction.

In a recent work~\cite{Keshav2025}, we proposed a spectral normalization framework for learning effective conductivities resulting from the solutions of elliptic problems.
The core idea is to exploit the Voigt-Reuss gap as a natural scaling for the admissible response and to reparameterize the effective tensor via a normalized SPD matrix with spectrum confined to the unit interval.
Learning is then carried out in the bounded unit-cube space of eigenvalues and orthogonal factors of the normalized tensor. The reconstruction of the constitutive tensor based on the normalized representation automatically enforces symmetry, positive definiteness, and Voigt-Reuss admissibility.
That prior study focused on scalar-valued elliptic problems such as thermal conduction.

\subsection{Objectives and scope.}
In this paper, we generalize the \VRNet{} framework to linear elasticity and demonstrate how a spectrally normalized surrogate can be used not only for fast, physically admissible forward prediction of effective stiffness tensors, but also as a robust engine for gradient-based inverse material design. Building on our earlier study for scalar elliptic problems~\cite{Keshav2025}, we first formalize a spectral normalization for generic symmetric positive definite constitutive tensors (of arbitrary order), including 3D small-strain elasticity as a special case. Any effective tensor is mapped through a Cholesky-like factor of the Voigt-Reuss difference into a dimensionless matrix with spectrum in $[0,1]$, yielding a task-agnostic target space with scale-free, well-conditioned losses and the guarantee that, after inverse mapping, all surrogate predictions are symmetric positive definite and confined to the Voigt-Reuss envelope.

Within this general framework, we consider two representative elasticity settings. First, we construct a \VRNet{} for 3D biphasic composites based on the open microstructure dataset~\cite{ulm2020_dataset,Prifling2021}, where each sample is described by isotropic averaged morphology descriptors and non-dimensional phase contrast parameters. This case probes the performance of spectral normalization in a purely descriptor-based, rotationally invariant forward problem. Second, we couple spectral normalization with a differentiable microstructure renderer and a convolutional neural network for 2D plane-strain composites generated from thresholded trigonometric fields~\cite{Boddapati2023-la,Boddapati2024-tb}. In this setting, the surrogate operates directly on parametric microstructure representations, enabling end-to-end differentiability of the map from design parameters to effective stiffness and, consequently, first-order inverse design with respect to full target tensors or tensor-valued functionals. Throughout, we restrict attention to linear elastostatics, but emphasize that the proposed spectral normalization is generic and applicable to other elliptic operators (e.g., thermal conductivity, diffusion, poroelasticity) and further.

The remainder of the paper is organized as follows. In \Cref{sec:spectral:bounds,sec:spectral:normalization} we recall the Voigt-Reuss bounds for elliptic problems and derive the associated spectral normalization, including its matrix formulation and numerical implementation. \Cref{sec:implementation_spectral_bounds} discusses how this normalization can be combined with generic machine-learning models. \Cref{sec:3d_mechanics} presents the 3D mechanical homogenization study based on descriptor inputs, while \Cref{sec:2d_mechanics} introduces the differentiable microstructure renderer, the convolutional \VRNet{}, and a series of forward and inverse design experiments for highly anisotropic, high-contrast 2D composites. The conclusions summarize the main findings and outline opportunities for extending \VRNet{} to richer G-closure problems and multi-physics design.

\section{Notation}
\label{sec:notation}

The spatial average of a quantity over a domain $\cA$ with measure $A = \lvert \cA \rvert$ is defined as 
\begin{equation}
	\langle\cdot\rangle_{\cA}=\dfrac{1}{A} \int_{\cA}(\cdot) \; \mathrm{d} A \; .
\end{equation}
In the sequel, boldface lowercase letters denote vectors, boldface uppercase letters denote 2-tensors, and blackboard bold uppercase letters (e.g., $\ffC$) denote 4-tensors. We use single ($\cdot$) and double contractions $(:)$. With respect to an orthonormal basis $\fe_i$ ($i\in \{1, 2, 3\}$), they are defined as
($\fA$: order $n$ tensor; $\fB$: order $m$ tensor)
\begin{align}
    \fA &= A_{i_1 i_2\cdots i_{n-1}i_n} \fe_{i_1} \otimes \cdots \otimes \fe_{i_n}, \qquad
    \fB = B_{j_1 j_2\cdots j_{m-1}j_m} \fe_{j_1} \otimes \cdots \otimes \fe_{j_m}, \\
    \fA \cdot \fB &= A_{i_1 i_2 \cdots i_{n-1} k}B_{k j_2 \cdots j_m}
    \fe_{i_1} \otimes \cdots \fe_{i_{n-1}} \otimes \fe_{j_2} \otimes \cdots \fe_{j_m} \; , \\
    \fA : \fB &= A_{i_1 i_2 \cdots i_{n-2} kl}B_{kl j_3 \cdots j_m}
    \fe_{i_1} \otimes \cdots \fe_{i_{n-2}} \otimes \fe_{j_3} \otimes \cdots \fe_{j_m} .
\end{align}
using Einstein summation over repeated indices. The tensor product is denoted by $\otimes$. Using the above notation in conjunction with the operator
\begin{align}
    \nabla &= \pd{}{x_i} \fe_i \; , 
\end{align}
The (right) divergence and the (right) gradient of a general tensor read
\begin{align}
    \fA \cdot \nabla &= A_{i_1 \cdots i_{n-1} j, j} \fe_{i_1} \otimes \cdots \otimes \fe_{i_{n-1}}, &
    \fA \otimes \nabla &= A_{i_1 \cdots i_{n}, j} \fe_{i_1} \otimes \cdots \otimes \fe_{i_n} \otimes \fe_j \; .
\end{align}
Relative Frobenius error is given by,
\begin{align}
  \mathcal{E}_\mathsf{F}(\ull{A}, \ull{B}) &= \frac{\|\ull{A} - \ull{B}\|_\mathsf{F}}{\|\ull{A}\|_\mathsf{F}} = \frac{\sqrt{\sum^{m,n}_{i,j} (A_{ij} - B_{ij})^2}}{\sqrt{\sum^{m,n}_{i,j} A_{ij}^2}}, \qquad \text{where } \ull{A}, \ull{B} \in \mathbb{R}^{m \times n}.
\end{align}

Generally, $\WH{(\cdot)}$ denotes predicted quantities, $\WT{(\cdot)}$ denotes normalized quantities, and $\ol{(\cdot)}$ denotes homogenized quantities.

For convenience, a six-dimensional orthonormal basis $\{\fB_1, \dots, \fB_6\}$ of the space of symmetric second-order tensors~$Sym(\ffR^3)$ in a Mandel-like notation is proposed:
\begin{align}
\left.
\begin{array}{rlrlrl}
    \fB_1 &= \fe_1 \otimes \fe_1, \ &
    \fB_2 &= \fe_2 \otimes \fe_2, \ &
    \fB_3 &= \fe_3 \otimes \fe_3, \ \\
    \fB_4 &= \frac{\sqrt{2}}{2} \lb \fe_1 \otimes \fe_2 + \fe_2 \otimes \fe_1 \rb, \ &
    \fB_5 &= \frac{\sqrt{2}}{2} \lb \fe_1 \otimes \fe_3 + \fe_3 \otimes \fe_1 \rb, \ &
    \fB_6 &= \frac{\sqrt{2}}{2} \lb \fe_2 \otimes \fe_3 + \fe_3 \otimes \fe_2 \rb\; . \end{array}\right. \label{eq:ONB:strain}
\end{align}
Then any symmetric second-order tensor $\fU\in Sym(\ffR^3)$ and any fourth-order tensor~$\ffY$ with left and right sub-symmetry can be expressed by a vector $\ul{u} \in \ffR^6$ and a matrix $\ull{Y}\in \ffR^{6\times 6}$ via\footnote{In two dimensions vectors are of size $3$ and matrices are $3 \times 3$, with basis tensor $\WT{\fB}_1=\fB_1, \WT{\fB}_2=\fB_2, \WT{\fB}_3=\fB_4$.}
\begin{align}
    u_i &= \fU : \fB_i, &
    Y_{ij} &= \fB_i : \ffY : \fB_j \; .
\end{align}
The following identities hold for symmetric tensors of order 2 ($\fU, \fV$) and 4 ($\ffC, \ffS$) and their respective vector representations $\ul{u}, \ul{v}, \ull{C}, \ull{S}$:
\begin{align}
    \fU \cdot \fV & \leftrightarrow \ul{u}^\mathsf{T} \ul{v} = \ul{u} \cdot \ul{v} = \ul{v}^\mathsf{T} \ul{u}, &
    \ffC : \fU & \leftrightarrow \ull{C} \; \ul{u}, &
    \ffC^{-1} : \fV & \leftrightarrow \ull{C}^{-1} \; \ul{v} \; .
\end{align}

\section{Foundations}

\subsection{Löwner bounds for elliptic problems}
\label{sec:spectral:bounds}

We build on our recent work~\cite{Keshav2025} and extend the framework to 3D mechanical problems.
We consider spectral normalization of a symmetric positive definite (SPD) constitutive tensor $\ffY$ (of arbitrary order) that maps an input $\fX$ and a general tensor $\fZ$, i.e.
\begin{align}
\fZ \cdot \ffY \cdot \fX &= \fX \cdot \ffY \cdot \fZ \, , &
\fX \cdot \ffY \cdot \fX & > 0 \, . \label{eq:prerequisite}
\end{align}
Let $\fX$ be a kinematic quantity (e.g., strain or temperature gradient). Many stationary physical problems can be written as the elliptic equation
\begin{align}
\lb \ffY \cdot \fX \rb \cdot \nabla &= \fzero \, . \label{eq:pde}
\end{align}
This covers, for instance, elasticity, steady-state heat conduction, and diffusion. In a homogenization setting, we set the boundary condition $\la \fX \ra = \fe_i$, where $\fe_i$ denotes the $i$-th unit tensor and $i$ runs over all admissible loadings (e.g., $i\in{1,2,3}$ for thermal problems or $i\in{1,\dots,6}$ for small-strain elasticity). The corresponding solutions are denoted by $\fX_i$. With the effective fluxes $\ol{\fQ}_i=\la \ffY\cdot \fX_i\ra$, the effective constitutive tensor $\ol{\ffY}$, with respect to an orthonormal basis $\fe_1,\dots,\fe_m$\footnote{For instance, the basis $\fB_1, \dots, \fB_6$ from \cref{eq:ONB:strain} will be used for $\fe_1, \dots, \fe_6$ for elastic homogenization.}, admits the matrix representation
\begin{align}
\ol{\ffY} & \leftrightarrow \ol{\ull{Y}} \; , \qquad
\lb \ol{\ull{Y}} \rb_{ij} = \ol{\fQ}_i \cdot \fe_j \quad (i,j= 1, \dots, m).
\label{eq:Ybar:matrix}
\end{align}
Using standard variational principles for the primal and dual problems (see, e.g.,~\cite{nemat2013micromechanics,zohdi2005}), the Voigt and Reuss bounds for $\ol{\ffY}$ are given by the weighted arithmetic and harmonic means of $\ffY$, respectively,
\begin{align}
\Voigt{\ol{\ffY}} &= \la \ffY \ra \; , &
\Reuss{\ol{\ffY}} &= \lb \la \ffY^{-1} \ra \rb^{-1} \; . \label{eq:voigt:reuss}
\end{align}
The ordering is understood in the Löwner sense, i.e., for any $\ol{\fX}$:
\begin{align}
\Reuss{\ol{\ffY}} \preceq \ol{\ffY} \preceq \Voigt{\ol{\ffY}}
\quad \Leftrightarrow \quad
\ol{\fX} \cdot \Reuss{\ol{\ffY}} \cdot \ol{\fX} \leq
\ol{\fX} \cdot \ol{\ffY} \cdot \ol{\fX} \leq
\ol{\fX} \cdot \Voigt{\ol{\ffY}} \cdot \ol{\fX} \; .
\label{eq:loewener}
\end{align}
In the $m\times m$ matrix representation (e.g., $m=3$ for 3D thermal problems and $m=6$ for 3D small-strain elasticity), this reads
\begin{align}
\Reuss{\ol{\ull{Y}}} \preceq \ol{\ull{Y}} \preceq \Voigt{\ol{\ull{Y}}}
\quad \Leftrightarrow \quad
\ol{\ul{x}} \cdot \Reuss{\ol{\ull{Y}}} \cdot \ol{\ul{x}} \leq
\ol{\ul{x}} \cdot \ol{\ull{Y}} \cdot \ol{\ul{X}} \leq
\ol{\ul{x}} \cdot \Voigt{\ol{\ull{Y}}} \cdot \ol{\ul{x}}
\;\; \forall \, \ol{\ul{x}}\in\ffR^m .
\label{eq:loewener:matrix}
\end{align}
In the sequel, we use the matrix form for convenience.

\subsection{Spectral normalization}
\label{sec:spectral:normalization}

The difference between the upper and lower bounds is symmetric and positive semi-definite and thus admits a diagonalization
\begin{align}
\ull{Q}_0 \; \ull{\Lambda}_0 \; \ull{Q}_0^\mathsf{T}
&= \Voigt{\ull{\ol{Y}}} - \Reuss{\ull{\ol{Y}}} \; .
\end{align}
In many situations, $\ull{\Lambda}_0$ is positive definite. If not, truncating to eigenvalues greater than a numerical tolerance $\epsilon>0$\footnote{Note that $\epsilon$ is a purely numerical parameter; in exact algebra, one may set $\epsilon=0$.} yields
\begin{align}
\ullWT{Q}_0 \; \ullWT{\Lambda}_0 \; \ullWT{Q}_0^\mathsf{T}
= \Voigt{\ull{\ol{Y}}} - \Reuss{\ull{\ol{Y}}} + \cO(\epsilon)
= \ull{L} \; \ull{L}^\mathsf{T} \; .
\label{eq:bound:diff}
\end{align}
From that, we can define a matrix $\ull{L}$ (closely related to Cholesky factorization) and its pseudo-inverse\footnote{PseudoInverse refers to the Moore-Penrose pseudo inverse \cite{Penrose1955}.} ~$\ull{L}^+$ via
\begin{align}
    \ull{L} &= \ullWT{Q}_0 \sqrt{\ullWT{\Lambda}_0}, &
    \ull{L}^+ &=  \sqrt{\ullWT{\Lambda}^{-1}_0} \ullWT{Q}^\mathsf{T}_0 \; .
\end{align}
Here $\WT{\bullet}$ is used to denote the column truncation of $\ull{Q}_0$ and the column- and row-truncation of $\ull{\Lambda}_0$ to strictly positive eigenpairs padded by zeros to maintain the dimension $m \times m$ for practical reasons. Note that $\ull{L}$ and $\ull{L}^+$ depend only on the trivial Voigt and Reuss bounds, respectively; i.e., they are independent of further microstructural features.
Symmetric application of the pseudo-inverse to \cref{eq:loewener:matrix} yields
\begin{align}
    0 \leq \colorbox{lightgray!45}{$\ullWT{Y}$} = \ull{L}^+ \lb \Voigt{\ull{\ol{Y}}} - \ull{Y} \rb {\ull{L}^+}^\mathsf{T} \leq \ull{L}^+ \lb \Voigt{\ull{\ol{Y}}} - \Reuss{\ull{\ol{Y}}} \rb {\ull{L}^+}^\mathsf{T} = \ull{I} \; . 
\end{align}
The dedimensionalized linear operator \colorbox{lightgray!45}{$\ullWT{Y}$} can be expressed through an orthogonal matrix $\ullWT{Q}\in Orth(\ffR^m)$ and a diagonal matrix $\ullWT{\Lambda} = \text{diag} ( {\xi_{\lambda}}_1, \dots , {\xi_{\lambda}}_m)$ with ${\ul{\xi}_{\lambda}} \in [0, 1]^m$ via
\begin{align}
    \colorbox{lightgray!45}{$\ullWT{Y}$} &= \ullWT{Q} \; \ullWT{\Lambda} \; \ullWT{Q}^\mathsf{T} \, . \label{eq:Ytilde}
\end{align}
Thereby, the effective constitutive output $\ol{\ull{Y}}$ is expressed as a function of the first-order bounds and the normalized matrix $\ullWT{Y}$ through
\begin{align}
    \ull{\ol{Y}} &= \Voigt{\ull{\ol{Y}}} - \ull{L} \; \lb \ullWT{Q} \; \ullWT{\Lambda} \; \ullWT{Q}^\mathsf{T} \rb \ull{L}^\mathsf{T} = \Voigt{\ull{\ol{Y}}} - \ull{L} \; \colorbox{lightgray!45}{$\ullWT{Y}$} \; \ull{L}^\mathsf{T}.
\end{align}
As a result of the spectral normalization, it suffices to predict \colorbox{lightgray!45}{$\ullWT{Y}$}, being a symmetric positive semi-definite matrix devoid of physical dimension and with a spectrum confined to the unit interval, e.g., by forecasting $(\ullWT{Q}, \ullWT{\Lambda})$.

\begin{remark}
	For thermal problems, the conductivities in isotropic multi-phasic materials are always different, yielding strictly positive eigenvalues in $\ull{\Lambda}_0$. In mechanical problems, the situation can become technically more intricate: Considering a biphasic isotropic elastic medium, the material parameters of the phases can be expressed through the bulk modulus $K_i$ and the shear modulus $G_i$ of phase $i\in\{0, 1\}$. In case either of the two matches, the difference \cref{eq:bound:diff} will contain 0-eigenvalues. For instance, a matching bulk modulus $K_0=K_1$ induces
	\begin{align}
		\Voigt{\ol{\ffC}}-\Reuss{\ol{\ffC}} &= 2 \ol{\Delta G} \; \ffP_2^\mathrm{iso}, &
		\ol{\Delta G} &= \Voigt{G} - \Reuss{G}, &  \ffP_2^\mathrm{iso} &  = \ffI^\mathrm{sym} - \frac{\fI \otimes \fI}{3} \; .
	\end{align}
	The matrices $\ull{L}$ and $\ull{L}^+$ then get
	\begin{align}
		\ull{L} &= \sqrt{2 \ol{\Delta G}} \ull{P}_2^\mathrm{iso}, &
		\ull{L}^+ &= \frac{1}{\sqrt{2 \ol{\Delta G}}} \ull{P}_2^\mathrm{iso}, &
		\ull{P}_2^\mathrm{iso}&= \sV
		2/3 & -1/3 & -1/3 & 0 & 0 &0 \\
		-1/3 & 2/3 & -1/3 & 0 & 0 &0 \\
		-1/3 & -1/3 & 2/3 & 0 & 0 &0 \\
		0 &0 & 0 & 1 & 0 & 0 \\
		0 &0 & 0 & 0 & 1 & 0 \\
		0 &0 & 0 & 0 & 0 & 1 \\
		\eV \; .
		\end{align}
		While theoretically straightforward, the robust computation of $\ull{L}$ and $\ull{L}^+$ must be carefully implemented and tested for such limit cases. Note that the restriction to isotropy in the above is for illustrative purposes only, i.e., the code should be able to cope with general anisotropy. For technical reasons, a numerical threshold $\epsilon>0$ should be used to filter "near 0" eigenvalues in $\ull{\Lambda}_0$.
\end{remark}

\subsection{Implementation of the spectral normalization}
\label{sec:implementation_spectral_bounds}

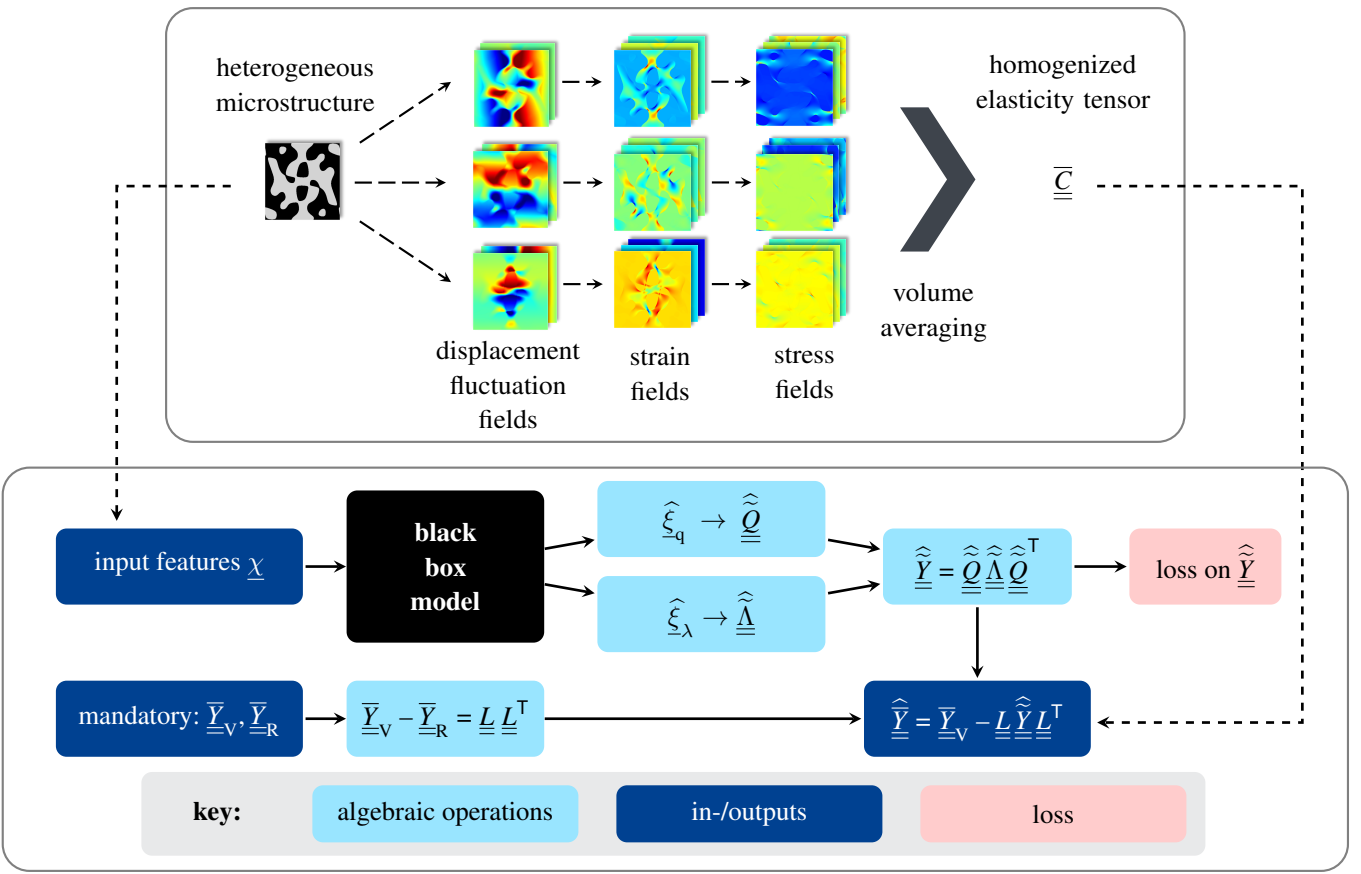
\begin{figure}[!htb]
  \centering
  \begin{tikzpicture}
    \draw[thick, rounded corners=10pt, gray] (-12.2,-0.5) rectangle (1.2,5.2);
    \node[anchor=south west, inner sep=0] (dnsimg) at (-11,0) {\includegraphics[page=6,scale=0.4,trim={0cm 0.9cm 1cm 2.8cm},clip]{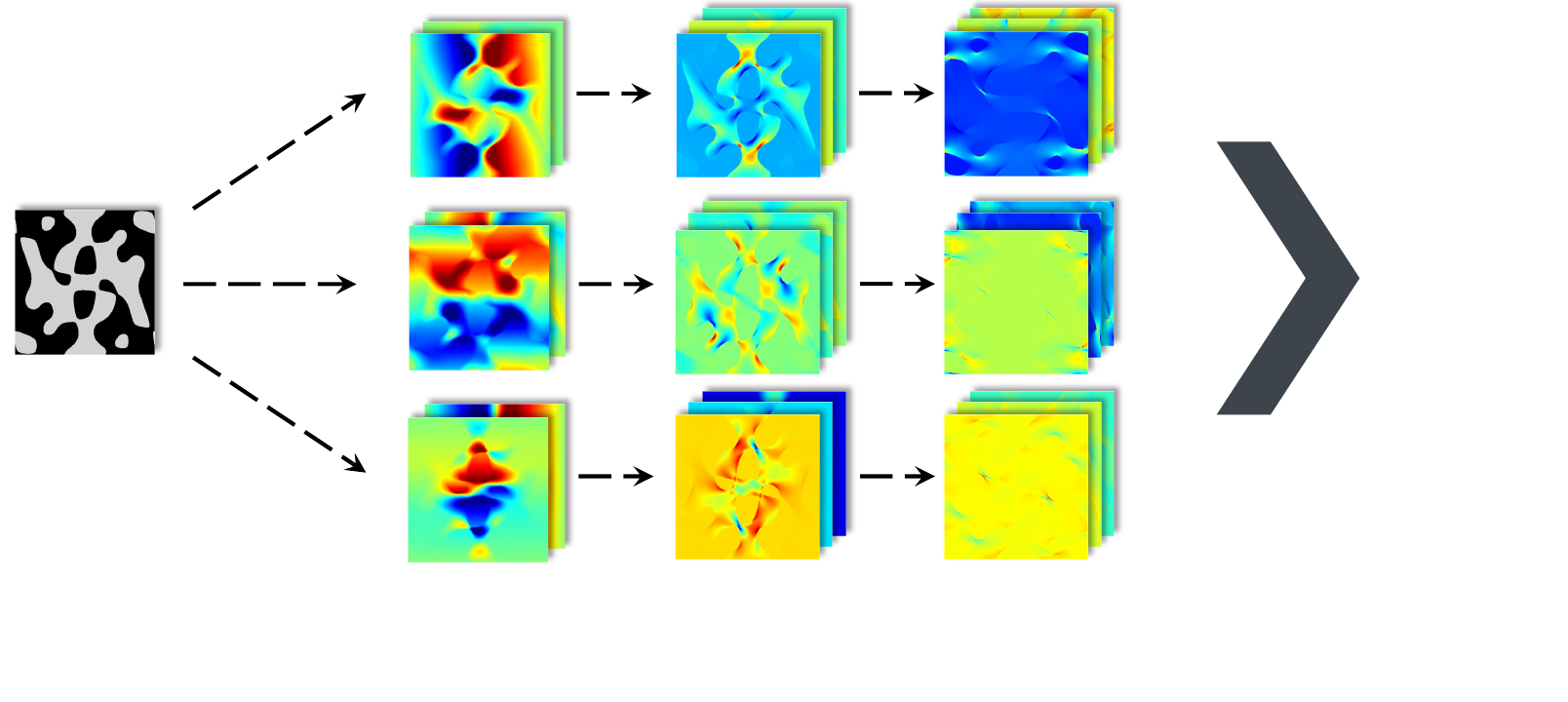}};
    \node[fill=white, text width=2.5cm, align=center] at (-10.5, 4.2) {heterogeneous microstructure};
    \node[fill=white, text width=2.0cm, align=center] at (-7.7, 0.25) {displacement\\ fluctuation fields};
    \node[fill=white, text width=1.0cm, align=center] at (-5.7, 0.4) {strain \\fields};
    \node[fill=white, text width=1.0cm, align=center] at (-3.8, 0.4) {stress \\fields};
    \node[fill=white, text width=1.5cm, align=center] at (-2.1, 1.2) {volume averaging};
    \node[fill=white, text width=2.65cm, align=center] at (-0.4, 4.2) {homogenized\\ elasticity tensor};
    \node[fill=white, text width=0.65cm, align=center] at (-0.4, 2.9) {$\ol{\ull{C}}$};
  \end{tikzpicture}
  \vspace{0.5cm}

  \begin{tikzpicture}[shorten >= 0pt, shorten <= 0pt, rounded corners=4pt]
	\begin{scope}[rectangle, minimum width=2cm, minimum height=1cm, fill=uniSlightblue]
		\node[rectangle, inner sep=0pt, minimum width=2cm, minimum height=1cm, fill=uniSblue, text=white, draw=none, line width=2pt] (input) at (-1.5, 0.5){
			\begin{minipage}{3.25cm}\centering input features $\ul{\chi}$\end{minipage}};
		\node[rectangle, inner sep=0pt, minimum width=2cm, minimum height=1cm, fill=uniSblue, text=white, draw=none, line width=2pt] (trivialinput) at (-1.5, -1.5){
			\begin{minipage}{3.25cm}\centering mandatory: $\ol{\ull{Y }}_\mathrm{V}, \ol{\ull{Y }}_\mathrm{R}$\end{minipage}};
		
		\node[fill=black, text=white, draw=none, inner sep=0pt, minimum height=2cm] (blackbox) at (2,0.5){
			\begin{minipage}{2.6cm}\bfseries\centering black\\box\\model\end{minipage}};
		
		\node[fill=uniSlblue40, draw=none, inner sep=0pt, minimum height=1cm] (Lbox) at (2,-1.5){
			\begin{minipage}{2.6cm}\centering $\ol{\ull{Y }}_\mathrm{V} - \ol{\ull{Y }}_\mathrm{R} = \ull{L}\;\ull{L}^{\mathsf{T}}$\end{minipage}};
		
		\node[rectangle, minimum width=3cm, inner sep=0pt, minimum height=1cm, fill=uniSlblue40, draw=none, line width=2pt] (out1) at (5.5, 1.125){
			\begin{minipage}{2.95cm}\centering $\ulWH{\xi}_\mathrm{q} \ \rightarrow\ \WH{\ullWT{Q}}$\end{minipage}};
		\node[rectangle, minimum width=3cm, inner sep=0pt, minimum height=1cm, fill=uniSlblue40, draw=none, line width=2pt] (out2) at (5.5, -0.125){
			\begin{minipage}{2.95cm}\centering $\ulWH{\xi}_\lambda \rightarrow \WH{\ullWT{\Lambda}}$\end{minipage}};
		
		\node[rectangle, minimum width=2cm, minimum height=1cm, fill=uniSlblue40, draw=none, line width=2pt] (out3) at (9., 0.5){
			\begin{minipage}{2.25cm}\centering $\WH{\ullWT{Y }} = \WH{\ullWT{Q}}\,\WH{\ullWT{\Lambda}} \, \WH{\ullWT{Q}}^\mathsf{T}$\end{minipage}};
		
		\node[rectangle, minimum width=2cm, inner sep=0pt, minimum height=1cm, fill=red!20, draw=none, line width=2pt] (loss) at (12., 0.5){
			\begin{minipage}{2.cm}\centering loss on $\WH{\ullWT{Y }}$\end{minipage}};
		
		\node[rectangle, minimum width=2cm, minimum height=1cm, text=white, fill=uniSblue, draw=none, line width=2pt] (out4) at (9,-1.5){
			\begin{minipage}{2.75cm}\centering $\WH{\ull{\ol{Y }}} = \ull{\ol{Y }}_\mathrm{V} - \ull{L} \, \WH{\ullWT{Y }} \, \ull{L}^\mathsf{T}$\end{minipage}};
		
		\begin{scope}[minimum height=0.75cm, minimum width=3.5cm,inner sep=0pt,yshift=-2.75cm, xshift=2cm]
			\node[draw=none,fill=uniSgray20!60, minimum height=1.1cm, minimum width=14cm, rectangle, rounded corners=4pt,line width=2pt] at (3,0) {};
			\node[rectangle, fill=uniSlblue40] (key1) at (0, 0) {algebraic operations};
			\node[rectangle, fill=uniSblue, text=white] (key2) at (4, 0) {in-/outputs};
			\node[rectangle, fill=red!20] (key3) at (8, 0) {loss};
			\node[draw=none,fill=none, minimum width=0cm] at (-3,0) {\textbf{key:}};
		\end{scope}
		
	\end{scope}
	
	\draw[line width=1pt, -stealth] (input) -- (blackbox);
	\draw[line width=1pt, -stealth] (trivialinput) -- (Lbox);
	\draw[line width=1pt, -stealth] (blackbox) -- (out1);
	\draw[line width=1pt, -stealth] (blackbox) -- (out2);
	\draw[line width=1pt, -stealth] (out1) -- (out3);
	\draw[line width=1pt, -stealth] (out2) -- (out3);
	\draw[line width=1pt, -stealth] (out3) -- (out4);
	\draw[line width=1pt, -stealth] (out3) -- (loss);
	\draw[line width=1pt, -stealth] (Lbox) -- (out4);
\end{tikzpicture}
  \begin{tikzpicture}[remember picture, overlay]
    \draw[dashed, black, line width=1pt, -stealth] (-14.0,8.8) -- (-15.5,8.8) -- (-15.5,4.4);
    \draw[dashed, black, line width=1pt, -stealth] (-2.6,8.8) -- (0.1,8.8) -- (0.1,1.75) -- (-2.6,1.75);
    \draw[thick, rounded corners=10pt, gray] (-17,-0.2) rectangle (0.7,5.1);
  \end{tikzpicture}
  \vspace{0.4cm}
  \caption{Two approaches for constructing the effective elasticity tensor $\ol{\ull{C}}$ from microstructure: (a) direct numerical simulation (DNS) workflow, and (b) a machine-learning surrogate model employing spectral normalization to enforce two-sided Löwner bounds. Note that $\ull{L}$ depends on the sample via the phase volume fractions and the individual phase's stiffness tensor.}
  \label{fig:MLmodel}
\end{figure}

The reparameterization through $\ullWT{Q}$, $\ullWT{\Lambda}$ can be combined with machine learning to predict $\ullWH{\WT{Y}} \approx \ullWT{Y}$.
Therefore, two distinct ingredients are needed (see also \Cref{fig:MLmodel}):
\begin{itemize}
    \item a vector-valued surrogate $\ulWH{\xi}_\lambda \in [0, 1]^m$ to model the eigenvalues in $\ullWTH{\Lambda}$ and
    \item a parameterized orthogonal matrix $\ullWTH{Q}\;(\ulWH{\xi}_\mathrm{q}) \in Orth(\ffR^m)$.
\end{itemize}
The first part is trivial to enforce, e.g., through a simple sigmoid function (or similar alternatives). Parameterization of the orthogonal matrix~$\ullWT{\WH{Q}}$ is more challenging and less commonly available. In our implementation\footnote{\href{https://github.com/DataAnalyticsEngineering/VoigtReussNet}{https://github.com/DataAnalyticsEngineering/VoigtReussNet}} using \texttt{pytorch}~\cite{Ansel_PyTorch_2_Faster_2024}, we deploy \python{torch.nn.parameterizations.orthogonal}\footnote{\href{https://pytorch.org/docs/stable/generated/torch.nn.utils.parametrizations.orthogonal.html}{\texttt{torch.nn.utils.parametrizations.orthogonal.html}}}. In practice, the black box surrogate model must provide $\ulWH{\xi}_\mathrm{q}$, matching the parameterization used.

A relative error with respect to the range of admissible responses $\Voigt{\ull{\ol{Y}}}-\Reuss{\ull{\ol{Y}}}$ of an approximation $\ullWH{\ol{Y}}$ of $\ull{\ol{Y}}$ can directly be obtained from the absolute error in the spectrally normalized $\ullWT{Y}$:
\begin{align}
    \phi( \ullWT{Y}, \ullWTH{Y} ) &= \frac{1}{\sqrt{m}}\; \Vert \ullWT{Y} - \ullWTH{Y} \Vert_\mathsf{F} \; , \label{eq:rel:error}
\end{align}
where $\left\| \cdot \right\|_\mathsf{F}$ denotes the Frobenius norm. This loss is insensitive to changes of unit (e.g., SI vs. imperial), and it can be combined with any other dedimensionalized loss. Therefore, we strongly recommend the use of a loss that is exclusively based on the normalized quantities $\ulWTH{Y}$ vs. $\ullWT{Y}$ instead of comparing the actual constitutive target $\ol{\ull{Y}}$ to the prediction $\ullWH{\ol{Y}}$. Notably, the error~$\phi( \ullWT{Y}, \ullWTH{Y} )$ is constrained to the unit interval $[0, 1]$, rendering it well interpretable for each sample as a relative error of the stiffness normalized to the range defined by the rigorous first-order bounds.
This procedure is also in line with the problem-specific weighting of losses in \cite{Fernandez2021}, which was also shown to be crucial for success when learning a model with outputs ranging over several orders of magnitude. Therefore, the loss will be based on \cref{eq:rel:error}.

\begin{remark}
    In our experience, it is prohibitive to compute the loss based on the parameters $\ul{\xi}_\mathrm{q}$, or on $\ul{\xi}_\lambda$ for the following reasons:
    \begin{itemize}
        \item Redundancy is possible, i.e. different $\ulWT{\xi}_\mathrm{q}$ can lead to the same $\ullWH{\WT{Q}}$.
        \item Depending on the chosen parameterization, errors in the parameters are hard to interpret since the usual norms of $\ulWT{\xi}_\mathrm{q}$ do not correspond straightforwardly to errors in the orthogonal matrix.
        \item As for $\ul{\xi}_\lambda$, the ordering of eigenvalues in the surrogate model could be different. The situation is even worse if eigenvalues show algebraic multiplicity, which renders the columns of $\ullWT{Q}$ nonunique.
    \end{itemize} 
    These problems are eliminated when using the suggested loss \cref{eq:rel:error} or any other loss relating $\ullWTH{Y}$ to $\ullWT{Y}$.
\end{remark}

As mentioned before, spectral normalization is independent of the type of machine-learning model to be used, i.e., it can be combined with neural networks (in their different flavors), kernel methods, random forests, and many more. The only two essential ingredients are that the outputs must be the vector $\ulWT{\xi}_\mathrm{\lambda}\in [0, 1]^m$ and the orthogonal matrix $\ullWT{Q}\in Orth(\ffR^m)$.

\section{Application to 3D linear mechanical homogenization}
\label{sec:3d_mechanics}
\begin{figure}[!htb]
    \centering
    \begin{tabular}{cccc}
        \subcaptionbox{Fiber system                         \label{fig:subfig1}}{\includegraphics[width=0.18\textwidth,trim={17cm 0 17cm 3cm},clip]{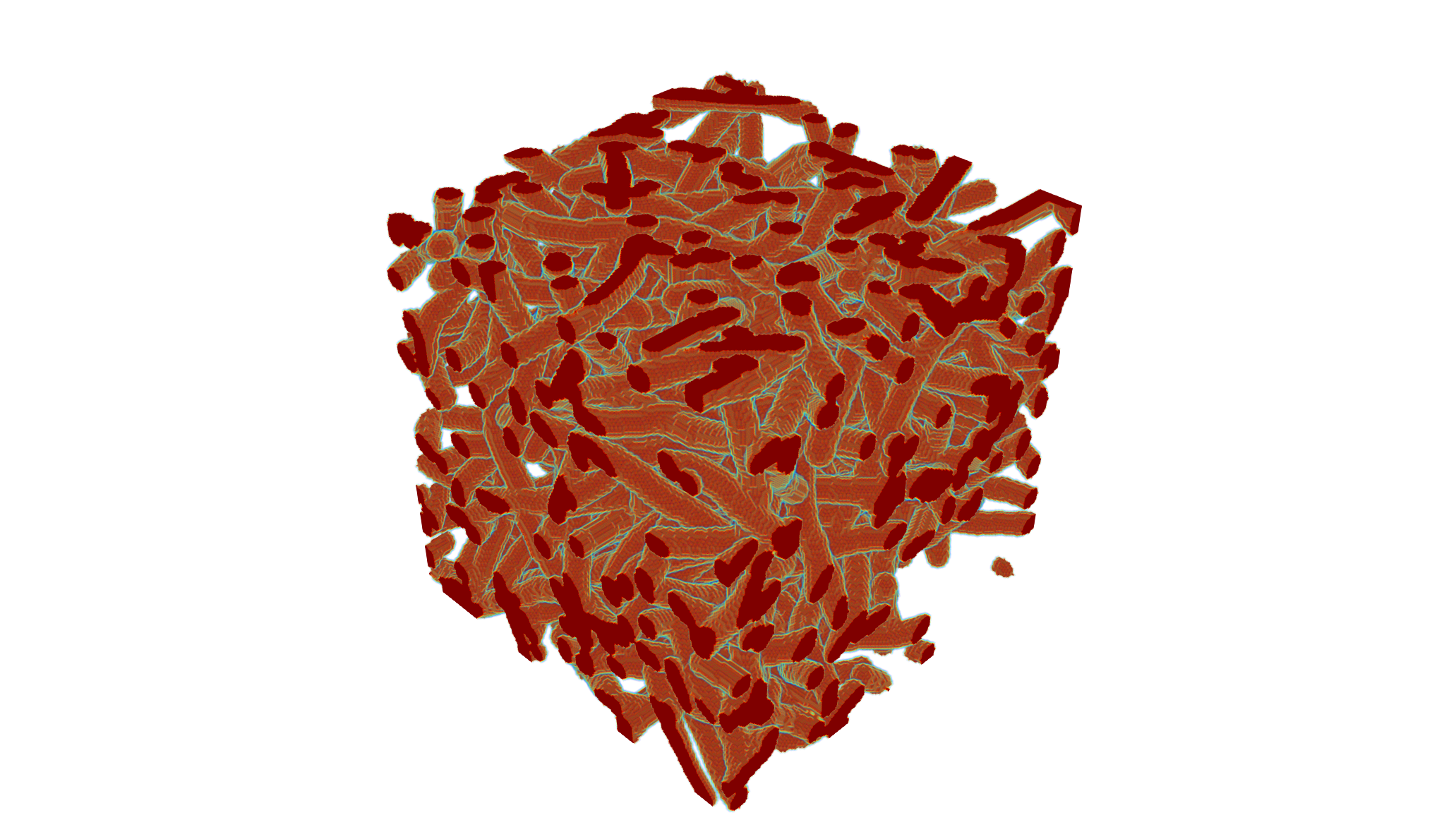}} & 
        \subcaptionbox{Channel system                       \label{fig:subfig2}}{\includegraphics[width=0.18\textwidth,trim={17cm 0 17cm 3cm},clip]{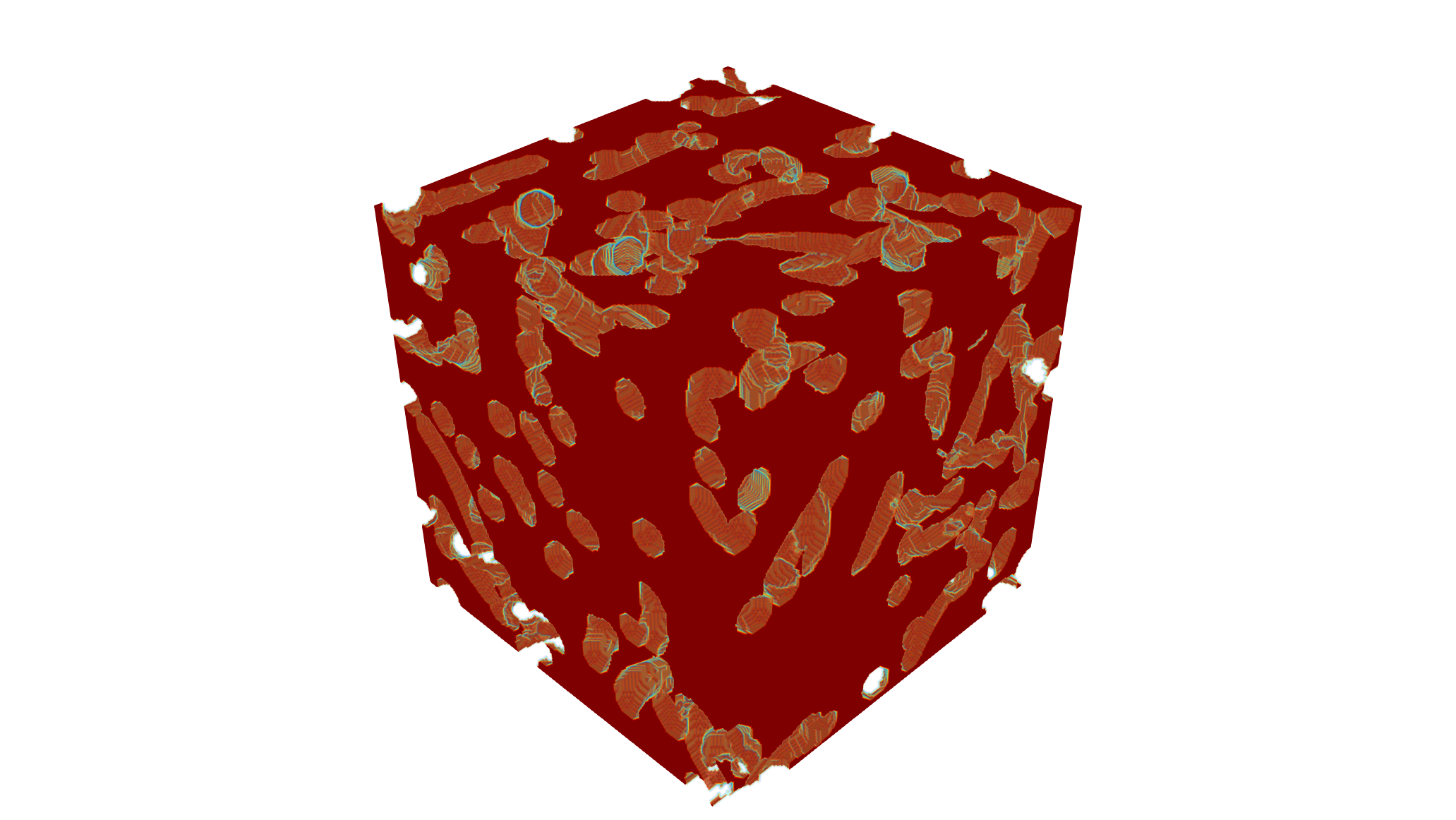}} & 
        \subcaptionbox{Level set of Gaussian random fields  \label{fig:subfig4}}{\includegraphics[width=0.18\textwidth,trim={17cm 0 17cm 3cm},clip]{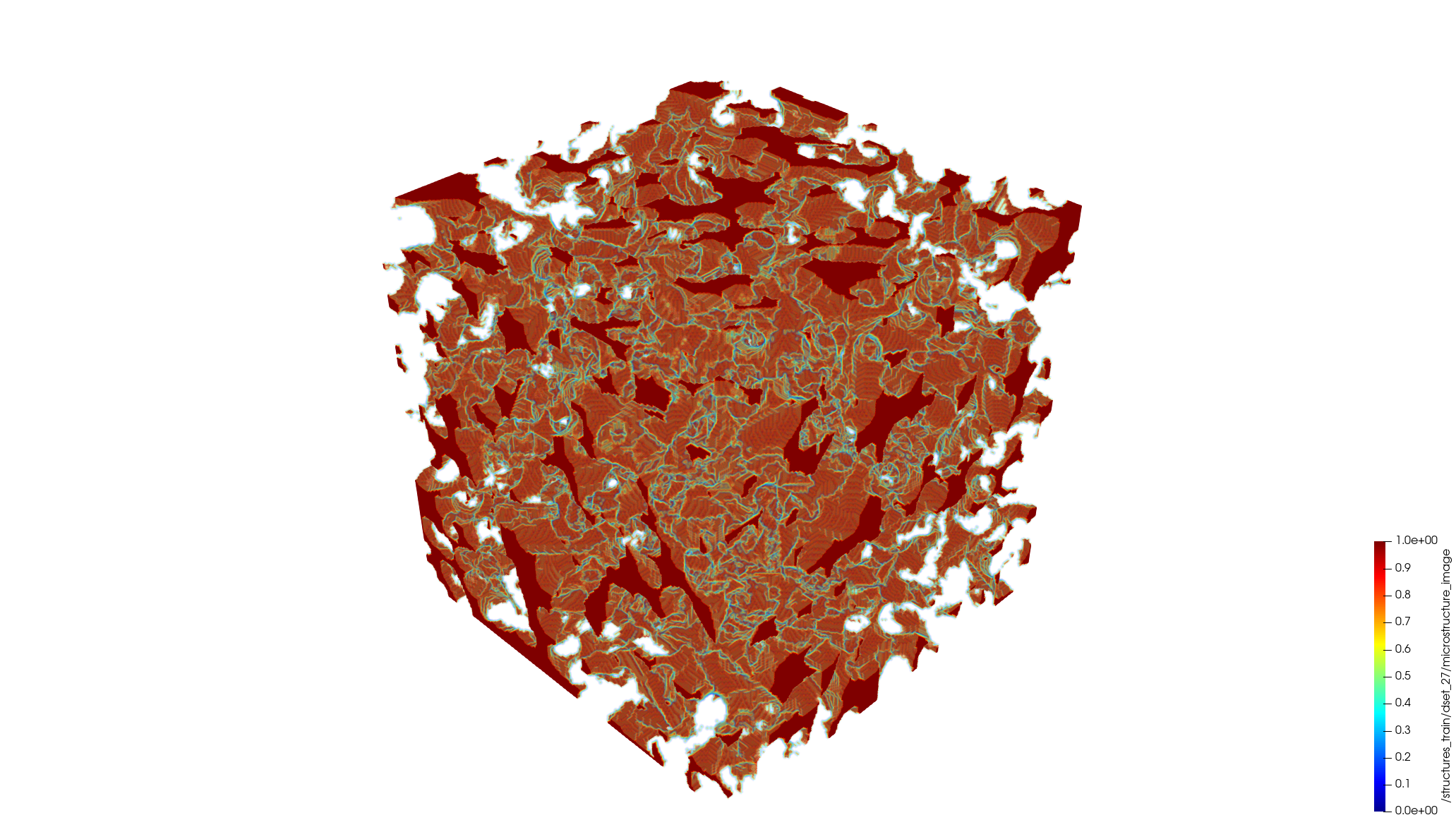}} & 
        \subcaptionbox{Spinodal system                      \label{fig:subfig5}}{\includegraphics[width=0.18\textwidth,trim={17cm 0 17cm 3cm},clip]{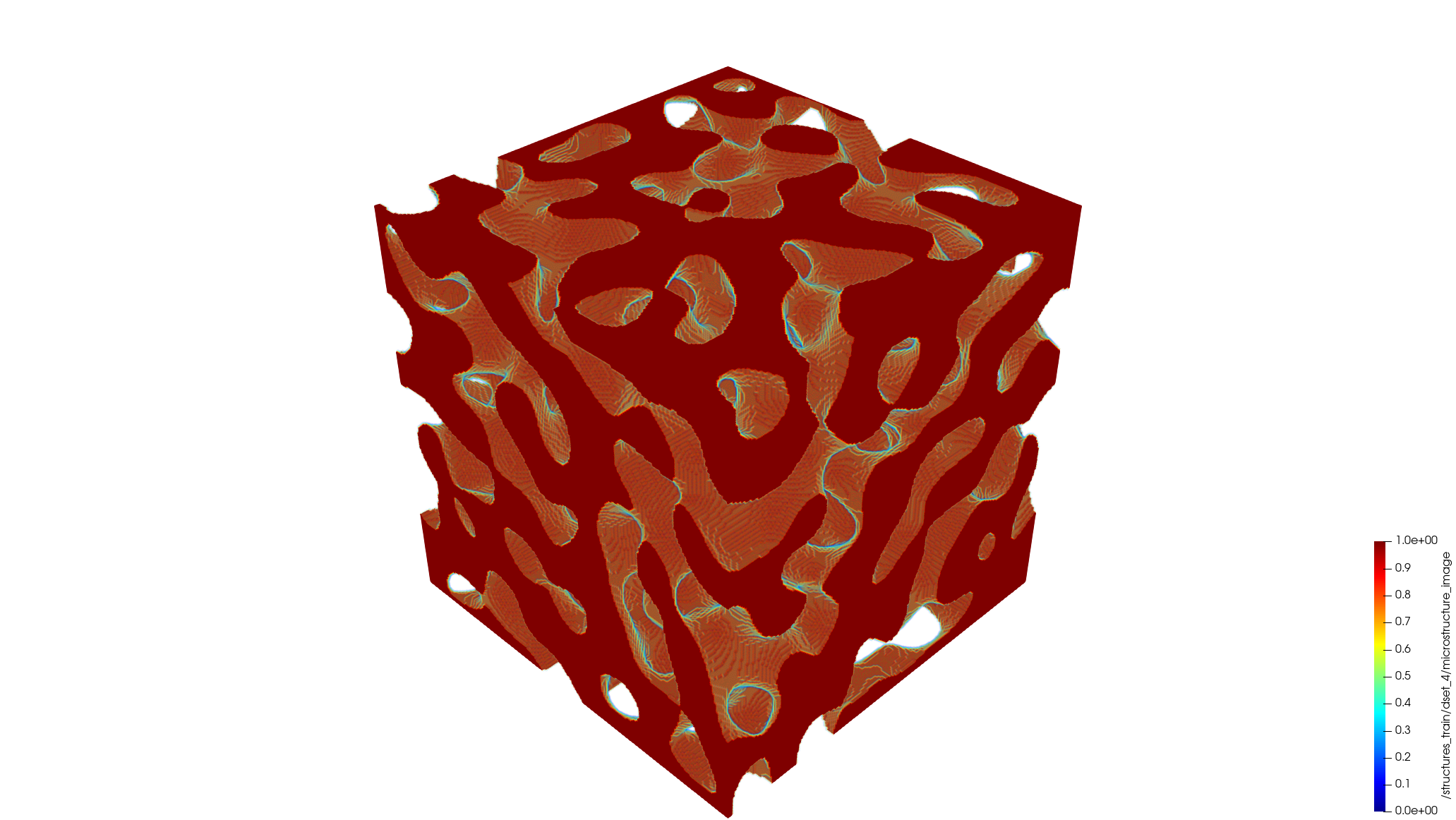}} \\[.7cm]
        \subcaptionbox{Hard ellipsoids                      \label{fig:subfig6}}{\includegraphics[width=0.18\textwidth,trim={17cm 0 17cm 3cm},clip]{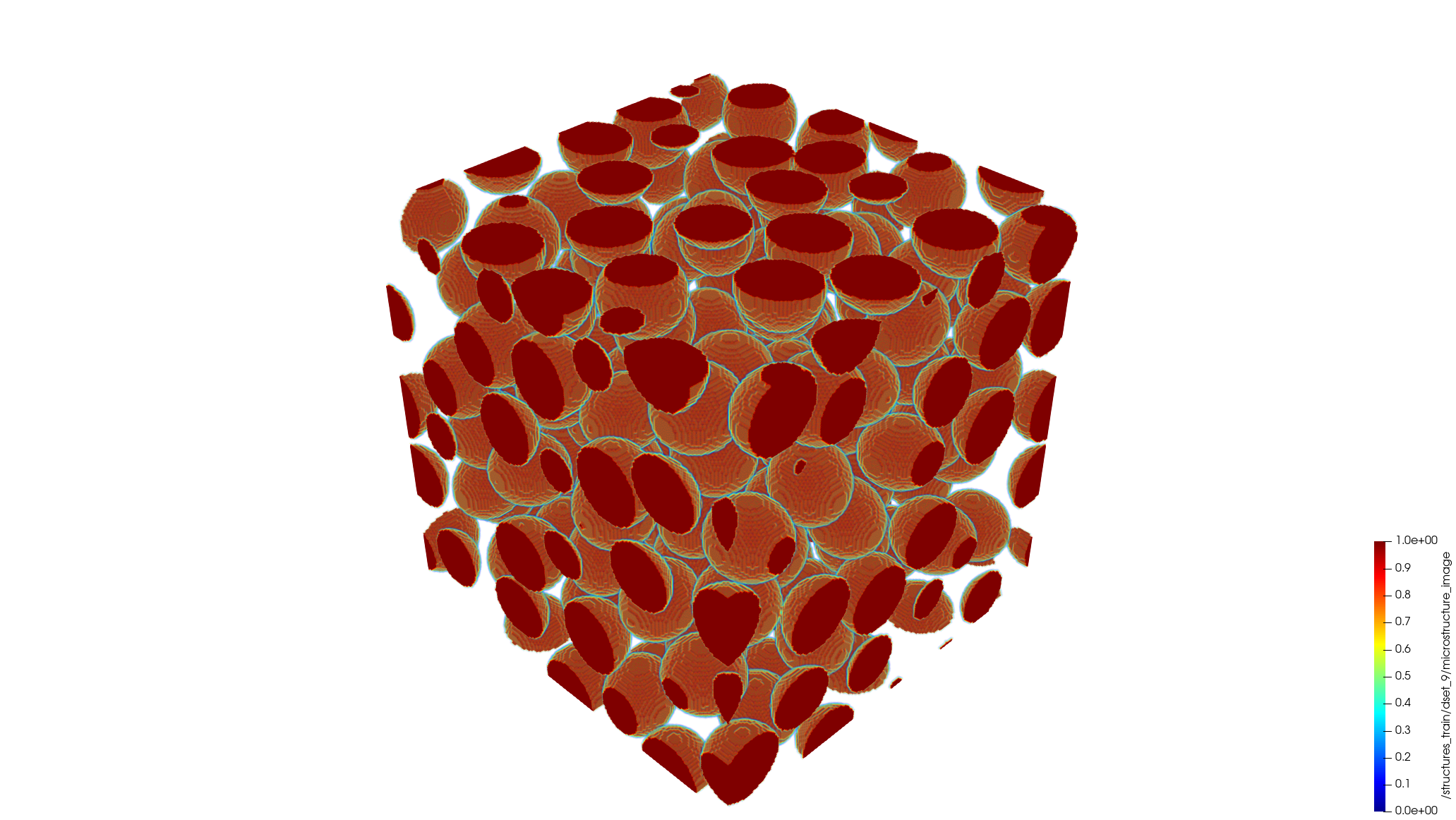}} &
        \subcaptionbox{Smoothed hard ellipsoids             \label{fig:subfig7}}{\includegraphics[width=0.18\textwidth,trim={17cm 0 17cm 3cm},clip]{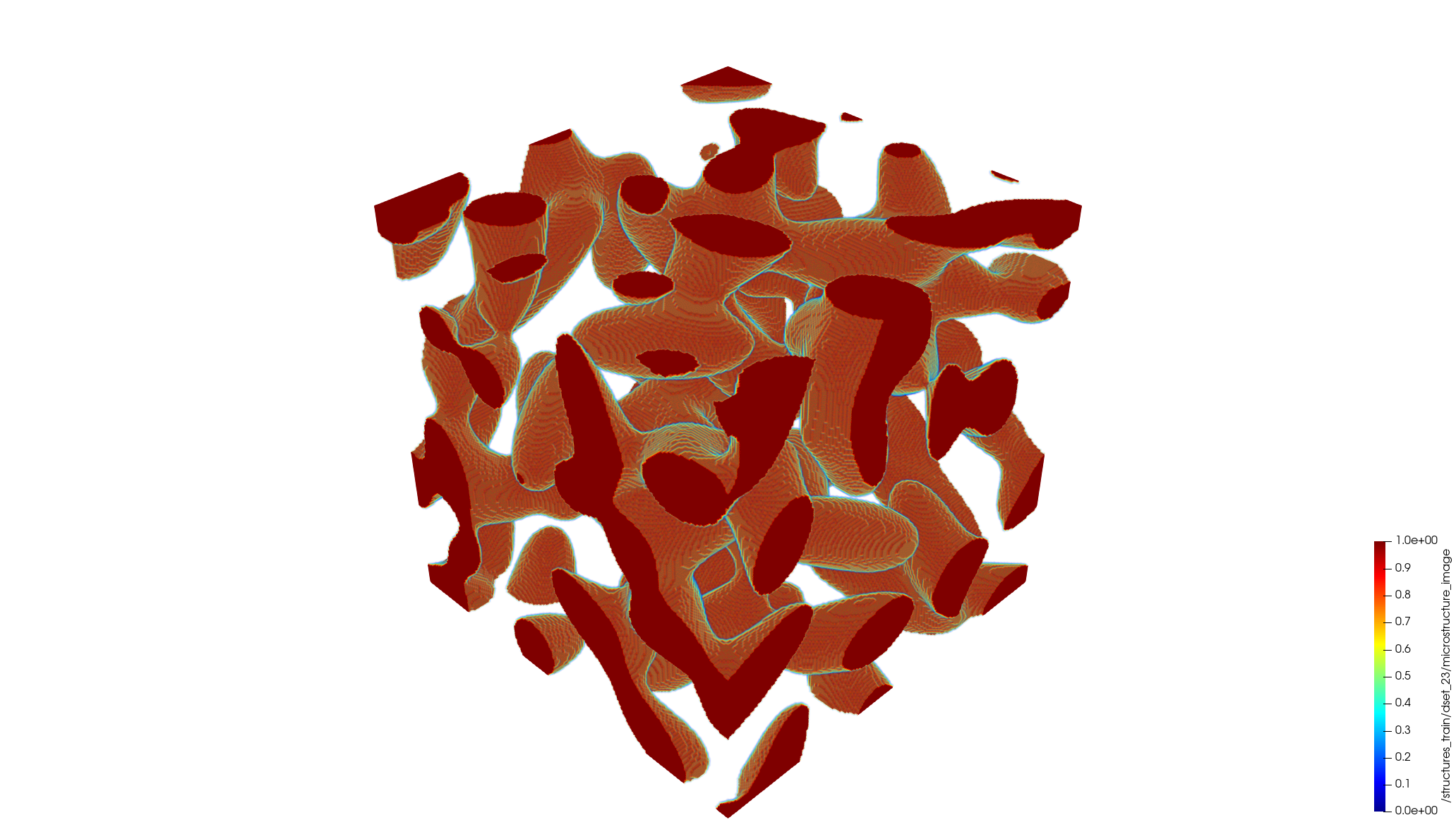}} &
        \subcaptionbox{Soft ellipsoids                      \label{fig:subfig8}}{\includegraphics[width=0.18\textwidth,trim={17cm 0 17cm 3cm},clip]{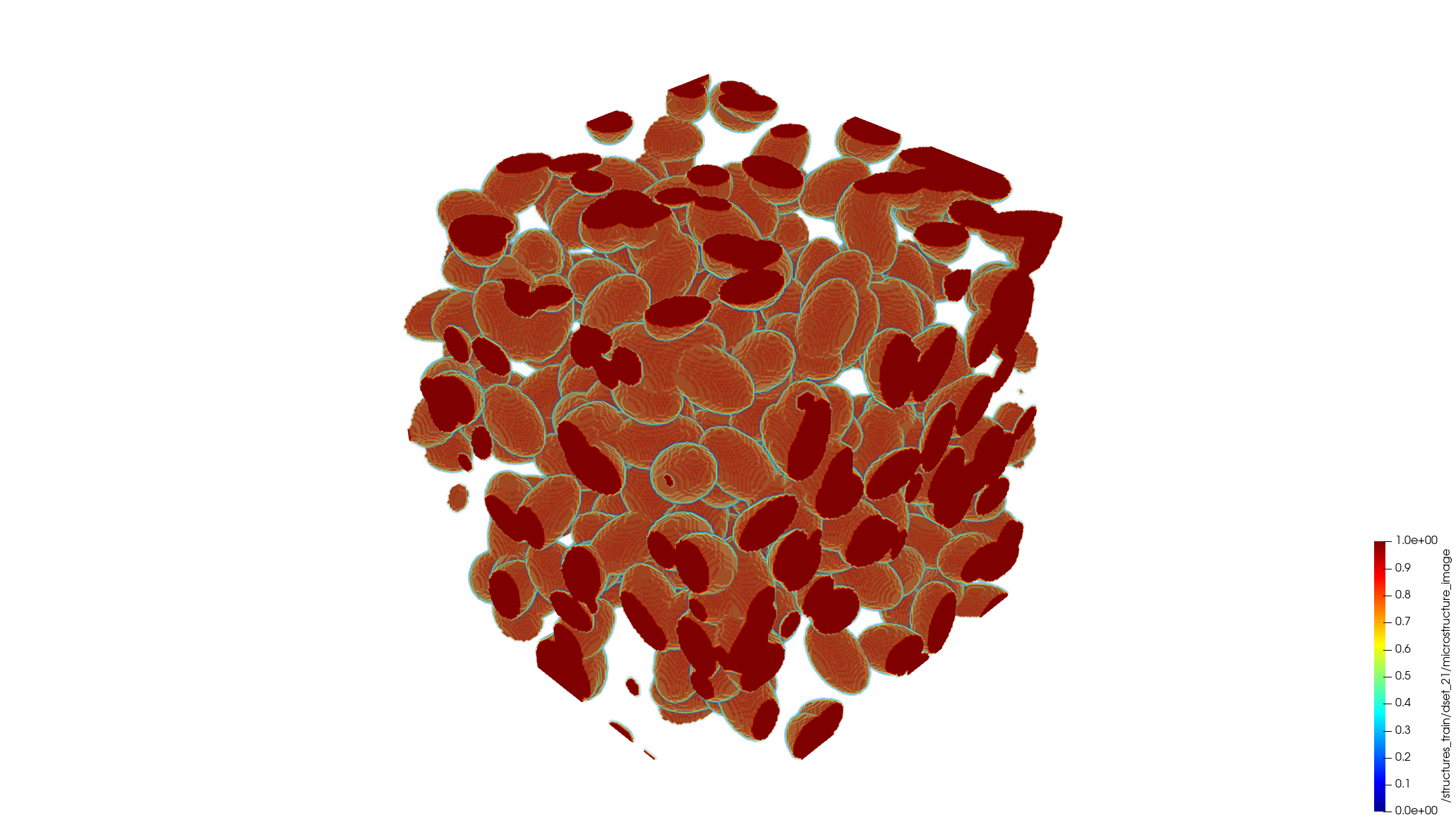}} & 
        \subcaptionbox{Smoothed soft ellipsoids             \label{fig:subfig9}}{\includegraphics[width=0.18\textwidth,trim={17cm 0 17cm 3cm},clip]{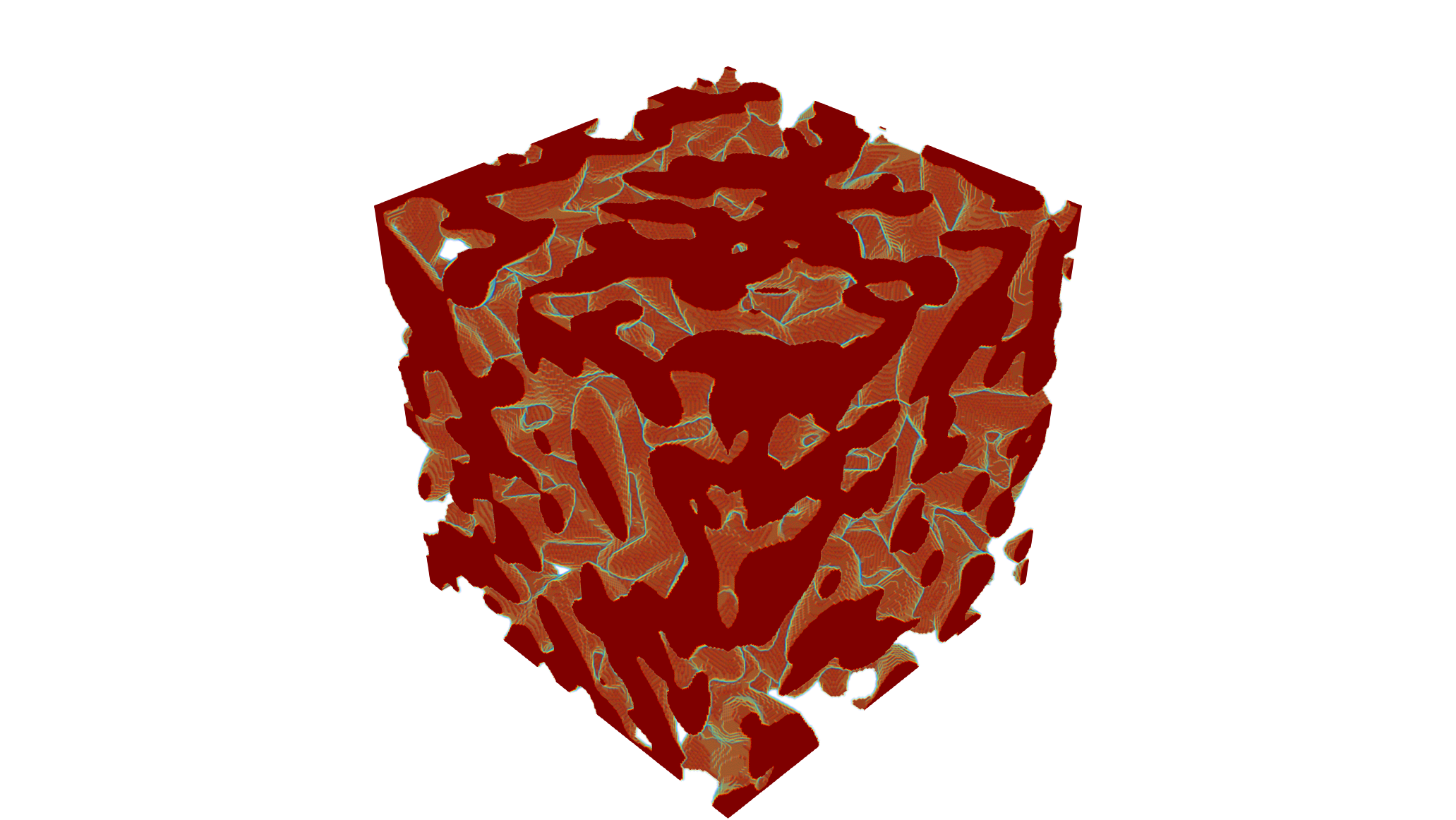}}
    \end{tabular}
    \caption{\protect Examples of 3D microstructures taken from \cite{ulm2020_dataset} for eight of the nine classes (spatial stochastic graphs omitted here). The inclusion phase is shown in red, and the matrix is shown as a void for visibility. Each class contributes $10{,}000$ samples.}
    \label{fig:example_3D_microstructures}
\end{figure}

Owing to the origins of the spectral normalization, we refer to our surrogate as \VRNet{} and apply it in a 3D mechanical setting. The goal is to predict the effective fourth-order stiffness tensor $\ol{\ull{C}} \in \mathrm{Sym}^+(\ffR^6)$ of periodic, biphasic composites from geometric descriptors of the unit cell and phase moduli. 
All microstructures are drawn from the open 3D dataset of \cite{ulm2020_dataset,Prifling2021}, which comprises nine stochastic model classes designed to span a broad morphological envelope (volume fraction, connectivity, length scales); see \Cref{fig:example_3D_microstructures}. 
Each microstructure is represented by a $192^3$ voxel image, and it is labeled by 236 image-derived scalar descriptors.\footnote{As in~\cite{Prifling2021}, these include porosity, specific surface area, constrictivity, chord-length and spherical contact distributions, geodesic tortuosity, two-point correlations, and size/shape statistics. They are predominantly isotropy-averaged descriptors; we return to this point below.}

We consider the isotropic bulk and shear moduli $(K_1, G_1)$ for phase~1 and $(K_2, G_2)$ for phase~2.
To reduce dimensionality, we dedimensionalize the remaining parameters relative to $K_1$, leading to three distinct parameters:
\begin{align}
    \pi_1 = \frac{K_2}{K_1}, 
    \qquad
    \pi_2 = \frac{G_1}{K_1}, 
    \qquad
    \pi_3 = \frac{G_2}{K_1} \, .
\end{align}
Due to the linearity of the elasticity problem, any global scaling by $K_1$ can be applied a posteriori. 
To cover a wide range of contrasts, we sample 
\begin{align}
  \pi_i = 10^{\pi_i^\star},\qquad \pi_i^\star \sim \cU(-2,2), \quad i=1,2,3,
\end{align}
i.e., $\pi_i \in [10^{-2},10^{2}]$ is uniformly sampled in the logarithmic space, covering four orders of magnitude in each of the parameters~$\pi_1, \pi_2, \pi3$.

For each microstructure, we draw triples $(\pi_1,\pi_2,\pi_3)$. We then solve periodic linear elasticity problems using our FFT-based homogenization code (FANS~\cite{FANS_github,Leuschner2017}) to obtain the effective elasticity tensor $\ol{\ull{C}}$ for $K_1=1$ (physical units are recovered by subsequent scaling by $K_1$). 
Overall, $1.18$ million microstructure-material combinations were considered, and for each of these, six simulations comprising $21.2$ million degrees of freedom were executed. The data were split into train/validation/test, see also~\cref{tab:3d_mech_datasets}.

\begin{figure}[!htb]
    \centering
    \includegraphics[scale=1.0]{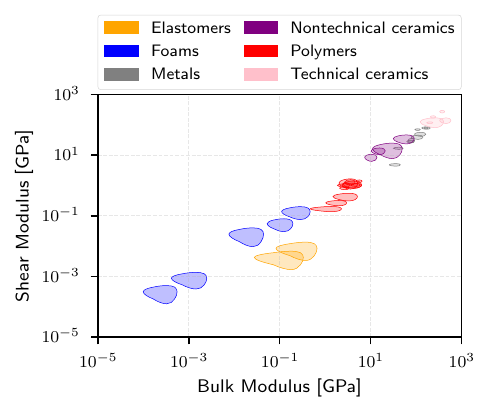}
    \includegraphics[scale=1.0]{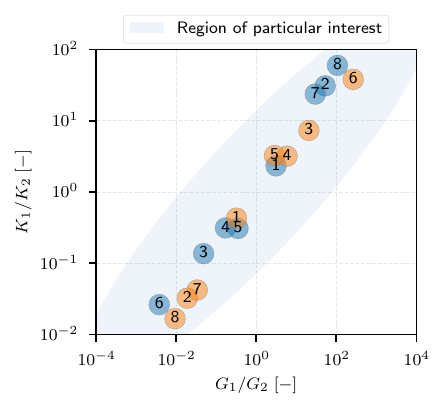}
    \caption{Phase contrasts of the bulk and shear moduli for the material pairs from~\cref{tab:binary_material_pairs}. Each material pair is represented twice, considering the possibility of the two materials being in either phase 1 or phase 2.}
    \label{fig:moduli_ratios}
\end{figure}

In the~\cref{fig:moduli_ratios} (Left), an Ashby-style scatter of individual engineering materials (elastomers, polymers, woods/foams, concretes, composites, metals, glasses/ceramics) is shown in the $(K, G)$ plane. 
The cloud is narrowly confined between lines $G \approx \alpha K$ with $\alpha = 3(1-2\nu)/(2(1+\nu))$ induced by the isotropic relation $G/K = 3(1-2\nu)/(2(1+\nu))$, i.e., material points fall on a low-dimensional manifold parameterized mainly by Poisson's ratio $\nu$. 
Typical ranges ($\nu\approx0.15$--0.35 for metals/ceramics, higher for many polymers, very high for near-incompressible elastomers) generate $\alpha$ spanning $\sim 10^{-2}$ (rubbery, nearly incompressible) to $\sim 1$ (low $\nu$ ceramics), which explains the visually linear alignment.
In the~\cref{fig:moduli_ratios} (Right), the plotted binary material pairs (each shown twice by swapping phase roles) inherit this proportional structure: both $K$- and $G$-contrasts largely move along similar $G=\alpha K$ rays because each constituent's Poisson ratio sits in the physically admissible band. Expressed in our non-dimensional parameters, this panel maps directly to the contrast plane $(1/\pi_1,\pi_2/\pi_3) = (K_1/K_2, G_1/G_2)$. 
The real pairs cluster within a corridor already densely covered by our log-uniform sampling $\pi_i \in [10^{-2},10^{2}]$; hence the training set spans (i) the physically prevalent region occupied by established engineering materials, (ii) high-contrast extrapolations (extreme $K$ or $G$ mismatch), and (iii) regimes approaching near-incompressibility (very small $G/K$).

\begin{table}[!htb]
    \centering
    \caption{Summary of the datasets considered for the 3D mechanical homogenization problem.}
    \label{tab:3d_mech_datasets}
    \begin{tabular}{lccc}
        \toprule
        \textbf{Dataset} & \textbf{Material parameters} & \textbf{Microstructures} & \textbf{Samples} \\[0.1cm]
        \midrule
        Train      & \multirow{3}{*}{$\pi_i = 10^{\pi^*_i},\ \pi^*_i \sim \cU(-2,2)$}           & 63,000 & 751,089 \\[0.1cm]
        Validation &                                                                            & 13,500 & 263,188 \\[0.1cm]
        Test       &                                                                            & 13,500 & 170,753 \\
        \bottomrule \\[-0.15cm]
        Total      &                                                                            & \textbf{90,000} & \textbf{1,185,030} \\ 
        \bottomrule
    \end{tabular}
\end{table}

\begin{table}[!htb]
    \centering
    \caption{Typical bulk and shear moduli for some common composite microstructured materials at room temperature}
    \label{tab:binary_material_pairs}
    \begin{tabular}{lccccccc} 
    \toprule
    \multicolumn{4}{c}{\textbf{Composite}}  & \multicolumn{2}{c}{\textbf{Phase 1} (GPa)} & \multicolumn{2}{c}{\textbf{Phase 2} (GPa)} \\
    \cmidrule(lr){5-6} \cmidrule(lr){7-8}
    {No.} & {} & {} & {} & {Bulk $K_1$} & {Shear $G_1$} & {Bulk $K_2$} & {Shear $G_2$}      \\
    \midrule
    1 & Steel         & -- & Aluminum          & 156 - 165 & 79.3   & 68 - 70  & 25.5       \\
    2 & Steel         & -- & Epoxy             & 156 - 165 & 79.3   & 5.2      & 1.5        \\
    3 & Epoxy         & -- & E-Glass           & 5.2       & 1.5    & 38       & 31         \\
    4 & Aluminum      & -- & Silicon carbide   & 68 - 70   & 25.5   & 220      & 150        \\
    5 & Cement        & -- & Aggregate         & 10 - 30   & 7      & 45 - 85  & 20         \\
    6 & Polyethylene  & -- & E-Glass           & 1         & 0.1    & 38       & 31         \\
    7 & Copper        & -- & Epoxy             & 123       & 44.7   & 5.2      & 1.5        \\
    8 & Tungsten      & -- & Epoxy             & 307 - 314 & 160    & 5.2      & 1.5        \\
    \bottomrule
    \end{tabular}
\end{table}

The surrogate for the 3D mechanical problem operates on a 239-dimensional feature vector where 236 scalars are morphology descriptors~\cite{Prifling2021}, and the remaining three encode the (dimensionless) elastic phase contrasts $\pi_1, \pi_2, \pi_3$.
All input features are range-normalized prior to training. 
As in~\cref{sec:spectral:normalization}, the \VRNet{} predicts normalized degrees of freedom $\ulWH{\xi}$ comprising of six eigenvalues $\lambda\{\ullWTH{C} \}^6_i \in [0, 1]$ and an orthogonal factor parameterized by $15$ scalars $(\WT{q}_1,\dots,\WT{q}_{15})$ also in $[0, 1]$, leading to a total of $21$ outputs.
The spectral inverse normalization $\cS^{-1}$ maps $\ullWTH{C}$ to $\ullWH{\ol{C}}$.

The regressor is realized as a fully connected feed-forward neural network with five hidden layers of widths [1024, 512, 256, 128, 128], taking the 239-dimensional input $\ul{\chi}$ and outputting the 21 normalized DOFs $\ulWH{\xi}$. Each hidden layer is followed by batch normalization and a mixed activation (combination chosen for stable convergence; identity components act as linear bypass channels).
Training uses the \texttt{AdamW} optimizer with a \texttt{ReduceLROnPlateau} scheduler (patience-based learning rate halving) until validation stagnation. The loss is the mean of the Voigt-Reuss normalized error measure~\cref{eq:rel:error} over all samples. It is dimensionless and confined to $[0,1]$. The model is trained on $\sim 7.5\times 10^5$ samples (train split). 

\begin{figure}[!htb]
    \centering
    \includegraphics[scale=1]{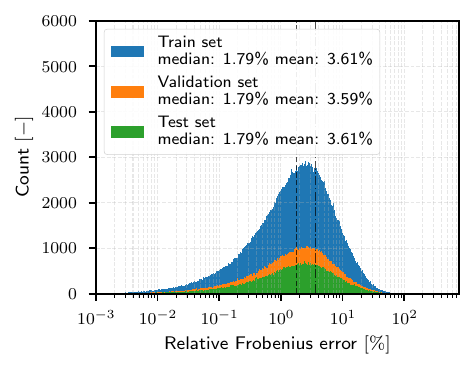}
    \includegraphics[scale=1,trim={0cm 0.0cm 7cm 0cm},clip]{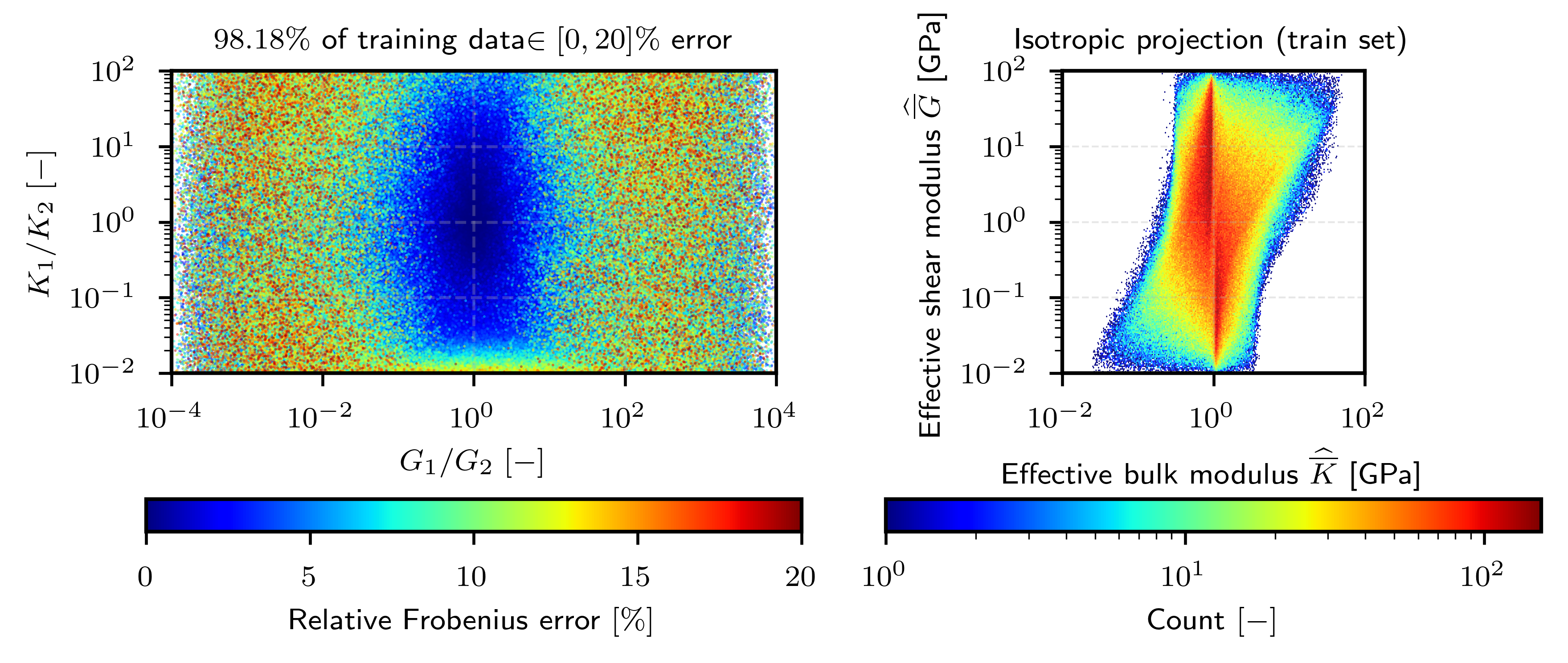}
    \caption{
    \textbf{Left:} Relative Frobenius error histograms $\cE_\mathsf{F}$ for train/val/test splits (nearly coincident; mean $\approx 3.6\%$, median $\approx 1.8\%$).
    \textbf{Right:} Relative Frobenius error of the predicted effective stiffness tensor for different phase contrasts. Plot shows the lower errors (0 - 20\%) covering 98.18\% of the data; extreme outliers are not shown}
    \label{fig:mech3d_rel_error_histogram}
\end{figure}

\Cref{fig:mech3d_rel_error_histogram} (left) shows the distribution of $\cE_\mathsf{F}$ for train/validation/test; the histograms nearly overlap with essentially identical mean/median, indicating strong generalization across the datasets.
The mean error is $\approx 3.6\%$, and the median is $\approx 1.8\%$, evidencing a consistent geometry and material parameter to the effective tensor mapping learned by \VRNet{}.
\Cref{fig:mech3d_rel_error_histogram} (right) shows the dependence of the relative error on phase contrast for the training data. The plot shows errors up to 20\% covering approximately 98.18\% of the data. The $\approx$1.8\% extreme outliers, which are not represented herein, typically appear when the phase contrast is near the limits of the logarithmic sampling (i.e., when moduli ratios are either very large or very small) or when the microstructure lies at the boundary of the considered morphological diversity. In addition, composite pairs representing real-world materials from ~\cref{tab:binary_material_pairs} fall within the region of relatively low error, which supports the model's capacity to achieve accurate predictions for practical composites such as, e.g., steel-aluminum, aluminum-SiC, and epoxy-glass systems.

\begin{figure}[!htb]
    \centering
    \begin{overpic}[scale=1]{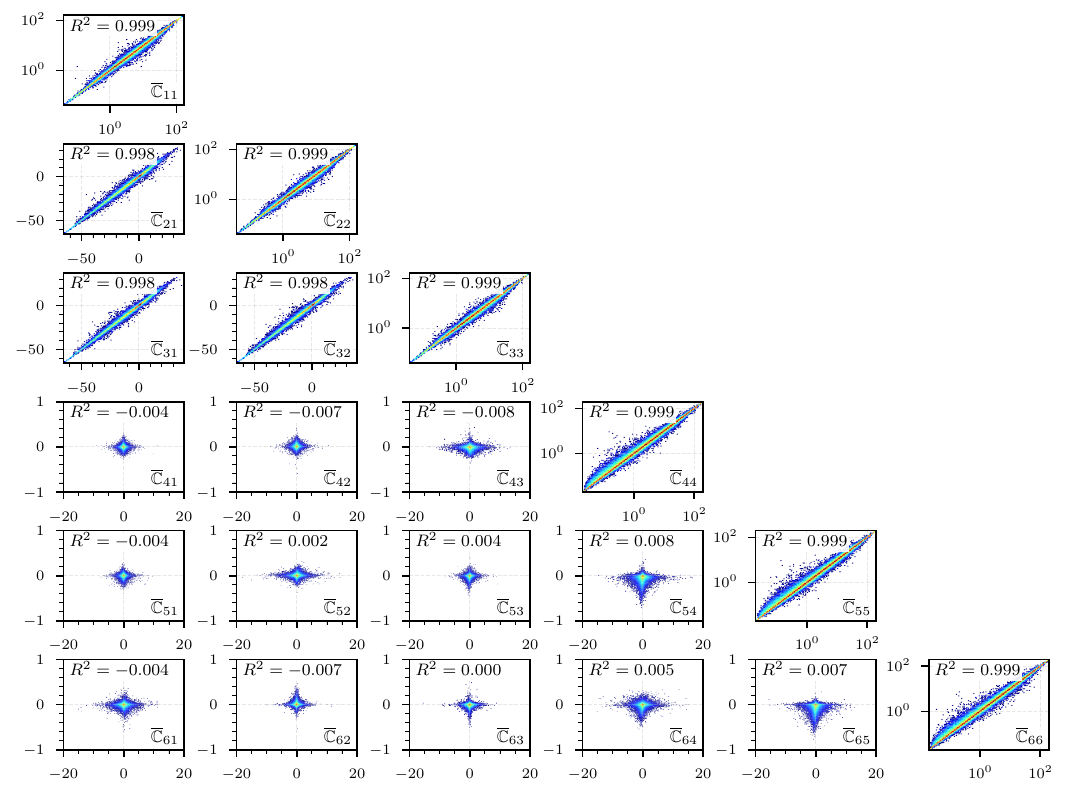}
        \put(50,55.5){%
            \begin{minipage}{0.5\textwidth}
                \centering
                \textbf{Symmetric}
            \end{minipage}
        }
    \end{overpic}
    \caption{
    Ground truth vs. prediction for all the stiffness components of $\ol{\ull{C}}$ (hexbin density) by the \VRNet{}. Each tile lists per-component $R^2$.
    All entries corresponding to isotropy achieve $R^2\ge 0.998$: the axial terms $\{\ol{C}_{11},\ol{C}_{22},\ol{C}_{33}\}$, shear terms $\{\ol{C}_{44},\ol{C}_{55},\ol{C}_{66}\}$, and the normal couplings $\{\ol{C}_{21},\ol{C}_{31},\ol{C}_{32}\}$.
    The remaining (anisotropy-revealing) components have $R^2\approx 0$.
    }
    \label{fig:mech3d_r2_matrix}
\end{figure}

\Cref{fig:mech3d_r2_matrix} assembles a component-wise view of predictive accuracy for the effective tensor $\ull{\ol{C}}$ by the \VRNet{} surrogate.
The lower triangular panels display the ground truth versus prediction (hexbin densities) for all individual components, along with their corresponding $R^2$ scores.
Two distinct regimes emerge sharply.
First, the isotropic block i.e., the three normal moduli $(\ol{C}_{11},\ol{C}_{22},\ol{C}_{33})$, the three shear moduli $(\ol{C}_{44},\ol{C}_{55},\ol{C}_{66})$, and the symmetric normal couplings $(\ol{C}_{21},\ol{C}_{31},\ol{C}_{32})$ exhibit near-perfect agreement with $R^2\ge 0.998$.
Second, all remaining (anisotropy-revealing) entries that encode departures from isotropy (shear-normal couplings and inter-shear couplings) show $R^2\approx 0$.
This pattern is expected given the 236 microstructure descriptors provided in~\cite{Prifling2021,ulm2020_dataset} are exclusively isotropic. 
The input feature map is invariant under $SO(3)$, hence there is no information available to break rotational degeneracy and predict anisotropic tensor parts that depend on orientation or directionality. 
Consequently, any surrogate can at best recover the isotropic projection of the effective stiffness $\ol{\ull{C}}_{\mathrm{iso}}$, where,
\begin{align}
  \ol{\ull{C}}_{\mathrm{iso}} &= \argmin_{\ol{\ull{C}}_{\mathrm{iso}} \in \mathrm{Sym}^+_{\mathrm{iso}}(6)} \Vert \ol{\ull{C}} - \ol{\ull{C}}_{\mathrm{iso}} \Vert_\mathsf{F}, \qquad
  \ol{\ull{C}}_{\mathrm{iso}} = 3 \; \ol{K} \; \frac{\fI \otimes \fI}{3} + 2 \;\ol{G} \; \ffP_2^\mathrm{iso}, \\
  \ol{K} &= \frac{1}{9} \sum_{i,j=1}^3 \ol{C}_{ij}, \qquad
  \ol{G} = \frac{1}{10} \lb \sum_{i=1}^6 \ol{C}_{ii} - 3 \;\ol{K}\rb, 
  \label{eq:isotropic_projection}
\end{align}
which fixes exactly those nine lower-triangular entries with high $R^2$.
All anisotropy revealing components lie in the orthogonal complement, i.e., $\ol{\ull{C}} - \ol{\ull{C}}_{\mathrm{iso}}$ and are a priori unpredictable from isotropic descriptors.

We trained three additional surrogates on the identical dataset (same train/validation splits), feature representation (same 239-dimensional inputs), same regressor and training protocol, but differing only in the output parameterization and loss function:
\begin{itemize}
  \item \textbf{Vanilla neural network:} directly predicts the 21 independent components of $\ol{\ull{C}}$ and reconstructs a symmetric matrix. This model has no built-in guarantee of positive definiteness or Voigt-Reuss admissibility. A mean squared error (MSE) loss on the components of $\ol{\ull{C}}$ is used for training.
  \item \textbf{Cholesky neural network~\cite{Xu2021}:} predicts the 21 independent components of the lower-triangular factor $\ullWH{L}$ and reconstructs $\ullWH{\ol{C}}=\ullWH{L}\;\ullWH{L}^\mathsf{T}$. This enforces $\ullWH{\ol{C}}\succeq 0$ by construction, but does not enforce two-sided Voigt-Reuss bounds. A MSE loss on the components of $\ull{L}$ is used for training.
  \item \textbf{Isotropic vanilla network (two-parameter predictor):} projects the target $\ol{\ull{C}}$ to the isotropic moduli $(\ol{K},\ol{G})$ via the orthogonal projection (\cref{eq:isotropic_projection}) and trains a two output network for $(\ol{K},\ol{G})$. The isotropic stiffness tensor $\ullWH{\ol{C}}_{\mathrm{iso}}$ is then reconstructed from the predicted moduli $(\WH{\ol{K}},\WH{\ol{G}})$. An output \texttt{ReLU} activation is used which ensures non-negativity of the predicted moduli and subsequently $\ullWH{\ol{C}}_{\mathrm{iso}} \succeq 0$. An MSE loss on $(\ol{K},\ol{G})$ is used for training.
  This directly addresses whether predicting only the two effective isotropic parameters is competitive when only isotropic descriptors are provided. 
\end{itemize}

\begin{figure}[!htb]
    \centering
    \includegraphics[scale=1]{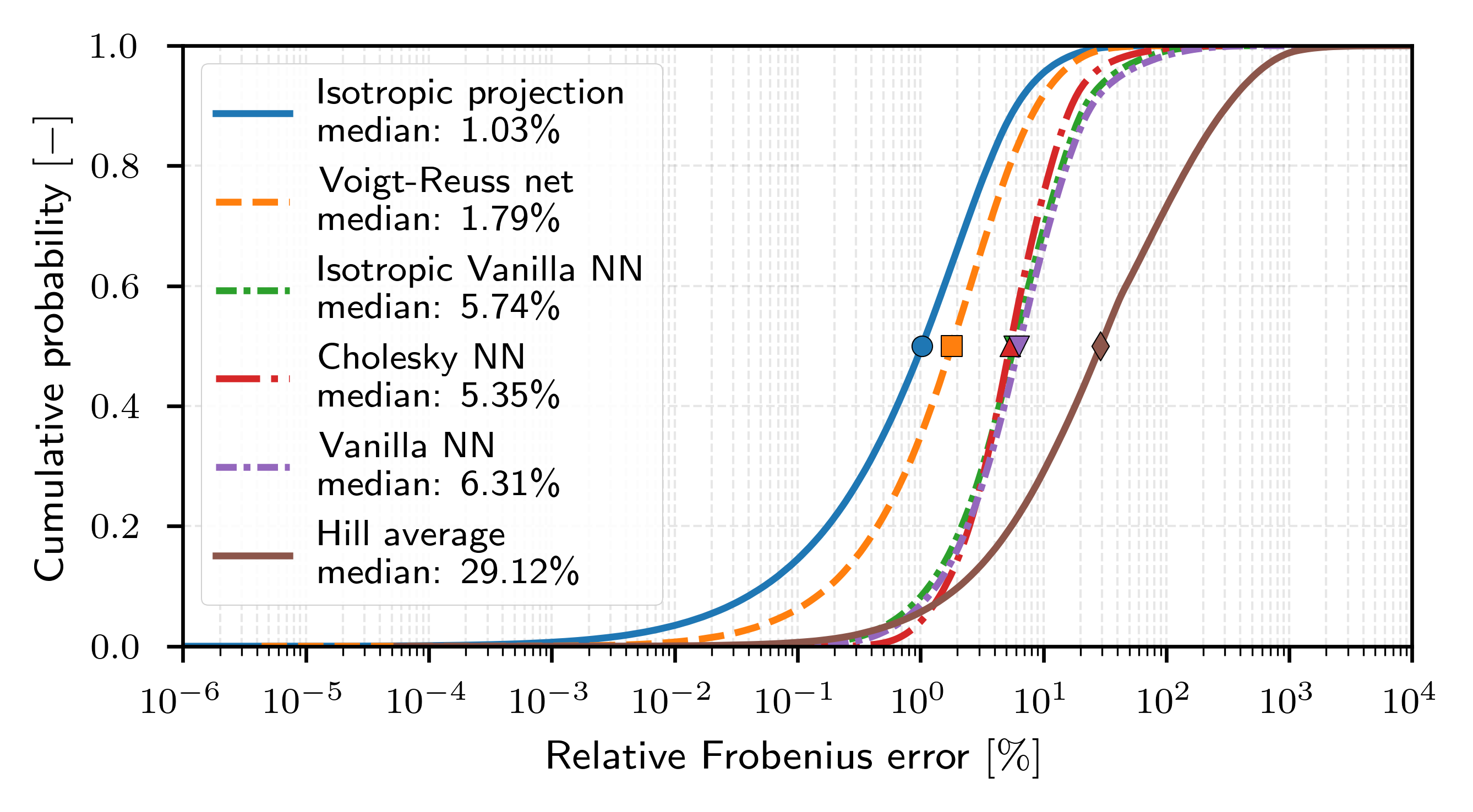}
    \caption{Empirical cumulative distribution functions of the relative Frobenius error $\cE_\mathsf{F}$ for the 3D mechanical problem.
    Curves compare \VRNet{} to three learned baselines (Cholesky network~\cite{Xu2021}, vanilla symmetric network, isotropic vanilla two-parameter network) and two analytical references (isotropic projection and Hill average).}
    \label{fig:mech3d_ecdf_train}
\end{figure}

\Cref{fig:mech3d_ecdf_train} compares empirical cumulative distribution functions (ECDFs) of the relative Frobenius error $\cE_\mathsf{F}$ for \VRNet{} against three learned baselines (Cholesky network~\cite{Xu2021}, vanilla neural network, isotropic vanilla two-parameter network), together with two analytical references, the isotropic projection and the classical Hill average,
\begin{align}
  \cE_{\mathrm{iso}}
  \;=\;
  \cE_\mathsf{F}\left(\ull{\ol{C}}\;,\;\ull{\ol{C}}_{\mathrm{iso}}\right), 
  \qquad
  \cE_{\mathrm{Hill}}
  \;=\;
  \cE_\mathsf{F}\left(\ull{\ol{C}}\;,\; \dfrac{\ull{\ol{C}}_{\mathrm{V}}+\ull{\ol{C}}_{\mathrm{R}}}{2}\right).
\end{align}
The isotropic projection error $\cE_{\mathrm{iso}}$ (median $1.03\%$) sets a natural lower floor for any surrogate that only receives isotropic descriptors as input.
In contrast, the Hill average incurs substantially larger discrepancies (median $29.12\%$), reflecting that simple mixing rules are not competitive once high material contrast and rich morphology are present.
The isotropic vanilla network explicitly targets only $(\ol{K},\ol{G})$, it remains noticeably less accurate than \VRNet{}, indicating that the bounded spectral representation is advantageous even in this isotropy-limited regime.
The \VRNet{} predictions (median $1.79\%$) lie very close to the isotropic floor and far below all the learned baselines, demonstrating that the network has learned to approximate the effective stiffness tensor with high accuracy within the information constraints imposed by the descriptors.
Quantitatively, \VRNet{} achieves the lowest mean/median errors and also strongly compresses the error spread among all learned models and, uniquely, produces zero violations of the Voigt and Reuss bounds by construction; see~\cref{tab:metrics_violations}.

As summarized in~\cref{tab:metrics_violations}, the vanilla neural network produces a non-negligible number of non-positive definite predictions, and both the vanilla and Cholesky networks violate the Voigt and Reuss bounds in a substantial fraction of cases, despite being trained on the same dataset, which we consider sufficiently large.
The isotropic vanilla baseline reduces (but does not eliminate) bound violations slightly by virtue of predicting only two non-negative moduli.
In contrast, \VRNet{} yields strictly admissible predictions for all samples and all splits, which is particularly important for use in inverse design and optimization.

Importantly, the residual gap of the \VRNet{} to the isotropic limit reflects both the limited information content of the current descriptor set, the finite network capacity, and specific hyperparameter choices of the network. In principle, more aggressive architectural and training optimization could move the \VRNet{} predictions even closer to this isotropic baseline, even without changing the inputs. 
Due to the invariance of the present descriptor set under $SO(3)$, the present learning task is effectively constrained to recovering the isotropic projection of $\ol{\ull{C}}$.
Providing anisotropy-sensitive descriptors (e.g., directional correlations, Minkowski tensors, band features, or convolutional image features) would supply the missing orientational information and make the anisotropy-revealing tensor components learnable.

These results confirm that the \VRNet{} achieves strong interpolation and generalization across an expansive space of 3D microstructures and elastic phase contrasts. It also maintains physically admissible predictions, adhering to the fundamental Voigt-Reuss bounds in the stiffness space, and consistently delivers accurate performance across the training, validation, and testing sets, with a median error of a few percent. Crucially, because the network was trained on a large and diverse dataset that includes real-world composite parameter ranges, it is well positioned to handle practical engineering scenarios where bulk and shear moduli vary within similar bounds.

\begin{table}[!htb]
\centering
\small
\caption{Relative Frobenius error (Mean/Median/Std) and constraint violation rates across models and data splits. Mean/Median/Std are reported in \%. Violations are reported as count and rate (\%). Total number of training samples: 751,089; validation samples: 263,188. Best results per metric are highlighted in \textbf{bold}. Testing results are omitted; performance and violation behavior on the test split are extremely similar to corresponding training and validation splits.}
\label{tab:metrics_violations}
\begin{tabular}{l
    c c
    c c
    c c
    c c}
\toprule
\textbf{Metric} 
& \multicolumn{2}{c}{\textbf{\VRNet{}}} 
& \multicolumn{2}{c}{\textbf{Cholesky neural network}~\cite{Xu2021}} 
& \multicolumn{2}{c}{\textbf{Isotropic Vanilla network}}
& \multicolumn{2}{c}{\textbf{Vanilla neural network}} \\
\cmidrule(lr){2-3}\cmidrule(lr){4-5}\cmidrule(lr){6-7}\cmidrule(lr){8-9}
& Training & Validation
& Training & Validation
& Training & Validation
& Training & Validation \\
\midrule
Median (\%) - $\cE_\mathsf{F}(\ull{\ol{C}}, \ullWH{\ol{C}} )$ 
& \bf 1.79 & \bf 1.79 
& 5.35 & 5.33 
& 5.74 & 5.73
& 6.31 & 6.28 \\[2pt]

Mean (\%) - $\cE_\mathsf{F}(\ull{\ol{C}}, \ullWH{\ol{C}} )$   
& \bf 3.61 & \bf 3.59 
& 8.77 & 8.76 
& 10.92 & 10.89
& 12.98 & 12.96 \\[2pt]

Std (\%) - $\cE_\mathsf{F}(\ull{\ol{C}}, \ullWH{\ol{C}} )$   
& \bf 5.50 & \bf 5.49 
& 13.53 & 13.59 
& 18.78 & 18.89
& 26.42 & 26.45 \\[2pt]

\addlinespace[2pt]
\makecell{Voigt bound violations \\ $\ullWH{\ol{C}} \not\preceq \uCvoigt $}
& \makecell{\bf 0\\(0\%)} & \makecell{\bf 0\\(0\%)}
& \makecell{170,578\\(22.7\%)} & \makecell{60,431\\(23.0\%)}
& \makecell{124,104\\(16.52\%)} & \makecell{43,774\\(16.63\%)}
& \makecell{178,365\\(23.8\%)} & \makecell{62,819\\(23.9\%)} \\[2pt]

\makecell{Reuss bound violations \\ $\ullWH{\ol{C}} \not\succeq \uCreuss$}
& \makecell{\bf 0\\(0\%)} & \makecell{\bf 0\\(0\%)}
& \makecell{171,259\\(22.8\%)} & \makecell{60,252\\(22.9\%)}
& \makecell{202,965\\(27.02\%)} & \makecell{71,732\\(27.26\%)}
& \makecell{216,511\\(28.8\%)} & \makecell{76,331\\(29.0\%)} \\[2pt]

\makecell{PD violations \\ $\ullWH{\ol{C}} \not\succeq 0$}
& \makecell{\bf 0\\(0\%)} & \makecell{\bf 0\\(0\%)}
& \makecell{0\\(0\%)} & \makecell{0\\(0\%)}
& \makecell{0\\(0\%)} & \makecell{0\\(0\%)}
& \makecell{26,889\\(3.6\%)} & \makecell{9,443\\(3.6\%)} \\
\bottomrule
\end{tabular}
\end{table}

\section{Application to highly anisotropic and high contrast 2D composites}
\label{sec:2d_mechanics}
\begin{figure}[!htb]
    \centering
    \includegraphics[scale=1.0]{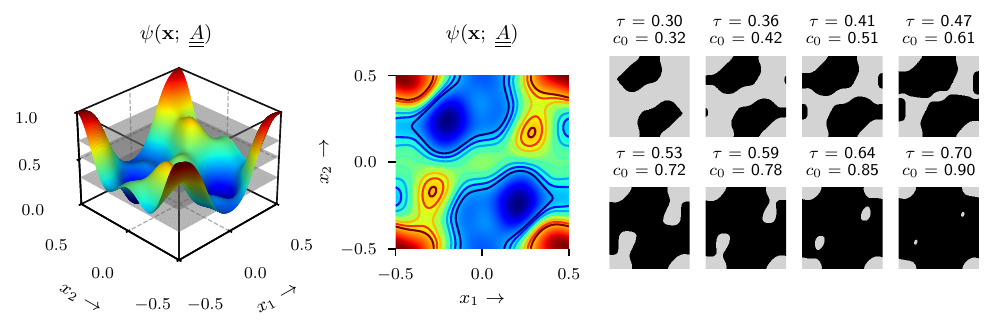}
    \caption{\protect{Schematic of the procedure to generate periodic biphasic unit cells from thresholded trigonometric fields~\cite{Boddapati2024-tb}}}
    \label{fig:threshold_explainer}
\end{figure}

In this section, we describe the application of the \VRNet{} to a 2D mechanical problem, specifically the prediction of the effective elasticity tensor $\ull{\ol{C}}\in Sym_+(\ffR^3)$ of periodic, biphasic composites under plane strain.
The data was generated and obtained from a prior study~\cite{Boddapati2024-tb}, which constructed a large database of unit cells~$\Omega=[-0.5, 0.5] \times [-0.5, 0.5]\subset\ffR^2$ and their homogenized elastic tensors, particularly targeting structured materials with extreme elastic anisotropy.
The data are generated with a previously documented procedure to sample large families of periodic, single-scale unit cells composed of two phases that can realize a wide range of anisotropic effective elastic tensors~\cite{Boddapati2023-la}. 
The construction is based on thresholded trigonometric fields with few spatial modes, which (i) guarantees periodicity by design, (ii) offers direct control over geometric symmetry through simple algebraic constraints on amplitudes, and (iii) yields a monotonic handle on the volume fraction over a wide range (1--99\% volume fraction of phase~1; see below for details) through the threshold parameter~$\tau\in[0, 1]$. A schematic of the procedure is shown in~\cref{fig:threshold_explainer}.

\begin{figure}
    \centering
    \includegraphics[scale=1.0]{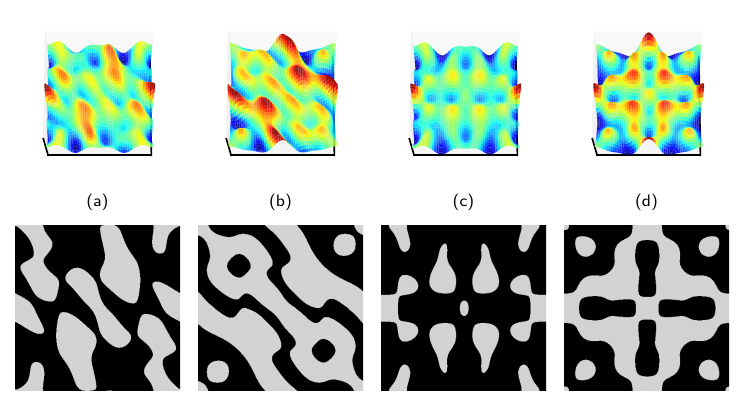}
    \caption{Examples of unit cells with different symmetry classes obtained by imposing simple algebraic constraints on the amplitude matrix $\ull{A}$. (a) No symmetry (random $\ull{A}$); (b) diagonal symmetry ($A_{m,n}=A_{n,m}$); (c) orthotropic symmetry ($A_{m,n}=A_{-m,n}=A_{m,-n}$); (d) square symmetry ($A_{m,n}=A_{n,m}=A_{-m,n}=A_{m,-n}$).}
    \label{fig:symmetry_classes}
\end{figure}

Let $\{\fe_1,\fe_2\}\subset\mathbb{R}^2$ be the orthonormal basis vectors of the 2D unit cell ($\fe_i\cdot \fe_j=\delta_{ij}$). 
For two odd integers $M_1, M_2\in\{1,3,5,\dots\}$ (the numbers of modes in the $\fe_1$ and $\fe_2$ directions, respectively), define the finite index sets
\begin{align}
\cI_1=\Big\{-\tfrac{M_1-1}{2},\ldots,0,\ldots,\tfrac{M_1-1}{2}\Big\}, 
\qquad 
\cI_2=\Big\{-\tfrac{M_2-1}{2},\ldots,0,\ldots,\tfrac{M_2-1}{2}\Big\}.
\end{align}
Given an amplitude matrix $\ull{A}=\big(A_{mn}\big)_{m\in\cI_1,\;n\in\cI_2} \in \ffR^{M_1 \times M_2}$ with entries sampled independently and identically from a uniform distribution $A_{mn}\sim\cU[-1,1]$, we form the periodic field $\levelset(\fx;A)$,
\begin{align}
  \levelset(\fx;A)
  = \sum_{m\in\cI_1}\sum_{n\in\cI_2}
    A_{mn}\;\cos\bigl(2\pi\;(m\;x_1+n\;x_2)\bigr),
  \quad
  \fx \in \Omega = \bigl\{\;x_1\;\fe_1 + x_2\;\fe_2 \;\big|\;
  x_1,x_2 \in \big[-\frac{1}{2},\;\frac{1}{2}\big)\;\bigr\}.
  \label{eq:cosine_field}
\end{align}
We then apply range normalization on $\levelset(\fx;\ull{A})$ such that the rescaled field $\levelsetWT(\fx;\ull{A}) \in [0,1]$,
\begin{align}
  \levelset_\mathrm{min} &= \min_{\fx\in\Omega} \levelset(\fx;\ull{A}), \quad
  \levelset_\mathrm{max} = \max_{\fx\in\Omega} \levelset(\fx;\ull{A}), \quad
  \levelsetWT(\fx;\ull{A}) = \frac{\levelset(\fx;\ull{A}) - \levelset_\mathrm{min}}{\levelset_\mathrm{max} - \levelset_\mathrm{min}} \in [0,1].
  \label{eq:range_rescaling}
\end{align}
For a threshold $\tau\in[0,1]$, the binary indicator~$\chi$ of the microstructure is
\begin{align}
  \chi(\fx;\ull{A},\tau) \;=\; 1_{\{\levelsetWT(\fx;\ull{A})\ge \tau\}},
  \qquad c_1(\tau;\ull{A}) \;=\; \frac{1}{|\Omega|}\int_\Omega \chi(\fx;\ull{A},\tau)\; \mathrm{d}\fx,
  \qquad c_0(\tau;\ull{A}) \;=\; 1 - c_1(\tau;\ull{A}).
  \label{eq:threshold_indicator}
\end{align}
By construction $c_0$ and $c_1$ vary monotonically with $\tau$, that is, $c_0(\tau) \leq c_0(\tau+\delta)$ for $\delta > 0$, and the inverse inequality for $c_1$ holds by construction. 
Symmetry classes of the microstructure can be enforced by simple constraints on $\ull{A}$ (see~\cref{fig:symmetry_classes}).
Increasing $(M_1, M_2)$ introduces finer features and generally reduces extreme anisotropy by distributing the power spectral density over more spatial frequencies~\cite{Boddapati2024-tb}.
Each unit cell $\chi(\fx;\ull{A},\tau)$ represents a biphasic composite.
Phase~0 (stiff - DM 8530 Grey60) and Phase~1 (soft - Tango Black) are modeled as isotropic linear elastic solids with $(E_0,\nu_0)=(1\;\mathrm{GPa},\;0.3), (E_1,\nu_1)=(1\;\mathrm{MPa},\;0.49)$,
as chosen in~\cite{Boddapati2024-tb}. These parameters representative for materials from a commercial multi-material Connex 3D printer. Note the impressive Young's modulus contrast of $10^3$ and a large contrast in bulk/shear response due to the near incompressibility of Tango black due to $\nu_1 \approx 0.5$. 
The data from \cite{Boddapati2024-tb} (2D plane strain homogenized elasticity tensors given in Voigt notation) is used to rewrite the linear stress-strain relation in Mandel notation~(see \cref{sec:notation}):
\begin{equation}
\ul{\ol{\sigma}} = \sV \ol{\sigma}_{11} \\ \ol{\sigma}_{22} \\ \sqrt{2}\; \ol{\sigma}_{12} \eV  = 
\sV
\ol{C}_{1111} &\ol{C}_{1122} & \sqrt{2}\; \ol{C}_{1112} \\
\ol{C}_{2211} &\ol{C}_{2222} & \sqrt{2}\; \ol{C}_{2212} \\
\sqrt{2}\; \ol{C}_{1211} & \sqrt{2}\; \ol{C}_{1222} & 2\; \ol{C}_{1212} \\
\eV : \sV \ol{\varepsilon}_{11} \\ \ol{\varepsilon}_{22} \\ \sqrt{2}\; \ol{\varepsilon}_{12} \eV
= \ol{\ull{C}} : \ul{\ol{\varepsilon}}.
\label{eq:plane_strain_mandel}
\end{equation}
The homogenized elasticity tensor $\ull{\ol{C}}$ is obtained by a standard two-scale first-order computational homogenization, i.e., by solving the corrector problems for three orthonormal macroscopic loadings.
Thermodynamic stability requires $\ol{C}_{1111},\ol{C}_{2222},\ol{C}_{1212}>0$, while off-diagonal couplings $\ol{C}_{1122},\ol{C}_{1112},\ol{C}_{2212}$ may be negative. Symmetry and positive definiteness of the plane strain stiffness $\ol{\ull{C}}$~\cref{eq:plane_strain_mandel} enforces the full set of positivity constraints (see \cref{sec:spectral:normalization}; also \cite{Keshav2025}) automatically.
In the study\cite{Boddapati2024-tb}, a large database was constructed by sampling many amplitude matrices $\ull{A}$ for different values of $M_1, M_2$ and by sweeping the threshold $\tau$; see~\cite{Boddapati2024-tb} for details.
In summary, the dataset construction involves:
\begin{itemize}
  \item \textbf{Mode grids.} Five sets of spatial modes $(M_1\times M_2)\in\cP$, with $\cP=\big\{\;3\times 3,\;5\times 5,\;7\times 7,\;9\times 9,\;11\times 11\;\big\}$.
  \item \textbf{Amplitude sampling.} For each mode set $p\in\cP$, $N_A=2,000$ independent amplitude matrices $\ull{A}^{(p,i)}$ with entries $A_{mn}\sim \cU[-1,1]$ are drawn. Fully random $\ull{A}$ are drawn to encourage low symmetry.
  \item \textbf{Threshold sweep.} For each $\ull{A}^{(p,i)}$, we evaluate $N_\tau=25$ equi-spaced thresholds $\{\tau_k\}_{k=1}^{N_\tau}\subset(0,1)$.
  \item \textbf{Filtering.} In the resulting $2,000\times 25 \time 5 = 250,000$ structures, to avoid trivial geometries (all phase 0; all phase 1), we retain only realizations whose phase volume fraction satisfies
  $c_0(\tau_k;\ull{A}^{(p,i)})\in[0.01,\;0.99]$.
  \item \textbf{Homogenized elasticity tensors.} Unit cells are rasterized at $100\times 100$ pixels and homogenized in plane strain with periodic boundary conditions, using the phase pairs $(E_1,\nu_1)$, $(E_2,\nu_2)$ to obtain $\ull{\ol{C}}$.
\end{itemize}
The data from \cite{Boddapati2024-tb} is used in the subsequent study, with additional simulations being performed using our FFT-based homogenization FANS \cite{FANS_github}.
For the resulting structures, we define a random split (80\% / 20\%) into training and validation sets for each mode set $(M_1, M_2)\in\cP$ (see~\cref{tab:2d_mech_counts}).
For each retained unit cell we store $(\ull{A}^{(p,i)}, \tau_k)$ ($i\in\{1, \dots, 2,000\}; k\in\{1, \dots, 25\}$) and the homogenized elasticity tensor $\ol{\ull{C}}^{(i, k)}_{(M_1, M_2)}$. After removal of inadmissible samples ($c_0<0.01$ or $c_0>0.99$), 219,378 samples remained; see~\cref{tab:2d_mech_counts}. Note that each sample is characterized entirely by the parameter vector $\ul{\mu}=[M_1, M_2, \ull{A}^{(i, k)}_{(M_1, M_2)}, \tau]$.\footnote{Matrices for $M_1=M_2<11$ can be represented by $11\times 11$ matrices padded with zeros, rendering the dimension of $\ul{\mu}$ constant for convenience (see~\cref{eq:center_pad}).}

\begin{table}[!htb]
  \centering
  \caption{\protect Summary of the 2D plane-strain elasticity dataset constructed by \cite{Boddapati2024-tb} from cosine-thresholded unit cells (post-filtering by $c_0\in[0.01,0.99]$). Column headers indicate the mode grid dimensions $(M_1 \times M_2)$ used to generate the amplitude matrix $\ull{A}$.}
  \label{tab:2d_mech_counts}
  \setlength{\tabcolsep}{8pt}
  \begin{tabular}{lccccc|c}
    \toprule
    & \multicolumn{5}{c}{\textbf{Mode grid $(M_1 \times M_2)$}} & \\
    \cmidrule(lr){2-6}
    & $3\times 3$ & $5\times 5$ & $7\times 7$ & $9\times 9$ & $11\times 11$ & \textbf{Total} \\
    \midrule
    \textbf{Train}      & 36,444 & 35,917 & 35,179 & 34,246 & 33,718 & 175,504 \\
    \textbf{Validation} & 9,111 & 8,979 & 8,795 & 8,562 & 8,427 & 43,874 \\
    \midrule
    \textbf{Total}      & 45,555 & 44,896 & 43,974 & 42,808 & 42,145 & 219,378 \\
    \bottomrule
  \end{tabular}
\end{table}

\begin{figure}
    \centering
    \includegraphics[scale=1.0]{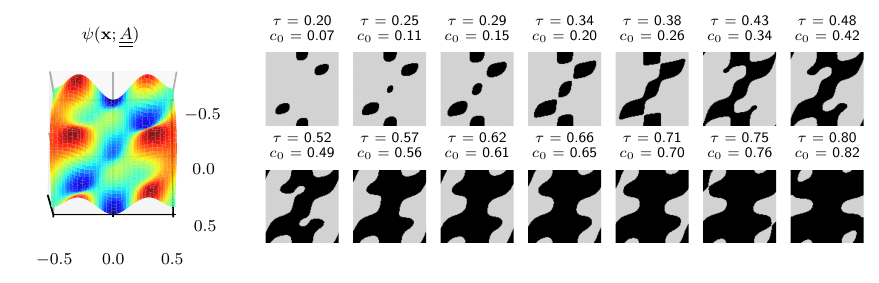}
    \caption{Examples of unit cells generated by thresholded trigonometric fields with a fixed amplitude matrix $\ull{A}$ and increasing threshold $\tau$, showing the monotonic increase in the volume fraction of phase 0 (stiff phase). The microstructures exhibit a wide range of microstructure topologies, demonstrating the challenging nature of the dataset.}
    \label{fig:threshold_sweep_microstructures}
\end{figure}

The prediction of effective elastic properties for periodic composites is fundamentally constrained by the classical G-closure problem. The G-closure describes the set of all possible effective elasticity tensors that can be realized by mixing two given phases in all possible microgeometries, subject to prescribed volume fractions and symmetry constraints. For high-contrast composites, where the elastic moduli of the two phases differ by several orders of magnitude, the G-closure set becomes extremely large and nontrivial. The variability of the microstructures in this dataset is substantial: the use of random trigonometric fields with varying numbers of modes and amplitude matrices generates a vast diversity of geometries, ranging from nearly isotropic to highly anisotropic and from simple to complex topologies. This diversity is further amplified by sweeping the threshold parameter $\tau$, which tunes the volume fraction and can induce topological transitions in the microstructure (see~\cref{fig:threshold_sweep_microstructures}). These factors render the learning problem particularly challenging. First, the mapping from microstructure to effective tensor is highly nonlinear and nonlocal: small changes in geometry can lead to large changes in the effective response, especially near percolation thresholds or in the high-contrast regime. Second, the admissible set of effective tensors (the G-closure) is a highly complicated set of symmetric positive definite matrices, strictly bounded by the upper Voigt and lower Reuss bounds. Third, the presence of extreme anisotropy and high contrast means that the model must generalize across a wide range of physical behaviors, including cases where subtle geometric features dominate the effective response. Consequently, accurately predicting the effective elasticity tensor in this space requires a surrogate model that is highly expressive and physically constrained, motivating the use of spectral normalization and the proposed generalization of the \VRNet{} \cite{Keshav2025} for mechanical problems.

\subsection{Neural surrogate architecture}
\label{subsec:2d_mech_model}

\begin{figure}
  \centering
  \begin{tikzpicture}[remember picture]
    \node[inner sep=0pt] (cnn_trunk) {\resizebox{1.0\textwidth}{!}{\input{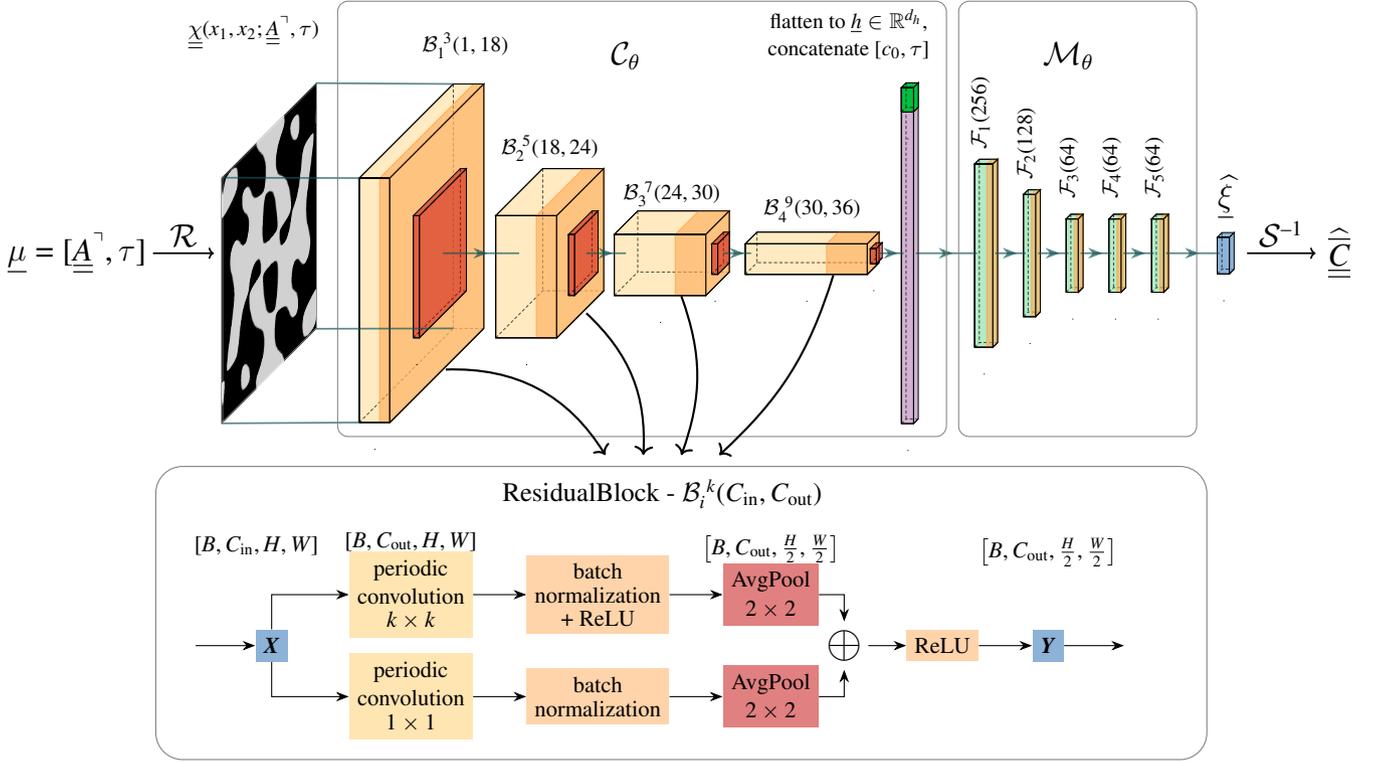}}};
    
    \coordinate (resblock1_point) at ([xshift=-3.1cm, yshift=-1.79cm]cnn_trunk.center);
    \coordinate (resblock2_point) at ([xshift=-1.25cm, yshift=-1.05cm]cnn_trunk.center);
    \coordinate (resblock3_point) at ([xshift=0cm, yshift=-0.825cm]cnn_trunk.center);
    \coordinate (resblock4_point) at ([xshift=2cm, yshift=-0.55cm]cnn_trunk.center);

  \end{tikzpicture}
  \vspace{0.5cm}
  \begin{tikzpicture}[remember picture]
    \node[inner sep=0pt] (resblock) {{\fontsize{8.5}{8.4}\selectfont
\begin{tikzpicture}[
    node distance=10mm, >=Stealth,
    box/.style={inner sep=3pt}
]
\node[fill=\ConvColor, fill opacity=0.5, text opacity=1, box,
            label={above:{$[B,C_{\text{out}},H,W]$}}] (pc) {%
                \shortstack{periodic \\convolution\\ $k \times k$}%
            };
\node[fill=\ConvReluColor, fill opacity=0.5, text opacity=1, box, right=7mm of pc] (bn) {\shortstack{batch \\normalization\\ + ReLU}};
\node[fill=\PoolColor, fill opacity=0.5, text opacity=1, box, right=7mm of bn,
            label={above:{$\big[B,C_{\text{out}},\frac{H}{2},\frac{W}{2}\big]$}}] (pool) { \shortstack{AvgPool\\ $2 \times 2$}};

\node[fill=\ConvColor, fill opacity=0.5, text opacity=1, box, below=2mm of pc] (pc1x1) {\shortstack{periodic \\convolution\\ $1 \times 1$}};
\node[fill=\ConvReluColor, fill opacity=0.5, text opacity=1, box, right=7mm of pc1x1] (bn_sc) {\shortstack{batch \\normalization}};
\node[fill=\PoolColor, fill opacity=0.5, text opacity=1, box, right=7mm of bn_sc] (pool_sc) {\shortstack{AvgPool\\ $2 \times 2$}};

\path (pc.north west) -- (pc1x1.south west) coordinate[pos=0.5] (xin_center);
\node[fill=\DcnvColor, fill opacity=0.75, text opacity=1, box, left=8mm of xin_center,
            label={[xshift=-2mm,yshift=9.75mm]above:{$[B,C_{\text{in}},H,W]$}}] (xin) {$\fX$};

\draw[->] (xin) |- (pc1x1.west);
\draw[->] (pc1x1) -- (bn_sc);
\draw[->] (bn_sc) -- (pool_sc);

\node[circle, draw=none, inner sep=0pt, minimum size=6mm,
            right=7.0cm of xin, font=\Large] (add) {$\oplus$};
\node[fill=\ConvReluColor, fill opacity=0.5, text opacity=1, box, draw=none, right=5mm of add] (outrelu) { ReLU};
\node[fill=\DcnvColor, fill opacity=0.75, text opacity=1, box, draw=none, right=7mm of outrelu, 
    label={[xshift=0mm,yshift=8.25mm]above:{$\big[B,C_{\text{out}},\frac{H}{2},\frac{W}{2}\big]$}}] (out) {$\fY$};

\draw[->] (xin) |- (pc);
\draw[->] (pc) -- (bn);
\draw[->] (bn) -- (pool);
\draw[->] (pool) -| (add);
\draw[->] (pool_sc) -| (add);
\draw[->] (add) -- (outrelu);
\draw[->] (outrelu) -- (out);

\path (add) -- (outrelu) coordinate[pos=1.0] (out_center);
\draw[->] (out) -- ++(1.0,0) node[below, pos=0.85] {};

\draw[<-] (xin) -- ++(-1.0,0) node[below, pos=0.85] {};

\end{tikzpicture}
}};
    \draw[rounded corners=10pt, gray] ([xshift=-0.5cm, yshift=-0.25cm]resblock.south west) rectangle ([xshift=1.0cm, yshift=0.85cm]resblock.north east);
    \node[above=5pt of resblock.north] {ResidualBlock - {$\cB_i^{\;k}(C_{\text{in}}, C_{\text{out}})$}};
    
    \coordinate (detail_point1) at ([xshift=-0.75cm, yshift=1cm]resblock.north);
    \coordinate (detail_point2) at ([xshift=-0.25cm, yshift=1cm]resblock.north);
    \coordinate (detail_point3) at ([xshift=0.25cm, yshift=1cm]resblock.north);
    \coordinate (detail_point4) at ([xshift=0.75cm, yshift=1cm]resblock.north);
    
    \begin{scope}[overlay]
      \draw[->, black, thick, bend left=30] (resblock1_point) to (detail_point1);
      \draw[->, black, thick, bend left=25] (resblock2_point) to (detail_point2);
      \draw[->, black, thick, bend left=20] (resblock3_point) to (detail_point3);
      \draw[->, black, thick, bend left=15] (resblock4_point) to (detail_point4);
    \end{scope}
  \end{tikzpicture}
  \caption{\protect \textbf{End-to-end differentiable surrogate.} The pipeline maps the parametric unit cell specification $(\ull{A}^\urcorner,\tau)$ to the effective plane-strain elasticity tensor $\ol{\ull{C}}$. \textbf{(Top)}: The center-padded amplitude matrix $\ull{A}^\urcorner$ together with threshold $\tau$, renders a soft microstructure image $\ull{\chi}$ via a differentiable renderer $\cR$. The image is processed by a convolutional trunk $\cC_\theta$ with four convolutional residual blocks~\cite{He2015-ws,He2016-bu} $\cB^{k}_i(C_\text{in}, C_\text{out})$. The resulting feature map is flattened to $\ul{h}$ and concatenated with auxiliary scalars $[c_0,\tau]$, then passed to the MLP $\cM_\theta$ with four linear layers to predict normalized spectral parameters $\ulWT{\xi}$. These are spectrally back-transformed to the effective tensor $\ullWH{\ol{C}}$ via the inverse spectral normalization $\cS^{-1}$. \textbf{(Bottom)}: Structure of a single residual block, two parallel convolutional layers with kernel size $k$ and a skipped connection along with batch normalization, ReLU activations, and average pooling.}
  \label{fig:CNN}
\end{figure}

Next, we describe the end-to-end differentiable surrogate used to map the parametric unit cell specification $(\ull{A},\tau)$ to the effective plane-strain elasticity tensor $\ol{\ull{C}}$. 
The model has three stages:
\begin{enumerate}[label=(\roman*)]
  \item A fixed-shape embedding $\cT$ that pads amplitude grids of various mode sets in $\mathcal{P}$ into a common $K\times K$ hyper mode set;
  Let the mode grid of a sample be $\ull{A}\in\mathbb{R}^{m\times n}$, where $m,n\in\{3,5,7,9,11\}$. 
  We define a centered embedding into $\ull{A}^{\urcorner}\in\mathbb{R}^{K\times K}$ with $K$ chosen such that $K \geq \max\{m, n\}$ for all possible $m, n$ (here, $K=11$):
  \begin{align}
    \cT:\ \mathbb{R}^{m\times n}\to\mathbb{R}^{K\times K},\quad
    \ull{A}^{\urcorner}_{ij} \;=\;
    \begin{cases}
      \ull{A}_{\;i-i_0,\;j-j_0}, & i_0\le i< i_0+m,\ \ j_0\le j< j_0+n,\\
      0,& \text{otherwise},
    \end{cases}
    \quad \text{where } 
    \begin{aligned}
      i_0= \frac{(K-m)}{2},\\[2pt]
      j_0= \frac{(K-n)}{2}.
    \end{aligned}
    \label{eq:center_pad}
  \end{align}
  Thus, every sample is represented by a fixed $K\times K$ matrix $\ull{A}^{\urcorner}$, enabling batched GPU training while preserving the spectral layout (modes remain centered).

  \item A differentiable microstructure renderer $\cR$ that synthesizes a soft microstructure from \(\ul{\mu}=(\ull{A}^{\urcorner},\tau)\);
  Given the grid resolution $(N_{x_1},N_{x_2})=(100,100)$, the scalar field is given by,
  \begin{equation}
    \ull{\levelset}(x_1,x_2;\ull{A}^{\urcorner}) \;=\; 
    \sum_{p=-(K-1)/2}^{(K-1)/2}\sum_{q=-(K-1)/2}^{(K-1)/2}
    {A}^{\urcorner}_{pq}\;\cos\big(2\pi(px_1+qx_2)\big), 
    \qquad (x_1,x_2)\in[-\tfrac12,\tfrac12]^2\cap \Pi_{N_{x_1},N_{x_2}},
    \label{eq:renderer_F}
  \end{equation}
  where $\Pi_{N_x,N_y}$ denotes the pixel lattice. We then apply range normalization~\cref{eq:range_rescaling} such that $\ull{\levelsetWT}\in[0,1]$. The scalar field~$\levelsetWT$ is thresholded with a \textit{soft} threshold $\tau$ and temperature \(T>0\),
  \begin{equation}
    \ull{\chi}(x_1,x_2;\ull{A}^{\urcorner},\tau) \;=\; 
    \sigma\left(\frac{\ull{\levelsetWT}(x_1,x_2;\ull{A}^{\urcorner})-\tau}{T}\right),
    \qquad \sigma(t)=\frac{1}{1+e^{-t}}.
    \label{eq:soft_threshold}
  \end{equation}
  By construction, the map $\cR(\ull{A}^{\urcorner},\tau)\mapsto \ull{\chi}$ is smooth; therefore, the gradients can flow from the loss to both \(\ull{A}^{\urcorner}\) and \(\tau\) (allowing inverse design later). 
  We use $T=0.0001$ to approximate a hard threshold while maintaining numerical stability and some degree of differentiability.

  \item A regressor predicting the normalized degrees of freedom $\WH{\ul{\xi}}$ (see~\cref{sec:spectral:normalization}).
  Let the rendered microstructure be a monochrome image $\ull{\chi}\in[0,1]^{N_{x_1}\times N_{x_2}}$   and let $\ul{s}=[c_0,\tau]^\mathsf{T}\in[0,1]^2$ be an auxiliary feature vector, where $c_0$ is the phase volume fraction of phase 0 in the rendered image $\ull{\chi}$ and $\tau$ is the threshold used.
  A convolutional trunk $\cC_\theta$ extracts an image descriptor $\ul{h}=\cC_\theta(\ull{\chi})\in\mathbb{R}^{d_h}$; see \cref{fig:CNN}.
  A multilayer perceptron (MLP) with a sigmoid head $\cM_\theta$ then regresses the normalized target from the concatenation $[\ul{h};\ul{s}]$, such that $\ul{\WH{\xi}}= \cM_\theta\big([\ul{h};\ul{s}]\big)=[\WH{\ul{\xi}}_\lambda, \WH{\ul{\xi}}_{\mathrm{q}}]^\mathsf{T}$.
  Here, \(\ul{\xi}_\lambda\in[0,1]^3\) are the normalized eigenvalues and \(\ul{\xi}_{\mathrm{q}}\in[0,1]^3\) parameterize an orthogonal matrix; both are produced with sigmoid activations to confine them to the unit cube.
  This is followed by the unique and deterministic spectral inverse transform $\cS^{-1}$ to $\WH{\ol{\ull{C}}}$.
\end{enumerate}

The final surrogate can be written as the composition,
\begin{align}
  \cM_\theta \;\circ\; \cC_\theta \;\circ\; \cR \;\circ\; \cT:\ (\ull{A},\tau) \mapsto \WH{\ul{\xi}} \in [0,1]^6,
\end{align}
where subscript $(\bullet)_\theta$ indicates a dependence on learnable parameters.
This construction ensures automatic feature detection from microstructure images, as well as end-to-end differentiability in $(\ull{A},\tau)$ for inverse design. Note that by replacing $\cR \circ \cT(\ull{A}, \tau)$ with any other image source, it is equally possible to generalize this approach to arbitrary microstructure sources. 
This construction can be straightforwardly extended to 3D by embedding 3D amplitude grids, rendering volumetric microstructures via trigonometric fields, and applying a 3D convolutional regressor, enabling end-to-end differentiable surrogates for 3D systems. However, the computational requirement for the 3D setting will increase considerably compared to the 2D setting.

An alternative would be to bypass the image representation and learn a regressor directly from $(c_0,\tau,\ull{A}^\urcorner)$. 
We deliberately adopt an image-based surrogate for two reasons.
First, convolutional encoders provide a strong inductive bias for periodic microstructures by exploiting locality and translation invariance. This typically improves sample efficiency compared to learning from $\ull{A}^\urcorner$ whose entries lack spatial semantics. 
Second, the image representation is microstructure generator agnostic, i.e., once trained, the convolutional surrogate can be applied to periodic microstructures from different generators or from voxelized experimental data, whereas a descriptor-only surrogate is intrinsically tied to the specific parameterization, e.g., $(\ull{A}^\urcorner,\tau)$ for now.
Importantly, rendering the microstructure does not increase the number of trainable weights (the generator itself is deterministic and parameter-free, unlike a decoder-type architecture), and it only changes the input representation. 
To explicitly quantify the difference in performance, we also train a descriptor-based variant of the \VRNet{} that maps directly from $(c_0,\tau,\ull{A}^\urcorner)$ to the normalized spectral degrees of freedom and compare its performance to the image-based counterpart in~\cref{tab:metrics_violations_2d}.

\subsection{Numerical experiments and results}
\label{subsec:2d_mech_results}

The convolutional trunk $\cC_\theta$ is a deep residual CNN with periodic convolutions and four stages. Each stage consists of a ResidualBlock with two periodic convolutional layers with increasing kernel sizes ($k_1=3$, $k_2=5$, $k_3=7$, $k_4=9$), batch normalization, and ReLU activations, followed by $2 \times 2$ average pooling. The number of output channels $C_\mathrm{out}$ increases as $18\to24\to30\to36$ across the four blocks, which implies $C_\mathrm{in}=1 \to 18 \to 24 \to 30$. After the final block, the feature map is flattened and concatenated with the auxiliary scalars $[c_0, \tau]$. This is passed to a multi-layer perceptron (MLP) with five hidden layers of sizes 256, 128, 64, 64, and 64, each with batch normalization and a mixed activation (\texttt{[SELU, Tanh, Sigmoid, Identity]}), and a final sigmoid-activated output layer of dimension 6. The model uses periodic convolutions throughout to preserve microstructure periodicity. Training uses the \texttt{AdamW} optimizer with an initial learning rate of $10^{-1}$, and a \texttt{ReduceLROnPlateau} scheduler (factor 0.5, patience 50 epochs).
The loss function is defined as the mean Voigt-Reuss normalized error $\phi$ (\cref{eq:rel:error}) over each batch,
\begin{align}
  \cL = \frac{1}{N_\text{batch}} \sum_{i=1}^{N_\text{batch}} \phi( \ullWT{{C}}^{(i)}, \ullWTH{{C}}^{(i)}),
\end{align}
where $N_\text{batch}$ is the batch size, $\ullWT{{C}}^{(i)}$ is the true normalized stiffness tensor, and $\ullWTH{{C}}^{(i)}$ is the predicted tensor for sample $i$ in the batch.

\subsubsection{Microstructure to property forecasting}

\begin{figure}[!htb]
    \centering
    \includegraphics[scale=0.925]{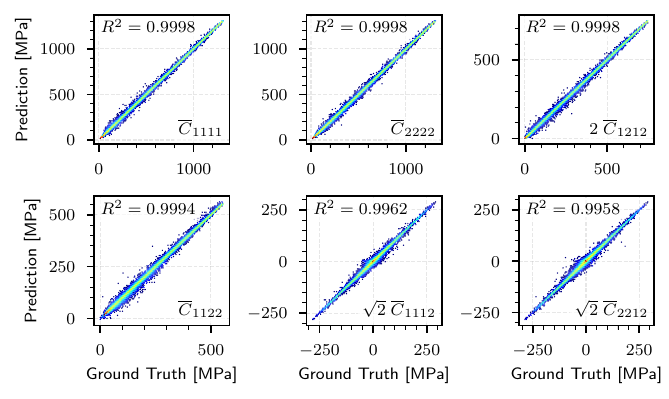}
    \includegraphics[scale=0.925]{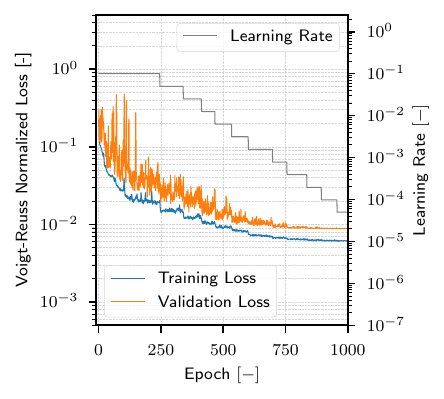}
    \caption{2D plane strain elasticity. \textbf{Left:} Ground truth vs. prediction for the six independent components of $\ol{\ull{C}}$.
    \textbf{Right:} The Voigt-Reuss normalized loss decays and saturates; the learning rate decreases with training via a ReduceLROnPlateau scheduler.}
    \label{fig:mech2d_spinodoid_vrnn_predictions}
\end{figure}

\Cref{fig:mech2d_spinodoid_vrnn_predictions} (left) compiles ground truth versus predicted entries of homogenized plane-strain stiffness $\ol{\ull{C}}$ for all samples. 
The six components tightly cluster on the identity line with a coefficient of determination $R^2>0.99$, indicating (i) negligible bias across magnitudes and signs (including shear-normal couplings $\ol{C}_{1112},\ol{C}_{2212}$), and (ii) uniformly low variance of the residuals. 
Notably, the spectral back-transform ensures that every predicted $\widehat{\ull{\ol{C}}}$ is symmetric positive definite and lies spectrally within Voigt-Reuss bounds; hence, even the few visible outliers remain physically admissible by construction.
The right plot in~\cref{fig:mech2d_spinodoid_vrnn_predictions} reports the loss of training and validation of the model.
Because $\phi$ for each sample is intrinsically bound to $[0,1]$, the training and validation losses provide a relative scale of progress.
Training converges smoothly with a small and stable generalization gap to subpercent errors relative to the admissible spectral range: $\cL_\text{train}\approx0.586 \%$ and $\cL_\text{val}\approx0.824 \%$ at saturation. In particular, the validation loss shows the same qualitative and nearly the same quantitative behavior as the training loss.
The scheduler lowers the learning rate from $10^{-1}$ to $5\times10^{-5}$, coincident with the flattening of both curves, which signals that the model has reached its accuracy floor under current capacity and data.
Crucially, the bounded, well-conditioned target space (all outputs in $[0,1]$) produces well-behaved gradients and loss descent without instability, a hallmark advantage of spectral normalization over direct regression in physical units. We also attribute the smooth training to the incorporation of the identity activation in our multi-activation layers, which acts as a sort of bypass inspired by the ResNet architecture.

\begin{figure}[!htb]
    \centering
    \includegraphics[scale=1]{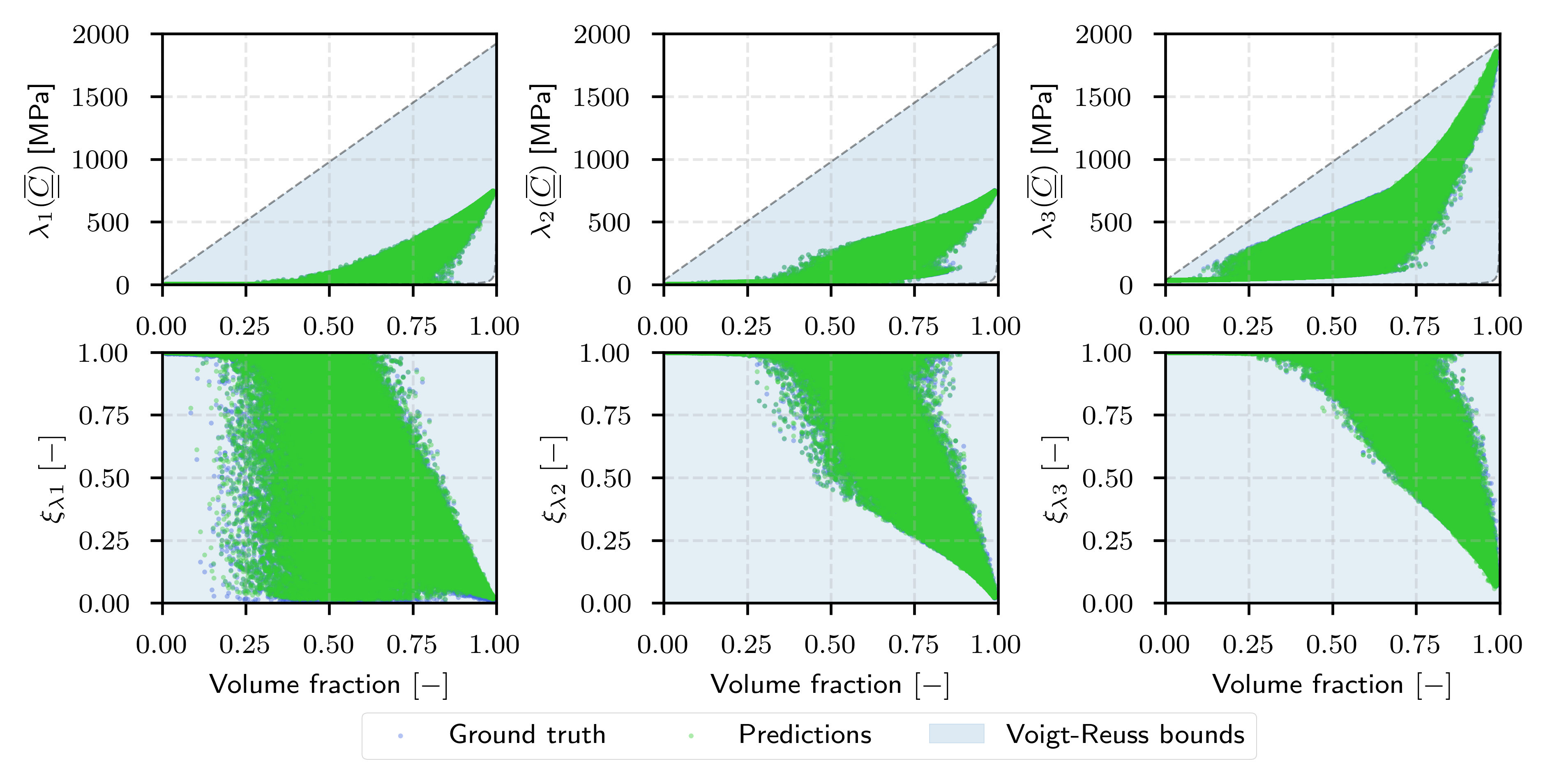}
    \caption{2D plane strain elasticity (validation set). 
    Distribution of eigenvalues $\{\lambda_i(\ol{\ull{C}})\}_{i=1}^3$ and the normalized eigenvalues $\{{\xi}_{\lambda_i}\}_{i=1}^3\in[0,1]$ with respect to the volume fraction.
    }
    \label{fig:mech2d_spinodoid_vrnn_eigenvalues}
\end{figure}

In~\cref{fig:mech2d_spinodoid_vrnn_eigenvalues} (top row) plots the three eigenvalues $\{\lambda_i(\ol{\ull{C}})\}_{i=1}^3$ of the effective stiffness tensor as a function of the volume fraction for all validation samples.
The shaded area indicates the Voigt-Reuss bounds, which collapse at the endpoints $c_0\to 0$ and $c_0 \to 1$ and widen near $c_0\approx 0.5$, where microstructural variability and anisotropy peak.
The ground truth eigenvalues (blue markers) span a wide range from $0$ to $2,000$ MPa, illustrating the extreme variability and high phase contrast of the dataset.
The predicted eigenvalues (green markers) tightly overlap with the ground truth across the entire range, demonstrating the model's ability to capture the effective stiffness even in challenging regimes accurately.
The bottom row shows the same eigenvalues after spectral normalization, i.e., $\{{\xi}_{\lambda_i}\}_{i=1}^3\in[0,1]$.
The normalization compresses the wide variation in physical units down to a bounded unit interval while preserving the ordering and relative positions of the eigenvalues.
The near-perfect overlap between ground truth and predictions in this normalized space illustrates why training in the bounded domain is numerically well-conditioned and stable.

To explicitly assess the benefits of spectral normalization, we compare four surrogates trained on the same datasets.
\begin{itemize}
  \item \textbf{Convolutional \VRNet{}}: The model described in~\cref{fig:CNN}, which predicts normalized spectral parameters and reconstructs $\ullWH{\ol{C}}$ by the spectral back-transform, thus guaranteeing symmetry, positive definiteness, and strict Voigt-Reuss admissibility.
  \item \textbf{Convolutional Cholesky neural network}~\cite{Xu2021}: A model with the same architecture (\cref{fig:CNN}), which predicts a Cholesky factor $\ullWH{L}$ such that $\ullWH{\ol{C}} = \ullWH{L} \; \ullWH{L}^\mathsf{T}$. This guarantees positive definiteness but does not enforce two-sided Voigt-Reuss bounds.
  \item \textbf{Convolutional vanilla network}: A model with the same architecture (\cref{fig:CNN}), which directly regresses the six independent stiffness components of $\ol{\ull{C}}$ in physical units without any constraint enforcement.
  \item \textbf{Descriptor-based \VRNet{}}: A model that bypasses microstructure rendering and the convolutional trunk, and instead maps the 123-dimensional descriptor vector $(c_0,\tau,\mathrm{vec}(\ull{A}^\urcorner))$ to the normalized spectral degrees of freedom using a fully connected network with hidden widths $[1024,\,512,\,256,\,128,\,128]$ each with batch normalization and mixed activations.
\end{itemize}

\begin{figure}[!htb]
    \centering
    \includegraphics[scale=1]{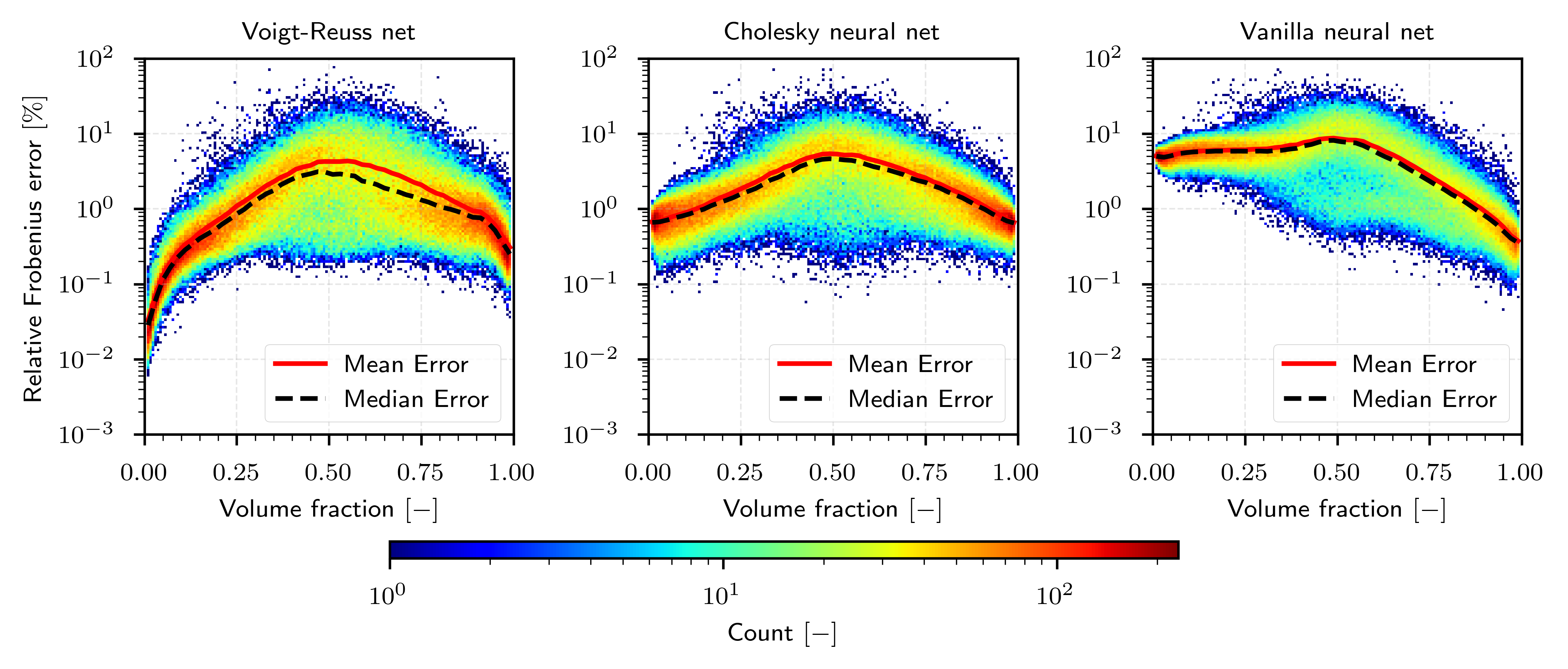}
    \caption{ 2D histogram of the relative Frobenius error $\cE_\mathsf{F}(\ol{\ull{C}}, \widehat{\ol{\ull{C}}})$ across volume fractions for different models on the validation set. The intensity indicates the number of samples in each 2D bin, with solid and dashed lines showing the mean and median errors, respectively.
    }
    \label{fig:mech2d_spinodoid_all_models_errors}
\end{figure}

\begin{table}[!htb]
\centering
\small
\caption{Relative error (Mean/Median/Std) and constraint violation rates across models and data splits for the 2D mechanical problem. Mean/Median/Std are reported in \%. Violations are reported as count and rate (\%). Total number of training samples: 175,494; validation samples: 43,874. Best results per metric are highlighted in \textbf{bold}.}
\label{tab:metrics_violations_2d}
\begin{tabular}{l
    c c
    c c
    c c
    c c}
\toprule
\textbf{Metric} 
& \multicolumn{2}{c}{\makecell{\bf (Image-based) \\ \bf \VRNet{}} }
& \multicolumn{2}{c}{\makecell{\bf (Image-based) \\ \bf Cholesky neural network}~\cite{Xu2021}} 
& \multicolumn{2}{c}{\makecell{\bf (Image-based) \\ \bf Vanilla neural network}} 
& \multicolumn{2}{c}{\makecell{\bf (Descriptor-based) \\ \bf \VRNet{}}} \\
\cmidrule(lr){2-3}\cmidrule(lr){4-5}\cmidrule(lr){6-7}\cmidrule(lr){8-9}
& Training & Validation
& Training & Validation
& Training & Validation
& Training & Validation \\
\midrule
Median (\%) - $\cE_\mathsf{F}(\ull{\ol{C}}, \ullWH{\ol{C}} )$
& \bf 0.77 & \bf 0.89
& 1.37 & 1.43
& 4.48 & 4.61
& 2.06 & 2.60 \\[2pt]

Mean (\%) - $\cE_\mathsf{F}(\ull{\ol{C}}, \ullWH{\ol{C}} )$    
& \bf 1.59 & \bf 2.41
& 2.30 & 2.78
& 4.73 & 5.04
& 4.56 & 8.16 \\[2pt]

Std (\%) - $\cE_\mathsf{F}(\ull{\ol{C}}, \ullWH{\ol{C}} )$    
& \bf 2.23 & \bf 3.93
& 2.46 & 3.60
& 3.99 & 4.50
& 6.60 & 16.31 \\[2pt]

\addlinespace[2pt]
\makecell{Voigt bound violations \\ $\ullWH{\ol{C}} \not\preceq \uCvoigt $}
& \makecell{\bf 0\\(0\%)} & \makecell{\bf 0\\(0\%)}
& \makecell{32\\(0.02\%)} & \makecell{6\\(0.01\%)}
& \makecell{7\\(0\%)} & \makecell{0\\(0\%)}
& \makecell{ 0\\(0\%)} & \makecell{ 0\\(0\%)} \\[2pt]

\makecell{Reuss bound violations \\ $\ullWH{\ol{C}} \not\succeq \uCreuss$}
& \makecell{\bf 0\\(0\%)} & \makecell{\bf 0\\(0\%)}
& \makecell{41,665\\(23.74\%)} & \makecell{10,542\\(24.03\%)}
& \makecell{77,921\\(44.40\%)} & \makecell{19,562\\(44.59\%)}
& \makecell{ 0\\(0\%)} & \makecell{ 0\\(0\%)} \\[2pt]

\makecell{PD violations \\ $\ullWH{\ol{C}} \not\succeq 0$}
& \makecell{\bf 0\\(0\%)} & \makecell{\bf 0\\(0\%)}
& \makecell{0\\(0\%)} & \makecell{0\\(0\%)}
& \makecell{48,937\\(27.89\%)} & \makecell{12,176\\(27.75\%)}
& \makecell{ 0\\(0\%)} & \makecell{ 0\\(0\%)} \\
\bottomrule
\end{tabular}
\end{table}

\begin{figure}[!htb]
    \centering
    \includegraphics[width=\textwidth]{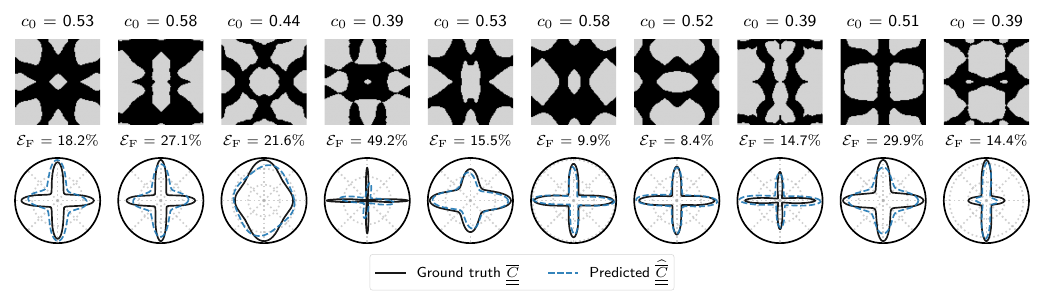}
    \caption{Out-of-distribution image-to-elasticity inference. 
    \textbf{Top:} Ten diffusion-generated binary microstructures taken from~\cite{Bastek2023-qt, Bastek2023-pp} (family disjoint from training). 
    \textbf{Bottom:} Young's modulus polar surfaces $E(\theta)$ from ground-truth homogenization (solid) and direct \VRNet{} predictions from the image (dashed).}    \label{fig:mech2d_ood_direct}
\end{figure}

\Cref{tab:metrics_violations_2d} shows that the image-based \VRNet{} attains the lowest median/mean/standard deviation errors while exhibiting zero violations of Voigt, Reuss, or positive definiteness across datasets.
In contrast, the image-based Cholesky network enforces positive definiteness by construction but still violates the Reuss bound for roughly one quarter of all samples, highlighting that SPD alone is insufficient to guarantee physically admissible homogenized stiffness tensors.
The image-based vanilla regressor performs worst and frequently produces inadmissible predictions, including both Voigt/Reuss violations and non-positive definite outputs.
Finally, the descriptor-based \VRNet{} confirms that direct use of $(\ull{A}^\urcorner,\tau)$ is feasible and remains strictly admissible due to the spectral back-transform, but its accuracy is inferior to the image-based \VRNet{} despite even higher parameter count of the fully connected surrogate.
This indicates that learning morphology-sensitive features from periodic microstructure images provides a beneficial inductive bias and improved sample efficiency, while also preserving the advantage that the same surrogate can be transferred to different microstructure sources where the generator parameters $(\ull{A}^\urcorner,\tau)$ are not available (see~\cref{fig:mech2d_ood_direct}).

\Cref{fig:mech2d_spinodoid_all_models_errors} presents a 2D histogram of the relative Frobenius error in the predicted stiffness tensor and its prediction as a function of volume fraction for all convolutional models on the validation set.
The intensity indicates the number of samples in each 2D bin, with solid and dashed lines showing the mean and median errors, respectively.
As expected, \VRNet{} errors are smallest near $c_0\to 0$ and $c_0\to 1$, where the spectral gap collapses, and largest near $c_0\approx 0.5$.
Nevertheless, the \VRNet{} mean, and median errors remain very modest (around 3\% and 2\%, respectively at $c_0\approx 0.5$), and even the rare worst-case errors (up to around 65\%) are still spectrally admissible by construction, i.e., they lie within the bounds.

To stress-test generalization, we consider ten binary microstructures sampled from an independent diffusion-based image generator (completely disjoint from the cosine-threshold family used for training).
We can evaluate image-only inference by feeding a binary microstructure image $\ull{\chi}\in\{0,1\}^{N_{x_1}\times N_{x_2}}$ directly into the convolutional trunk and regressor, followed by the spectral back-transform.
The predictor is
\begin{align}
  \ullWH{\ol{C}}
  \;=\;
  \cS^{-1}\left(
    \cM_\theta\big(
      \cC_\theta(\ull{\chi})
    \big)
  \right),
  \label{eq:direct_image_inference}
\end{align}
where $\cC_\theta$ is the periodic convolutional trunk, $\cM_\theta$ the MLP head producing the normalized DOFs $\ulWH{\xi}$, and $\cS$ the spectral normalization.
This path requires a single forward pass.
\Cref{fig:mech2d_ood_direct} shows ten diffusion-generated test images (top) and the corresponding plane-strain Young's modulus surfaces $E(\theta)$ (bottom), comparing ground truth (solid) versus direct \VRNet{} predictions (dashed). 
Despite pronounced out-of-distribution features (e.g., very slender beams, sharp junctions), the predicted $E(\theta)$ curves closely track the ground truth in both shape and level at most angles. 
The relative Frobenius errors at the tensor level are modestly higher than on the in-distribution data, consistent with the geometric gap between these images and the training manifold. However, all predictions remain strictly admissible by construction. 
Practically, this enables the instant screening of arbitrary microstructure images with a single model evaluation, while preserving physical guarantees. Additionally, when new input data becomes available, the existing model appears to be a good candidate for transfer learning, as it can be refined at a much lower cost than when starting from a random initialization.

\begin{figure}[!htb]
    \centering
    \vspace{1cm}
    \includegraphics[scale=1]{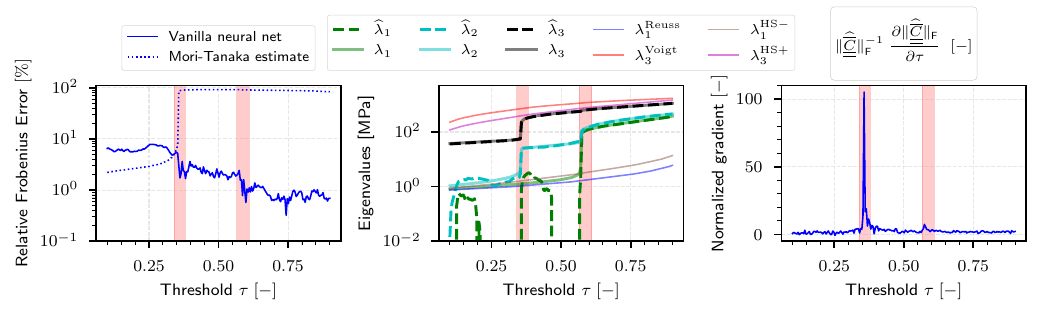} \\[0.7cm]
    \begin{tikzpicture}
        \node[anchor=south west,inner sep=0] (image) at (0,0) {\includegraphics[scale=1]{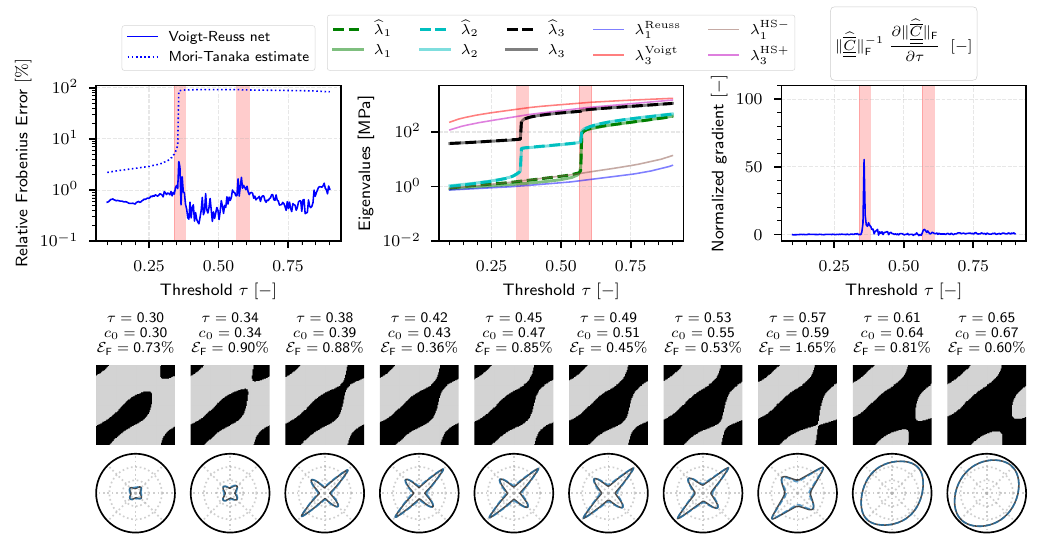}};
        \draw[red, thick] (3.1,0.15) rectangle ++(3.2,3.9);
        \draw[red, thin, fill=red, opacity=0.25] (4.0,2.55) rectangle ++(0.43,0.43);
        \draw[red, thin, fill=red, opacity=0.25] (5.6,2.55) rectangle ++(0.43,0.43);
        \draw[red, thick] (12.7,0.15) rectangle ++(3.2,3.9);
        \draw[red, thin, fill=red, opacity=0.25] (13.65,1.9) rectangle ++(0.43,0.43);
        \draw[red, thin, fill=red, opacity=0.25] (15.25,1.9) rectangle ++(0.43,0.43);
    \end{tikzpicture}
    \begin{tikzpicture}[remember picture, overlay]
      \draw[thick, rounded corners=10pt, gray!50] (-17.7,10.0) rectangle (0.0,15.8);
      \draw[thick, rounded corners=10pt, gray!50] (-17.7,0) rectangle (0.0,9.8);
      \node[fill=white, text width=4.0cm, align=center, rotate=0, rounded corners] at (-8.85,15.4) {Vanilla neural network};
      \node[fill=white, text width=4.0cm, align=center, rotate=0, rounded corners] at (-8.85,9.4) {\VRNet{}};
    \end{tikzpicture}
    \caption{Sensitivity to the threshold parameter $\tau$ for a fixed amplitude matrix $\ull{A}$.
    \textbf{Top row (vanilla NN):} Relative Frobenius error $\cE_\mathsf{F}(\ol{\ull{C}},\widehat{\ol{\ull{C}}})$ (solid) with Mori--Tanaka reference (dashed), eigenvalues $\{\lambda_i(\ol{\ull{C}})\}_{i=1}^3$ with Voigt/Reuss and Hashin--Shtrikman bounds (HS$\pm$), and the normalized gradient magnitude $\|\partial_\tau \widehat{\ol{\ull{C}}}\|_\mathsf{F}$.
    \textbf{Middle row (\VRNet{}):} Analogous diagnostics for \VRNet{}.
    \textbf{Bottom:} Ten frames along the $\tau$-path; red boxes mark the two topology changes (island merger/percolation) coinciding with error and gradient spikes, together with Young's modulus surfaces (log scale) of the ground truth $\ol{\ull{C}}$ (solid) and the \VRNet{} prediction $\widehat{\ol{\ull{C}}}$ (dashed).}
    \label{fig:mech2d_threshold_sweep}
\end{figure}
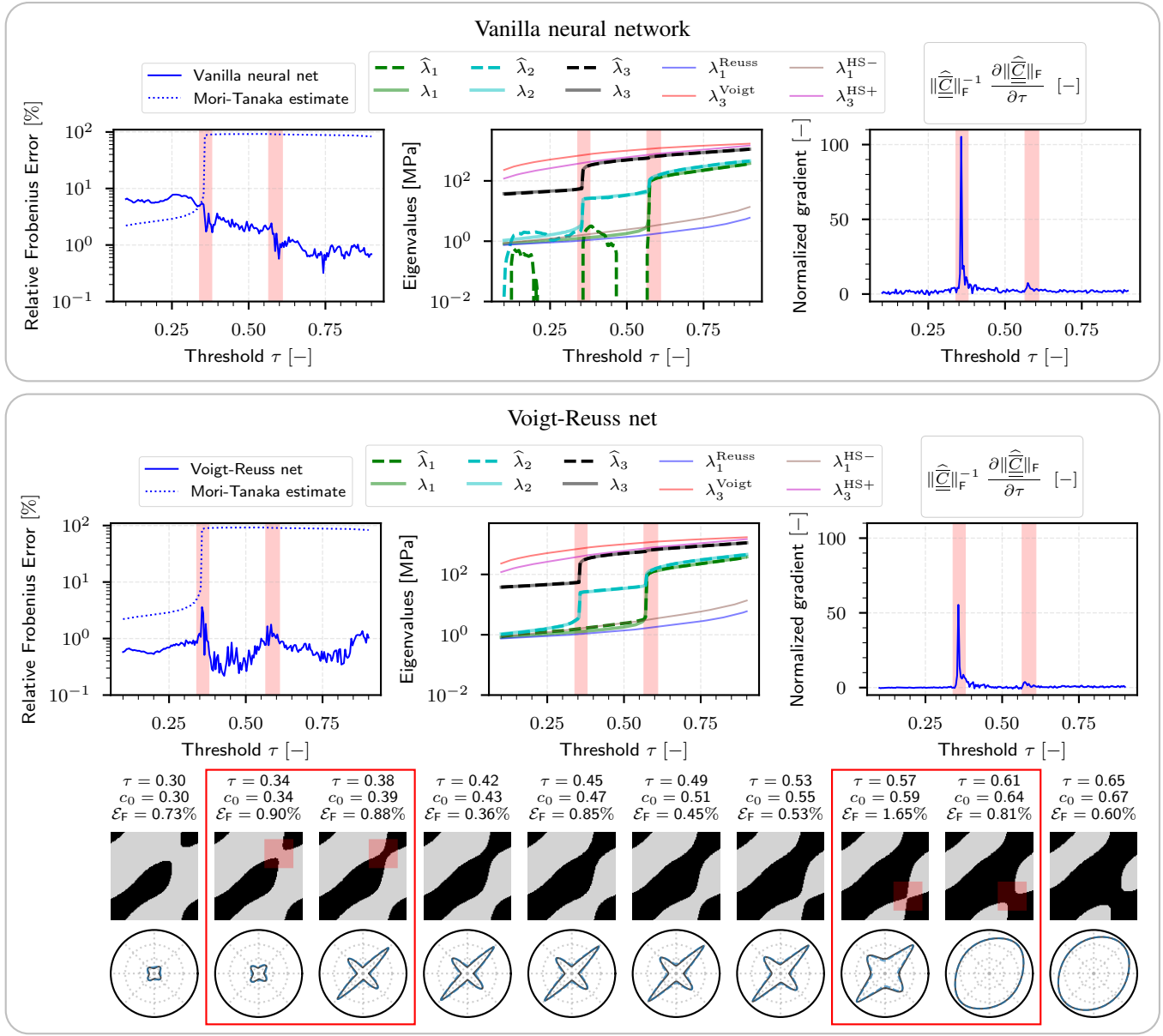

In order to systematically investigate the sensitivity of the surrogate to the threshold parameter $\tau$, we fix a random amplitude matrix $\ull{A}$ and sweep $\tau$ over 100 values from 0 to 1, rendering $\ull{\chi}(\;\cdot\;;\ull{A},\tau)$ and computing both the homogenized tensor $\ol{\ull{C}}$ and the surrogate predictions $\widehat{\ol{\ull{C}}}$.
We compare two surrogates, the vanilla surrogate and the \VRNet{} and include a classical Mori-Tanaka estimate (volume fraction with spherical Eshelby tensor) as a geometry-agnostic baseline.

\Cref{fig:mech2d_threshold_sweep} shows that \VRNet{} maintains sub-percent relative Frobenius errors for almost the entire $\tau$-path, with two pronounced spikes.
In contrast, the Mori-Tanaka estimate begins reasonably at low fractions but degrades sharply as $\tau$ increases: once the inclusion topology ceases to resemble well-separated, non-interacting particles (assumptions underlying Mori-Tanaka estimates), its error grows and eventually becomes unacceptable.
This behavior is expected in high-contrast media where connectivity and shape effects dominate the effective stiffness.
The two error peaks of the \VRNet{} align exactly with abrupt topology changes: previously disconnected islands coalesce into percolating channels (see the bottom strip of microstructures). 
These events are level-set percolation transitions of the field $\levelsetWT(\fx; \ull{A})$ under thresholding; they trigger discontinuous jumps in transport paths and load-bearing skeletons, hence non-smooth changes in $\ol{\ull{C}}$. 
The eigenvalue plot makes this explicit: the eigenvalues $\{\lambda_i(\ol{\ull{C}})\}_{i=1}^3$ show two clear jumps, and the \VRNet{} tracks both jumps essentially on target.
The Hashin-Shtrikman (HS$\pm$) and the Voigt-Reuss envelopes, which depend only on phase moduli and volume fraction, provide a geometry-agnostic benchmark; the ground truth and \VRNet{} eigenvalues remain within these envelopes as well.
The fact that the \VRNet{} resolves the spectral jumps is notable: The model only sees the amplitude matrix $\ull{A}$ and the threshold $\tau$, and consequently the rendered microstructure images, and learns the global response, including percolation-driven anisotropy while retaining strict physical admissibility. 
In contrast, the vanilla surrogate exhibits substantially higher errors along the same path and completely fails to predict the smallest and second-smallest eigenvalues reliably, and the predicted eigenvalues drop below the Reuss bound, reflecting a loss of physical admissibility.

To further probe sensitivity, differentiating the Frobenius norm of the predictions with respect to $\tau$ reveals that both surrogates produce informative gradient spikes precisely at the two critical thresholds.
These spikes signal non-smooth property changes induced by global connectivity flips, which are well-known triggers for abrupt property changes that are beyond the reach of first-order mean-field closures, such as Mori-Tanaka or self-consistent schemes.

Overall, \VRNet{} accurately captures the geometry-controlled locally non-smooth dependence of the effective stiffness on the threshold, including (multiple) percolation-induced spectral jumps, while remaining strictly admissible. In particular, the number of such spectral jumps varies among the microstructures studied; that is, it is not explicitly considered in the trained \VRNet{}, but rather implicitly through the data. Contrastingly, vanilla regression violates bounds and misrepresents the most sensitive eigenmodes.

The differentiable chain $(\ull{A},\tau)\to\widehat{\ol{\ull{C}}}$ of the \VRNet{} provides reliable gradients, except for genuine topological events, which appear as informative spikes that could be exploited in inverse design (e.g., to seek or avoid connectivity transitions).

\subsubsection{Inverse materials design}

\begin{figure}[!htb]
    \centering
    \includegraphics[scale=0.9]{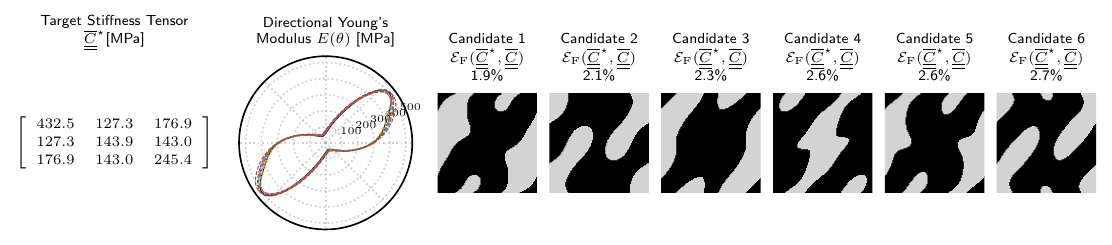}\vspace{-0.24cm}
    \includegraphics[scale=0.9]{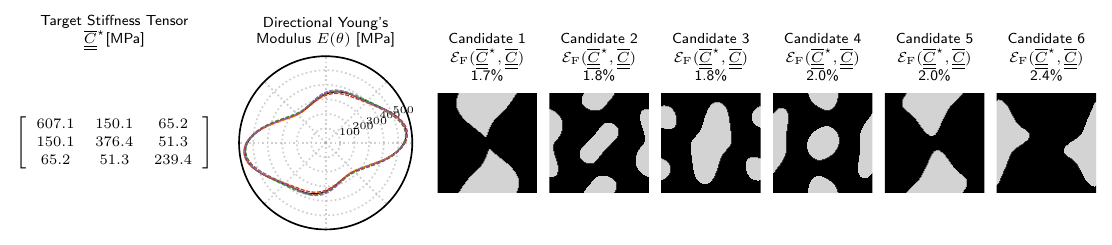}\vspace{-0.24cm}
    \includegraphics[scale=0.9]{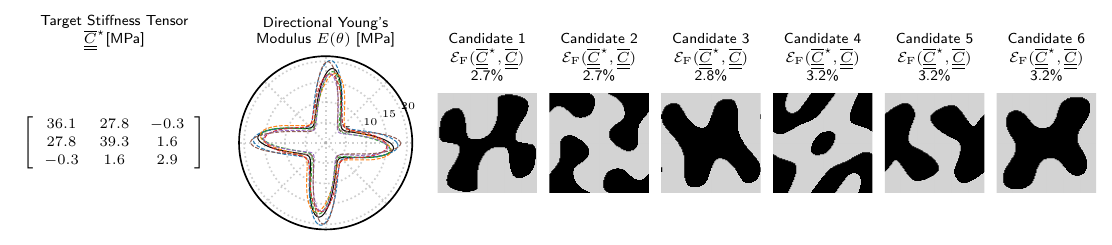}\vspace{-0.24cm}
    \includegraphics[scale=0.9]{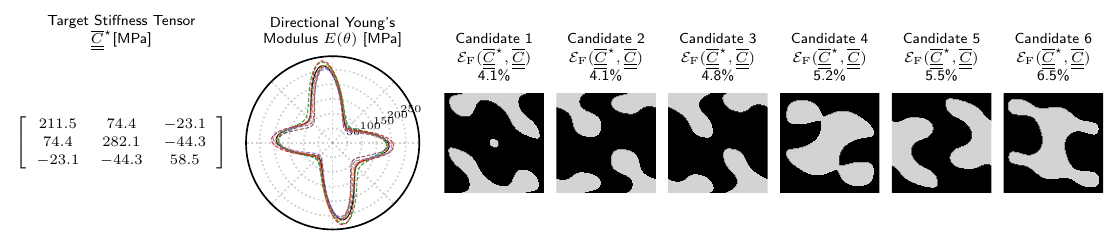}
    \caption{Inverse design to prescribed effective stiffnesses in plane strain. 
    For four distinct targets (rows), the first column lists the target tensor $\ol{\ull{C}}^\star$; the second overlays the Young's modulus polar surface $E(\theta)$ of the target (black) with the best six candidates (colored); the next six columns show the corresponding candidate microstructures with their relative Frobenius error $\cE_\mathsf{F}(\ull{\ol{C}}^{\star}, \ull{\ol{C}})$ in~\%. 
    Across all cases, the recovered $E(\theta)$ nearly overlaps the target, and the tensor errors are only a few percent.}
    \label{fig:mech2d_inverse_to_C}
\end{figure}

We finally demonstrate the application of the \VRNet{} model for the inverse design of microstructures to achieve prescribed effective plane-strain stiffness tensors $\ol{\ull{C}}^\star$.
Given a physically admissible target plane-strain stiffness $\ol{\ull{C}}^\star\in\mathrm{Sym}^+(\ffR^3)$, we seek unit cell parameters $(\ull{A},\tau)$ whose predicted effective tensor is as close to the target as possible. 
Let the differentiable \VRNet{} surrogate (\cref{subsec:2d_mech_model}) be written compactly as 
\begin{align}
\ullWH{\ol{C}}\;(\ull{A},\tau)
= \cS^{-1}\left( \cM_\theta\big( \cC_\theta(\cR(\cT(\ull{A}),\tau))\big) \right),
\end{align}
where $\cS$ denotes the spectral normalization. 
We minimize the (scale-free) distance
\begin{align}
  \cJ(\ullWH{\ol{C}}) = \min_{\ull{A},\;\tau\in(0,1)} 
  \cE_\mathsf{F}\big(\ol{\ull{C}}^\star, \widehat{\ol{\ull{C}}}\;(\ull{A},\tau)\big).
  \label{eq:invC_objective}
\end{align}
Since $\cT,\cR,\cC_\theta,\cM_\theta,\cB$ are all differentiable end-to-end, the gradients $\partial\cJ/\partial(\ull{A},\tau)$ are obtained by automatic differentiation. 
We solve~\cref{eq:invC_objective} in batched form: with $N_\mathrm{start}$ independent initializations, and optimize all $(\ull{A}^{(s)},\tau^{(s)})$ ($s = 1, \dots N_\mathrm{start}$) in parallel using the \texttt{AdamW} optimizer. 
The amplitude matrices $\ull{A}^{(s)}$ are initialized to random noise, and the threshold parameters $\tau^{(s)}$ are set to $0.5$.
At convergence, we verify a few realizations related to the lowest objective function values by full homogenization to report the true relative error, which is inaccessible during the inverse design phase.
\Cref{fig:mech2d_inverse_to_C} assembles four representative targets (one per row). 
For each target, the second column shows the directional Young's modulus surface computed from the plane-strain compliance tensor, overlaid for the target and the six best candidates. 
In all four cases, the recovered directional Young's modulus $E(\theta)$ almost perfectly matches the target in shape and amplitude, while the relative Frobenius error between the optimized candidates and the target tensor is only a few percent. 
We also observe a many-to-one effect: different unit cell candidates can realize near-identical effective tensors consistent with the nonuniqueness of G-closure. However, the optimizer reliably lands on physically meaningful, percolating, or laminated motifs when required. 
Practically, the batched formulation can yield dozens to hundreds of high-quality candidates in a single run; no auxiliary "inverse" network is needed, and one surrogate suffices for both forward prediction and backpropagated design.

\begin{figure}[!htb]
    \centering
    \begin{tikzpicture}
      \node[draw=gray!50, thick, rounded corners=10pt, inner sep=5pt] {
      \includegraphics[scale=1.0, trim={0cm 0.25cm 0cm 0.25cm},clip]{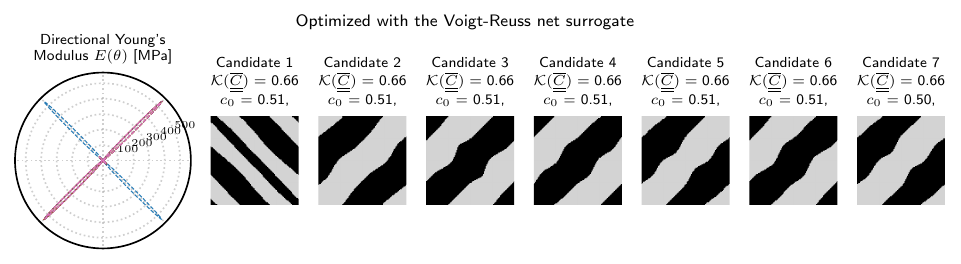}
      };
    \end{tikzpicture}
    \includegraphics[scale=1.0]{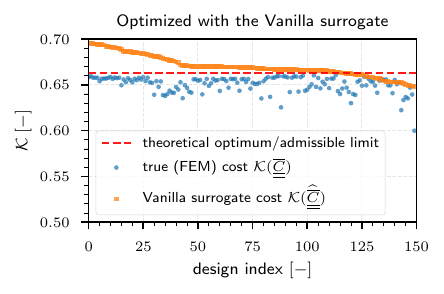}
    \hspace{1cm}
    \includegraphics[scale=1.0]{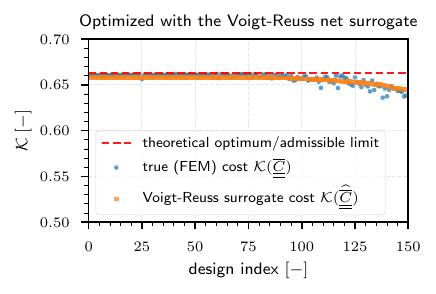}
    \caption{Inverse design for extreme shear-normal coupling.
    \textbf{Top:} For the \VRNet{}-driven optimization, polar surfaces of the directional Young's modulus $E(\theta)$ for the top candidates together with the corresponding optimized microstructures (ranked by surrogate objective).
    \textbf{Left/Right:} Candidate ranking by surrogate objective for the vanilla network (left) and the \VRNet{} (right), each for 150 harvested designs.
    For each candidate, we report the surrogate cost $\cK({\ullWH{\ol{C}}})$ (available during optimization) and the ground truth FEM evaluated cost $\cK(\ull{\ol{C}})$ computed a posteriori; the theoretical admissible limit is indicated for reference.
    The vanilla surrogate can assign costs beyond the admissible maximum, whereas the \VRNet{} objective converges to the theoretical limit while remaining unconditionally admissible.}
    \label{fig:mech2d_inverse_coupling}
\end{figure}

Beyond matching a prescribed tensor, we can pose design goals directly in terms of tensor functionals and optimize the generator parameters $(\ull{A},\tau)$ with gradients through the surrogate.
To maximize plane-strain shear-normal couplings, we consider the objective
\begin{align}
  \cK(\ullWH{\ol{C}})
  \;=\; \max_{\ull{A},\;\tau \in (0,1)} \frac{\sqrt{2}\,\left(\lvert\WH{\ol{C}}_{1112}\rvert + \lvert \WH{\ol{C}}_{2212}\rvert \right)}{\|\ullWH{\ol{C}}\|_\mathsf{F}}.
  \label{eq:coupling_objective}
\end{align}
We run the same multistart, fully batched \texttt{AdamW} optimization with two surrogates: the vanilla regressor and the \VRNet{} (identical architecture, inputs, dataset, and training schedule; only the output parameterization differs).
At convergence, we rank 150 candidates by surrogate cost and then evaluate the true objective $\cK(\ull{\ol{C}})$ using the homogenization solver.
\Cref{fig:mech2d_inverse_coupling} shows that the \VRNet{} recovers the known optimum i.e., a $45^\circ$ laminate with $c_0\approx 0.5$ and its surrogate objective saturates tightly at the theoretical admissible limit, consistent with strict Voigt-Reuss enforcement.
In contrast, the vanilla surrogate can overestimate the coupling and assign values that exceed the admissible maximum. This mismatch is revealed once the designs are re-evaluated with the homogenization solver, leading to significant discrepancies between surrogate and true objectives.

\begin{figure}[!htb]
    \centering
    \begin{tikzpicture}
      \node[draw=gray!50, thick, rounded corners=10pt, inner sep=5pt] {
      \includegraphics[scale=1.0, trim={0cm 0.0cm 0cm 0.25cm},clip]{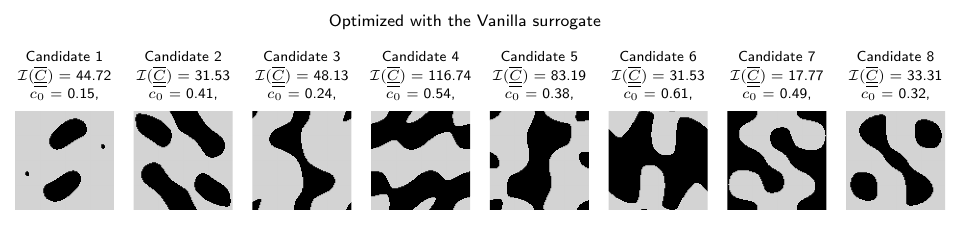}
      };
    \end{tikzpicture}\\[0.1cm]
    \begin{tikzpicture}
      \node[draw=gray!50, thick, rounded corners=10pt, inner sep=5pt] {
      \includegraphics[scale=1.0, trim={0cm 0.0cm 0cm 0.25cm},clip]{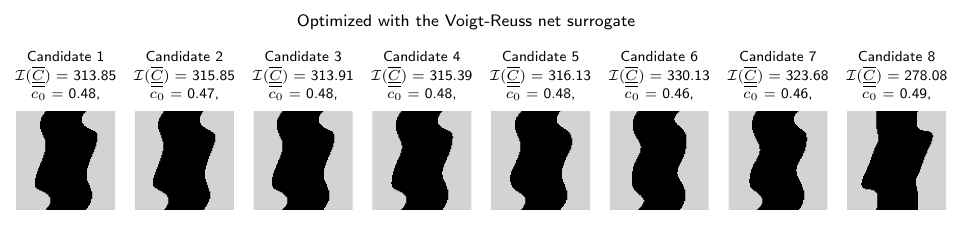}
      };
    \end{tikzpicture}
    \caption{Inverse design for maximizing spectral anisotropy via the eigenvalue ratio.
    \textbf{Top:} Designs obtained by optimizing the vanilla surrogate.
    \textbf{Bottom:} Designs obtained by optimizing the \VRNet{} surrogate.
    In each row, eight candidates are shown, ranked left-to-right by surrogate objective $\cI(\ullWH{\ol{C}})$, with the true solver evaluated objective $\cI(\ull{\ol{C}})$ printed above each microstructure.
    The vanilla optimization yields completely spurious candidates and poor surrogate-truth agreement, while \VRNet{} produces physically meaningful percolating motifs and substantially better alignment between surrogate and true objectives.}
    \label{fig:mech2d_inverse_eigratio}
\end{figure}

As a second functional objective, we maximize the ratio of the largest to the smallest eigenvalue,
\begin{align}
  \cI(\ullWH{\ol{C}})
  \;=\; \max_{\ull{A},\;\tau\in(0,1)} \frac{\lambda_{\max}\big(\ullWH{\ol{C}}\big)}{\lambda_{\min}\big(\ullWH{\ol{C}}\big)} \, ,
  \label{eq:eigratio_objective}
\end{align}
again using the same multistart, fully batched \texttt{AdamW} setup and ranking candidates by surrogate cost before recomputing $\cI(\ull{\ol{C}})$ with the homogenization solver.
\Cref{fig:mech2d_inverse_eigratio} highlights a qualitative difference: optimizing the vanilla surrogate tends to exploit non-physical regions of its output space, producing candidates that do not translate into high true eigenvalue ratios.
In contrast, the \VRNet{} driven optimization yields structured, percolating designs that consistently realize large true eigenvalue ratios while remaining admissible.

\section{Conclusions}

\subsection{Summary}
Across all settings, i.e. in both 2D and 3D mechanical problems, and for an extremely wide range of material contrasts, we demonstrated that spectral normalization as the major building block of the \VRNet{} provides a principled scaffold for learning effective constitutive tensors that are guaranteed to be symmetric positive definite and to lie between the first order Voigt and Reuss bounds in the Löwner sense. By mapping every sample into a bounded target space that is isomorphic to a unit hyper-cube, our surrogates possess robust training dynamics, well-conditioned, scale-free losses, and return physically admissible predictions by construction.

We formalized a universal reparameterization introduced in \cite{Keshav2025} and extended it to mechanical homogenization problems. Notably, the presence of zero eigenvalues in the difference of Voigt and Reuss bounds, e.g., triggered by matching single isotropic material constants, causes some technical issues that must be considered with care.
The learning objective leads to a task-agnostic loss $\phi(\ullWT{Y},\ullWTH{Y})\in[0,1]$ that directly measures errors relative to the admissible Voigt-Reuss gap, decoupling learning from units and improving numerical stability.

In the case of 3D mechanical homogenization as a forward problem, we trained a \VRNet{} surrogate using 236 isotropic microstructure descriptors from \cite{ulm2020_dataset,Prifling2021} along with three non-dimensional phase parameters $(\pi_1,\pi_2,\pi_3)$, on $>1.18$ Mio high-fidelity DNS labels of effective elasticity tensors.
Because the inputs are rotationally invariant, the surrogate recovers the isotropic projection of $\ol{\ull{C}}$ nearly perfectly with the components related to isotropy attain $R^2\ge0.998$ while anisotropy-revealing entries remain unpredictable.
At the tensor level, histograms of the relative Frobenius error of the training, validation, and testing datasets overlap tightly (mean $\approx 3.6\%$, median $\approx 1.8\%$), indicating strong generalization across microstructures and contrasts.
ECDFs reveal that the isotropic projection forms a lower floor (median $1.03\%$), and \VRNet{} sits close to the floor (median $1.79\%$), outperforming the Hill average (median $29.12\%$) by far.
To explicitly assess the benefit of the spectral parameterization, we additionally trained three further surrogates on the identical dataset and inputs (vanilla neural network, Cholesky neural network, and an isotropic two-parameter $(\ol{K},\ol{G})$ predictor).
The \VRNet{} achieves the lowest errors and uniquely produces strictly admissible predictions with zero Voigt/Reuss violations, whereas the alternative surrogates exhibit substantially larger errors paired with frequent bound violations (and the vanilla neural network also produces non-SPD outputs); see~\cref{tab:metrics_violations}.
Realistic composite material pairings (e.g., Al-SiC, epoxy-glass, etc.) lie squarely in the low-error regime of the \VRNet{}.

In the context of the 2D mechanical homogenization (forward and inverse), we introduced an end-to-end differentiable surrogate, enabling a hybrid convolutional and multi-layer perceptron based surrogate that ingests $(\ull{A},\tau)$ and outputs a spectrally admissible $\ol{\ull{C}}$.
Forward predictions reach an excellent coefficient of determination of $R^2>0.99$ across all components of the elasticity tensor and subpercent Voigt-Reuss normalized losses.
To compare the performance of the \VRNet{}, we also trained a Cholesky and vanilla regressor (same architecture and training protocol), and also a descriptor-based \VRNet{} that maps directly from $(c_0,\tau,\mathrm{vec}(\ull{A}^{\urcorner}))$. \Cref{tab:metrics_violations_2d} shows that the image-based \VRNet{} attains the lowest mean/median/standard deviation errors with zero violations, while the descriptor-based \VRNet{} remains admissible but is less accurate, highlighting the inductive bias and transfer potential of image-based features.
Crucially, the differentiable chain $(\ull{A},\tau)\mapsto \ol{\ull{C}}$ of the \VRNet{} exposes informative gradients and resolves non-smooth percolation events (eigenvalue jumps) along threshold sweeps seamlessly, while the corresponding vanilla regressor exhibits significantly larger errors and violates the bounds.

Leveraging differentiability and strict admissibility, we performed inverse design to target full tensors and functionals, recovering classical optima (e.g., $45^\circ$ laminates at $c_0 \approx 0.5$ for extreme shear-normal coupling) and diverse near-optimal candidates in a single batched optimization run.
By prescribing a single target effective stiffness tensor $\ol{\ull{C}}^\star$, the batch run of the optimizer can generate a variety of structurally distinct candidate microstructures that achieve nearly identical mechanical properties, with relative Frobenius errors as low as a few percent.
This many-to-one mapping underscores the richness of the G-closure set, where multiple geometric realizations can produce equivalent effective responses, offering designers flexibility in selecting structures based on additional criteria such as manufacturability, weight, or connectivity.
Moreover, the batched optimization approach enables simultaneous exploration of hundreds of candidates, ensuring diversity while guaranteeing that every proposed design lies within the physically admissible Voigt-Reuss bounds, thus avoiding infeasible or unstable configurations. It should also be noted that the same machinery supports inverse tasks without auxiliary networks, enabling large-batch, first-order exploration of G-closure subsets while remaining inside the Voigt-Reuss bounds at every iterate.

Overall, \VRNet{} delivers accurate, admissible, and data-efficient surrogates for multiscale materials, unifying forward prediction and gradient-based design under a single, physically grounded normalization.

\subsection{Discussion}

The results in 3D and 2D together highlight two complementary roles of the \VRNet{} framework. 
First, when coupled with hand-crafted 3D microstructure descriptors, it acts as a physically constrained surrogate for effective stiffness that is robust to large contrast variation but fundamentally limited by the information content of the input features (here, isotropic descriptors). 
Second, in the 2D setting with a differentiable renderer, the surrogate is a fully image-based, end-to-end differentiable map from parametric microstructure generators $(\ull{A},\tau)$ to effective tensors, opening the door to gradient-based inverse design.
Both use cases share the same spectral normalization backbone and loss, suggesting that the methodology is not tied to any particular architecture but rather to the Löwner-bounded reparameterization of constitutive tensors.

The 3D study illustrates a key structural limitation that is often implicit in data-driven homogenization, i.e., when the feature map is invariant under $SO(3)$, no surrogate can recover orientation-dependent anisotropy. 
In our case, the descriptors~\cite{Prifling2021} encode only isotropy-averaged information, so the model can learn at best the isotropic projection $\ol{\ull{C}}_{\mathrm{iso}}$ of the effective stiffness. 
The sharp split in the $R^2$-matrix (near-perfect prediction for isotropic components, and $R^2\approx0$ for anisotropy-revealing entries) is not a failure of the proposed model but a faithful reflection of the available input information. 
This is a diagnostic for feature sufficiency, and anisotropic components remain unpredictable without orientation-sensitive descriptors (e.g., directional correlations, band features, Minkowski tensors, or full CNN-based encoders).
The ECDFs of the relative Frobenius errors make this explicit: the isotropic projection error seems to be a lower bound for the error of the studied surrogates using the descriptors, and the \VRNet{} is found closest to that lower bound.
In such an isotropy-dominated regime, one could predict only two effective parameters (e.g., $(\ol{K},\ol{G})$) instead of a full tensor. 
Our additional isotropic surrogate outputting only effective isotropic moduli shows that this is not competitive with respect to prediction accuracy: despite explicitly targeting $(\ol{K},\ol{G})$, it remains substantially less accurate than \VRNet{}, indicating that the spectral normalization remains beneficial in this case, too.
Moreover, the vanilla and Cholesky surrogates exhibit frequent violations of the Voigt and Reuss bounds despite being trained on the identical network, dataset, and inputs, while \VRNet{} uniquely yields zero violations by construction (\cref{tab:metrics_violations}).

The 2D experiments, by contrast, operate with a much richer generative parameterization of microstructures via thresholded trigonometric fields instead of a limited number of microstructure classes with random realizations as in the extensive 3D example. Here, the bottleneck is rather massive variability of the microstructures paired with a distinct phase contrast. The corresponding G-closure subset is, therefore, much richer, encompassing, e.g., massively anisotropic tensors. 
Embedding the amplitude matrix~$\ull{A} \in \ffR^{M_1\times M_2}$ into a fixed hyper-grid of size $K \geq M_1, M_2$, rendering the microstructure image based on $\ull{A}^{\urcorner}\in\ffR^{K\times K}$ (here: $K=11$), and regressing the normalized spectral degrees of freedom together yield a surrogate that is sensitive to connectivity, captures the effects due to topology changes, and can replicate extreme anisotropy, all that while remaining numerically stable.
The threshold-sweep study shows that the \VRNet{} tracks sharp eigenvalue jumps at percolation events and exposes these transitions through spikes in the threshold-related sensitivity~$\partial \|\ullWH{\ol{C}}\|_\mathsf{F}/\partial\tau$, while classical Mori-Tanaka estimates deteriorate as their dilute, non-interacting assumptions are increasingly violated. 
The vanilla surrogate, evaluated on the identical sweep, exhibits substantially larger errors, fails to capture the smaller eigenvalues reliably, and produces eigenvalues that oftentimes violate the Reuss bound. 
Even though both surrogates are differentiable and provide gradients, only \VRNet{} maintains physically meaningful predictions throughout the path, including across non-smooth connectivity transitions.

The inverse-design studies highlight the distinctive benefits of a spectrally normalized surrogate:
\begin{itemize}
  \item \textbf{Robust gradients.} All outputs live in $[0,1]^m$, and the renderer $\cR$ is smooth in $(\ull{A},\tau)$; hence, gradients remain well-behaving even at extreme phase contrasts, enabling stable first-order optimization without manual rescaling or ad-hoc clipping. 
  \item \textbf{Batched optimization.} Optimizing hundreds of candidates in one GPU pass provides a diverse family of designs (useful for screening, multi-criteria trade-offs, or robustness analysis) essentially at the cost of a "single run".
  \item \textbf{Single model, forward and backward.} Unlike frameworks that pair a forward surrogate with a separate inverse network~\cite{Kumar2020} or require differentiable simulators~\cite{Pundir2025}, our pipeline uses one spectrally normalized surrogate both for prediction and for backpropagating arbitrary differentiable objectives (tensor matching, eigenvalue ratios, directional moduli, compliance measures, volume constraints, etc.). This takes away the effort for designing, training and validating the inverse model.
  \item \textbf{Guaranteed admissibility.} For every intermediate design and the final design the surrogate respects Voigt-Reuss bounds in the Löwner sense; i.e. the optimizer never leaves the feasible spectral simplex, avoiding unphysical attractors that commonly plague unconstrained neural inverse design.
  \item  \textbf{Empirical coverage of G-closure.} The cosine-threshold parameterization spans a rich, multimodal subset of the G-closure. We recover both classical optima and diverse microstructures that realize nearly identical effective property tensors, directly reflecting the intrinsic non-uniqueness of microstructure-to-property mappings.
\end{itemize}

At a higher level, the framework can be viewed as a “spectral layer” that decouples physical admissibility from architectural choice.
Any regressor that produces normalized spectral parameters $\ulWH{\xi}$ can be wrapped by $\cS^{-1}$ to yield outputs that are symmetric positive definite and automatically bounded in the Löwner sense.
This separation of concerns should generalize beyond small-strain elasticity to other symmetric positive definite effective outputs emerging from linear elliptic PDEs, and to multi-property surrogates where several coupled tensors must remain jointly admissible.

There are, however, clear limitations and avenues for improvement. 
In 3D, the present results are intentionally constrained by isotropic descriptors. To capture full anisotropy, one must either augment the feature set with orientation-sensitive statistics or transition to 3D CNNs operating directly on voxelized microstructures, at a much higher computational cost. 
In 2D, the cosine-threshold family, while expressive, samples only a subset (albeit a considerably sized subset) of all possible microstructures. Enriching the generator (e.g., using hierarchical patterns or by incorporating manufacturing-aware constraints) would expand the reachable portion of the G-closure.
Moreover, the current formulation is deterministic: incorporating uncertainty quantification (e.g., Bayesian surrogates or ensemble methods) would allow one to propagate microstructural variability and quantify confidence in inverse designs, which is essential for safety-critical applications.

Finally, the combination of spectral normalization and end-to-end differentiability makes \VRNet{} compatible with existing structural and topology-optimization workflows.
Because the surrogate provides fast, differentiable, and physically admissible predictions, it can act as a “constitutive oracle” inside larger multi-scale loops, enabling gradient-based co-design of architecture and material.
The same machinery supports inverse tasks without auxiliary networks, enabling large-batch, first-order exploration of G-closure subsets while staying inside Voigt-Reuss bounds at every iterate.
In this sense, \VRNet{} offers a unifying, physics-grounded interface between classical homogenization theory and modern deep-learning-based materials design.

\section*{Acknowledgments}
Funded by Deutsche Forschungsgemeinschaft (DFG, German Research Foundation) under Germany's Excellence Strategy - EXC 2075 - 390740016.
Contributions by Felix Fritzen are funded by Deutsche Forschungsgemeinschaft (DFG, German Research Foundation) within the Heisenberg program - DFG-FR2702/10 - 517847245. 
This work is also supported as part of the consortium NFDI-MatWerk, funded by the Deutsche Forschungsgemeinschaft (DFG, German Research Foundation) under the National Research Data Infrastructure - NFDI 38/1 - 460247524. 
We acknowledge the support of the Stuttgart Center for Simulation Science (SimTech).

\subsection*{Financial disclosure}
None reported.

\subsection*{Conflict of interest}
The authors declare no potential conflict of interest.

\section*{Data availability statement}
Upon acceptance of the manuscript, the problem datasets will be published, as well as a simple implementation of the spectral normalization method in \texttt{Pytorch}.

\bibliography{literature}

@ARTICLE{Keshav2025,
  title     = "Spectral Normalization and {Voigt-Reuss} net: A universal
               approach to microstructure-property forecasting with physical
               guarantees",
  author    = "Keshav, Sanath and Herb, Julius and Fritzen, Felix",
  journal   = "GAMM-Mitteilungen",
  publisher = "Wiley",
  volume    =  48,
  number    =  3,
  month     =  sep,
  year      =  2025,
  copyright = "http://creativecommons.org/licenses/by/4.0/",
  language  = "en",
  doi       = "10.1002/gamm.70005",
}

@ARTICLE{Boddapati2023-la,
  title     = "Single-test evaluation of directional elastic properties of
               anisotropic structured materials",
  author    = "Boddapati, Jagannadh and Flaschel, Moritz and Kumar, Siddhant
               and De Lorenzis, Laura and Daraio, Chiara",
  journal   = "J. Mech. Phys. Solids",
  publisher = "Elsevier BV",
  volume    =  181,
  number    =  105471,
  pages     = "105471",
  month     =  dec,
  year      =  2023,
  language  = "en",
    doi = {10.1016/j.jmps.2023.105471}
}

@ARTICLE{Boddapati2024-tb,
  title     = "Planar structured materials with extreme elastic anisotropy",
  author    = "Boddapati, Jagannadh and Daraio, Chiara",
  journal   = "Mater. Des.",
  publisher = "Elsevier BV",
  volume    =  246,
  number    =  113348,
  pages     = "113348",
  month     = Oct,
  year      =  2024,
  copyright = "http://creativecommons.org/licenses/by-nc/4.0/",
  language  = "en",
  doi = {10.1016/j.matdes.2024.113348}
}

@BOOK{Milton2002,
  title     = "The theory of composites",
  author    = "Milton, Graeme W",
  publisher = "Cambridge University Press (Virtual Publishing)",
  series    = "Cambridge Monographs on Applied and Computational Mathematics",
  month     =  dec,
  year      =  2009,
  address   = "Cambridge, England"
}

@BOOK{Torquato2002,
  title     = "Random heterogeneous materials",
  author    = "Torquato, Salvatore",
  publisher = "Springer",
  series    = "Interdisciplinary applied mathematics",
  month     =  mar,
  year      =  2013,
  address   = "New York, NY",
  language  = "en"
}

@ARTICLE{Nguyen2025-cn,
  title         = "Deep learning-aided inverse design of porous metamaterials",
  author        = "Nguyen, Phu Thien and Heider, Yousef and Kochmann, Dennis M
                   and Aldakheel, Fadi",
  month         =  jul,
  year          =  2025,
  copyright     = "http://creativecommons.org/licenses/by/4.0/",
  archivePrefix = "arXiv",
  primaryClass  = "cs.LG",
  eprint        = "2507.17907"
}

@ARTICLE{Heider2025-il,
  title     = "A multiscale {CNN-based} intrinsic permeability prediction in
               deformable porous media",
  author    = "Heider, Yousef and Aldakheel, Fadi and Ehlers, Wolfgang",
  journal   = "Appl. Sci. (Basel)",
  publisher = "MDPI AG",
  volume    =  15,
  number    =  5,
  pages     = "2589",
  month     =  feb,
  year      =  2025,
  copyright = "https://creativecommons.org/licenses/by/4.0/",
  language  = "en"
}

@ARTICLE{Kumar2020-vq,
  title     = "Inverse-designed spinodoid metamaterials",
  author    = "Kumar, Siddhant and Tan, Stephanie and Zheng, Li and Kochmann,
               Dennis M",
  journal   = "Npj Comput. Mater.",
  publisher = "Springer Science and Business Media LLC",
  volume    =  6,
  number    =  1,
  month     =  jun,
  year      =  2020,
  copyright = "https://creativecommons.org/licenses/by/4.0",
  language  = "en"
}

@ARTICLE{Bastek2023-qt,
  title     = "Inverse design of nonlinear mechanical metamaterials via video
               denoising diffusion models",
  author    = "Bastek, Jan-Hendrik and Kochmann, Dennis M",
  journal   = "Nat. Mach. Intell.",
  publisher = "Springer Science and Business Media LLC",
  volume    =  5,
  number    =  12,
  pages     = "1466--1475",
  month     =  dec,
  year      =  2023,
  copyright = "https://creativecommons.org/licenses/by/4.0",
  language  = "en"
}

@MISC{Bastek2023-pp,
    title     = "Inverse-design of nonlinear mechanical metamaterials via video
               denoising diffusion models: Dataset and model checkpoints",
    author    = "Bastek, Jan-Hendrik",
    publisher = "ETH Zurich",
    year      =  2023,
    howpublished = "ETH Zurich research collection",
    doi = {10.3929/ethz-b-000629716}
}

@ARTICLE{Zheng2023-gv,
  title    = "Unifying the design space and optimizing linear and nonlinear
              truss metamaterials by generative modeling",
  author   = "Zheng, Li and Karapiperis, Konstantinos and Kumar, Siddhant and
              Kochmann, Dennis M",
  journal  = "Nat. Commun.",
  volume   =  14,
  number   =  1,
  pages    = "7563",
  month    =  nov,
  year     =  2023,
  language = "en"
}

@ARTICLE{Otto2025-gt,
  title     = "Data-driven inverse design of spinodoid architected materials",
  author    = "Otto, Alexandra and Rosenkranz, Max and Kalina, Karl A and
               K{\"a}stner, Markus",
  journal   = "GAMM-Mitteilungen",
  publisher = "Wiley",
  volume    =  48,
  number    =  4,
  month     =  nov,
  year      =  2025,
  copyright = "http://creativecommons.org/licenses/by/4.0/",
  language  = "en"
}

@ARTICLE{Rosenkranz2025-ej,
  title     = "Data-efficient inverse design of spinodoid metamaterials",
  author    = "Rosenkranz, Max and K{\"a}stner, Markus and Sbalzarini, Ivo F",
  journal   = "Integr. Mater. Manuf. Innov.",
  publisher = "Springer Science and Business Media LLC",
  month     =  oct,
  year      =  2025,
  copyright = "https://creativecommons.org/licenses/by/4.0",
  language  = "en"
}

@ARTICLE{Hashin1963-am,
  title     = "A variational approach to the theory of the elastic behaviour of
               multiphase materials",
  author    = "Hashin, Z and Shtrikman, S",
  journal   = "J. Mech. Phys. Solids",
  publisher = "Elsevier BV",
  volume    =  11,
  number    =  2,
  pages     = "127--140",
  month     =  mar,
  year      =  1963,
  language  = "en"
}

@inproceedings{He2015-ws,
    title={Deep residual learning for image recognition},
    author={He, Kaiming and Zhang, Xiangyu and Ren, Shaoqing and Sun, Jian},
    booktitle={Proceedings of the IEEE conference on computer vision and pattern recognition},
    pages={770--778},
    organization = {IEEE},
    year={2016}
}

@InProceedings{He2016-bu,
    author="He, Kaiming
    and Zhang, Xiangyu
    and Ren, Shaoqing
    and Sun, Jian",
    editor="Leibe, Bastian
    and Matas, Jiri
    and Sebe, Nicu
    and Welling, Max",
    title="Identity Mappings in Deep Residual Networks",
    booktitle="Computer Vision -- ECCV 2016",
    year="2016",
    publisher="Springer International Publishing",
    address="Cham",
    pages="630--645",
    organization = {},
    isbn="978-3-319-46493-0"
}

@Inbook{Allaire2002,
author="Allaire, Gr{\'e}goire",
title="The mathematical modeling of composite materials",
bookTitle="Shape Optimization by the Homogenization Method",
year="2002",
publisher="Springer New York",
address="New York, NY",
pages="91--188",
isbn="978-1-4684-9286-6",
doi="10.1007/978-1-4684-9286-6_2",
url="https://doi.org/10.1007/978-1-4684-9286-6_2"
}

@Inbook{Cherkaev2000,
author="Cherkaev, Andrej",
title="Obtaining G-Closures",
bookTitle="Variational Methods for Structural Optimization",
year="2000",
publisher="Springer New York",
address="New York, NY",
pages="261--277",
isbn="978-1-4612-1188-4",
doi="10.1007/978-1-4612-1188-4_10",
url="https://doi.org/10.1007/978-1-4612-1188-4_10"
}

@article{EIDEL2023115741,
title = {Deep CNNs as universal predictors of elasticity tensors in homogenization},
journal = {Computer Methods in Applied Mechanics and Engineering},
volume = {403},
pages = {115741},
year = {2023},
issn = {0045-7825},
doi = {https://doi.org/10.1016/j.cma.2022.115741},
url = {https://www.sciencedirect.com/science/article/pii/S004578252200696X},
author = {Bernhard Eidel}
}

@article{Bhattacharya2024,
author = {Bhattacharya, Kaushik and Kovachki, Nikola B. and Rajan, Aakila and Stuart, Andrew M. and Trautner, Margaret},
title = {Learning Homogenization for Elliptic Operators},
journal = {SIAM Journal on Numerical Analysis},
volume = {62},
number = {4},
pages = {1844-1873},
year = {2024},
doi = {10.1137/23M1585015},
URL = {https://doi.org/10.1137/23M1585015},
eprint = {https://doi.org/10.1137/23M1585015}

}

@article{NGUYEN2026106418,
title = {Universal Fourier Neural Operators for periodic homogenization problems in linear elasticity},
journal = {Journal of the Mechanics and Physics of Solids},
volume = {206},
pages = {106418},
year = {2026},
issn = {0022-5096},
doi = {https://doi.org/10.1016/j.jmps.2025.106418},
url = {https://www.sciencedirect.com/science/article/pii/S0022509625003928},
author = {Binh Huy Nguyen and Matti Schneider}
}

@article{Pundir2025,
  title = {Simplifying FFT-based methods for solid mechanics with automatic differentiation},
  volume = {435},
  ISSN = {0045-7825},
  url = {http://dx.doi.org/10.1016/j.cma.2024.117572},
  DOI = {10.1016/j.cma.2024.117572},
  journal = {Computer Methods in Applied Mechanics and Engineering},
  publisher = {Elsevier BV},
  author = {Pundir,  Mohit and Kammer,  David S.},
  year = {2025},
  month = feb,
  pages = {117572}
}

@Article{Penrose1955,
  author    = {Penrose, R.},
  journal   = {Mathematical Proceedings of the Cambridge Philosophical Society},
  title     = {A generalized inverse for matrices},
  year      = {1955},
  issn      = {1469-8064},
  number    = {3},
  pages     = {406--413},
  volume    = {51},
  doi       = {10.1017/s0305004100030401},
  publisher = {Cambridge University Press (CUP)},
}

@Article{moulinec1998,
  Title                    = {{A numerical method for computing the overall response of nonlinear composites with complex microstructure}},
  Author                   = {Moulinec, H. and Suquet, P.},
  Journal                  = {Computer Methods in Applied Mechanics and Engineering},
  Year                     = {1998},
  Pages                    = {69--94}
}

@Article{Schneider2021,
  author    = {Schneider, Matti},
  journal   = {Acta Mechanica},
  title     = {{A review of nonlinear FFT-based computational homogenization methods}},
  year      = {2021},
  issn      = {1619-6937},
  month     = mar,
  number    = {6},
  pages     = {2051--2100},
  volume    = {232},
  doi       = {10.1007/s00707-021-02962-1},
  publisher = {Springer Science and Business Media LLC},
}

@Article{Rassloff2025,
  author    = {Ra{\ss}loff, Alexander and Seibert, Paul and Kalina, Karl A. and Kästner, Markus},
  journal   = {Computational Mechanics},
  title     = {{Inverse design of spinodoid structures using Bayesian optimization}},
  year      = {2025},
  issn      = {1432-0924},
  month     = feb,
  doi       = {10.1007/s00466-024-02587-w},
  publisher = {Springer Science and Business Media LLC},
}

@article{Karniadakis2021,
  title = {Physics-informed machine learning},
  volume = {3},
  ISSN = {2522-5820},
  url = {http://dx.doi.org/10.1038/s42254-021-00314-5},
  DOI = {10.1038/s42254-021-00314-5},
  number = {6},
  journal = {Nature Reviews Physics},
  publisher = {Springer Science and Business Media LLC},
  author = {Karniadakis,  George Em and Kevrekidis,  Ioannis G. and Lu,  Lu and Perdikaris,  Paris and Wang,  Sifan and Yang,  Liu},
  year = {2021},
  month = may,
  pages = {422–440}
}

@article{Xu2021,
  title = {Learning constitutive relations using symmetric positive definite neural networks},
  volume = {428},
  ISSN = {0021-9991},
  url = {http://dx.doi.org/10.1016/j.jcp.2020.110072},
  DOI = {10.1016/j.jcp.2020.110072},
  journal = {Journal of Computational Physics},
  publisher = {Elsevier BV},
  author = {Xu,  Kailai and Huang,  Daniel Z. and Darve,  Eric},
  year = {2021},
  month = mar,
  pages = {110072}
}

@article{Aldakheel2023,
  title = {Efficient multiscale modeling of heterogeneous materials using deep neural networks},
  volume = {72},
  ISSN = {1432-0924},
  url = {http://dx.doi.org/10.1007/s00466-023-02324-9},
  DOI = {10.1007/s00466-023-02324-9},
  number = {1},
  journal = {Computational Mechanics},
  publisher = {Springer Science and Business Media LLC},
  author = {Aldakheel,  Fadi and Elsayed,  Elsayed S. and Zohdi,  Tarek I. and Wriggers,  Peter},
  year = {2023},
  month = apr,
  pages = {155–171}
}

@article{Leuschner2017,
  title = {Fourier-Accelerated Nodal Solvers (FANS) for homogenization problems},
  volume = {62},
  ISSN = {1432-0924},
  url = {http://dx.doi.org/10.1007/s00466-017-1501-5},
  DOI = {10.1007/s00466-017-1501-5},
  number = {3},
  journal = {Computational Mechanics},
  publisher = {Springer Science and Business Media LLC},
  author = {Leuschner,  Matthias and Fritzen,  Felix},
  year = {2017},
  month = Nov,
  pages = {359–392}
}

@article{combo2022,
  doi = {https://doi.org/10.1007/s00466-022-02232-4},
  year = {2023},
 vol = {71},
number ={1},
pages = {191-212},
  publisher = {Springer Science and Business Media {LLC}},
  author = {Sanath Keshav and Felix Fritzen and Matthias Kabel},
  title = {{FFT}-based homogenization at finite strains using composite boxels ({ComBo})},
  journal = {Computational Mechanics}
}

@misc{FANS_github,
  title = {FANS - An open-source, efficient, and parallel implementation of the Fourier-Accelerated Nodal Solvers for microscale multiphysics problems.},
  url = {https://github.com/DataAnalyticsEngineering/FANS},
  author = {Keshav, Sanath and Rieg, Florian and Desai, Ishaan and Sigg, Moritz and Haag, Claudius and Fritzen, Felix},
  howpublished = {Github repository},
  year = {2024},
}

@book{nemat2013micromechanics,
  title={Micromechanics: Overall Properties of Heterogeneous Materials},
  author={Nemat-Nasser, S. and Hori, M. and Achenbach, J.D.},
  isbn={9781483291512},
  series={ISSN},
  url={https://www.sciencedirect.com/bookseries/north-holland-series-in-applied-mathematics-and-mechanics/vol/37/suppl/C},
  year={2013},
  publisher={Elsevier Science}
}

@article{lissner2024,
  title = {Microstructure homogenization: human vs machine},
  volume = {11},
  ISSN = {2213-7467},
  url = {http://dx.doi.org/10.1186/s40323-024-00275-1},
  DOI = {10.1186/s40323-024-00275-1},
  number = {1},
  journal = {Advanced Modeling and Simulation in Engineering Sciences},
  publisher = {Springer Science and Business Media LLC},
  author = {Li{\ss}ner,  Julian and Fritzen,  Felix},
  year = {2024},
  month = nov 
}

@article{lissner2019,
  title = {Data-Driven Microstructure Property Relations},
  volume = {24},
  ISSN = {2297-8747},
  url = {http://dx.doi.org/10.3390/mca24020057},
  DOI = {10.3390/mca24020057},
  number = {2},
  journal = {Mathematical and Computational Applications},
  publisher = {MDPI AG},
  author = {Li{\ss}ner,  Julian and Fritzen,  Felix},
  year = {2019},
  month = may,
  pages = {57}
}

@article{Prifling2021,
  title = {Large-Scale Statistical Learning for Mass Transport Prediction in Porous Materials Using 90, 000 Artificially Generated Microstructures},
  volume = {8},
  ISSN = {2296-8016},
  url = {http://dx.doi.org/10.3389/fmats.2021.786502},
  DOI = {10.3389/fmats.2021.786502},
  journal = {Frontiers in Materials},
  publisher = {Frontiers Media SA},
  author = {Prifling,  Benedikt and R\"{o}ding,  Magnus and Townsend,  Philip and Neumann,  Matthias and Schmidt,  Volker},
  year = {2021},
  month = dec 
}

@misc{ulm2020_dataset,
  author       = {Benedikt Prifling and
                  Magnus R{\"o}ding and
                  Philip Townsend and
                  Matthias Neumann and
                  Volker Schmidt},
  title        = {Large-scale statistical learning for mass
                   transport prediction in porous materials using
                   90,000 artificially generated microstructures
                  },
  month        = nov,
  year         = 2021,
  howpublished = {Zenodo},
  doi          = {10.5281/zenodo.4047774},
  url          = {https://doi.org/10.5281/zenodo.4047774},
}

@article{Wiener1912,
  title = {Die {{Theorie}} Des {{Mischk{\"o}rpers}} F{\"u}r Das {{Feld}} Der Station{\"a}ren {{Str{\"o}mung}} 1 {{Die Mittelwerts{\"a}tze}} F{\"u}r {{Kraft}}, {{Polarisation}} Und {{Energie}} / von {{Otto Wiener}}},
  author = {Wiener, Otto},
  year = {1912},
  journal = {Abhandlungen der K{\"o}niglich S{\"a}chsischen Gesellschaft der Wissenschaften ; 61,6},
  publisher = {Teubner},
  address = {Leipzig},
  urldate = {2024-12-31},
}

@article{Voigt1889,
  title = {Ueber Die {{Beziehung}} Zwischen Den Beiden {{Elasticit{\"a}tsconstanten}} Isotroper {{K{\"o}rper}}},
  author = {Voigt, W.},
  year = {1889},
  journal = {Annalen der Physik},
  volume = {274},
  number = {12},
  pages = {573--587},
  issn = {1521-3889},
  doi = {10.1002/andp.18892741206},
  urldate = {2024-12-31},
  copyright = {Copyright {\copyright} 1889 WILEY-VCH Verlag GmbH \& Co. KGaA, Weinheim},
  langid = {english},
}

@article{Reuss1929,
  title = {Berechnung Der {{Flie{\ss}grenze}} von {{Mischkristallen}} Auf {{Grund}} Der {{Plastizit{\"a}tsbedingung}} F{\"u}r {{Einkristalle}} .},
  author = {Reuss, A.},
  year = {1929},
  journal = {{ZAMM - Journal of Applied Mathematics and Mechanics}},
  volume = {9},
  number = {1},
  pages = {49--58},
  issn = {1521-4001},
  doi = {10.1002/zamm.19290090104},
  urldate = {2025-01-09},
  copyright = {Copyright {\copyright} 1929 WILEY-VCH Verlag GmbH \& Co. KGaA, Weinheim},
  langid = {english},
}

@article{Fernandez2021,
  title = {Material modeling for parametric,  anisotropic finite strain hyperelasticity based on machine learning with application in optimization of metamaterials},
  volume = {123},
  ISSN = {1097-0207},
  url = {http://dx.doi.org/10.1002/nme.6869},
  DOI = {10.1002/nme.6869},
  number = {2},
  journal = {International Journal for Numerical Methods in Engineering},
  publisher = {Wiley},
  author = {Fernández,  Mauricio and Fritzen,  Felix and Weeger,  Oliver},
  year = {2021},
  month = nov,
  pages = {577–609}
}

@article{Linden2023,
  title = {Neural networks meet hyperelasticity: A guide to enforcing physics},
  volume = {179},
  ISSN = {0022-5096},
  url = {http://dx.doi.org/10.1016/j.jmps.2023.105363},
  DOI = {10.1016/j.jmps.2023.105363},
  journal = {Journal of the Mechanics and Physics of Solids},
  publisher = {Elsevier BV},
  author = {Linden,  Lennart and Klein,  Dominik K. and Kalina,  Karl A. and Brummund,  J\"{o}rg and Weeger,  Oliver and K\"{a}stner,  Markus},
  year = {2023},
  month = oct,
  pages = {105363}
}

@ARTICLE{Klein2025,
  title     = "Neural networks meet hyperelasticity: A monotonic approach",
  author    = "Klein, Dominik K and Hossain, Mokarram and Kikinov, Konstantin
               and Kannapinn, Maximilian and Rudykh, Stephan and Gil, Antonio J",
  journal   = "Eur. J. Mech. A Solids",
  publisher = "Elsevier BV",
  volume    =  116,
  number    =  105900,
  pages     = "105900",
  month     =  mar,
  year      =  2026,
  copyright = "http://creativecommons.org/licenses/by/4.0/",
  language  = "en"
}

@article{Farhat2022,
  title = {A mechanics‐informed artificial neural network approach in data‐driven constitutive modeling},
  volume = {123},
  ISSN = {1097-0207},
  url = {http://dx.doi.org/10.1002/nme.6957},
  DOI = {10.1002/nme.6957},
  number = {12},
  journal = {International Journal for Numerical Methods in Engineering},
  publisher = {Wiley},
  author = {As’ad,  Faisal and Avery,  Philip and Farhat,  Charbel},
  year = {2022},
  month = mar,
  pages = {2738–2759}
}

@article{Kumar2020,
  title = {Inverse-designed spinodoid metamaterials},
  volume = {6},
  ISSN = {2057-3960},
  url = {http://dx.doi.org/10.1038/s41524-020-0341-6},
  DOI = {10.1038/s41524-020-0341-6},
  number = {1},
  journal = {npj Computational Materials},
  publisher = {Springer Science and Business Media LLC},
  author = {Kumar,  Siddhant and Tan,  Stephanie and Zheng,  Li and Kochmann,  Dennis M.},
  year = {2020},
  month = jun 
}

@book{zohdi2005,
  title = {An Introduction to Computational Micromechanics: Corrected Second Printing},
  ISBN = {9783540323600},
  ISSN = {1613-7736},
  url = {http://dx.doi.org/10.1007/978-3-540-32360-0},
  DOI = {10.1007/978-3-540-32360-0},
  journal = {Lecture Notes in Applied and Computational Mechanics},
  publisher = {Springer Berlin Heidelberg},
  author = {Zohdi,  Tarek and Wriggers,  Peter},
  year = {2005}
}

@inproceedings{Ansel_PyTorch_2_Faster_2024,
author = {Ansel, Jason and Yang, Edward and He, Horace and Gimelshein, Natalia and Jain, Animesh and Voznesensky, Michael and Bao, Bin and Bell, Peter and Berard, David and Burovski, Evgeni and Chauhan, Geeta and Chourdia, Anjali and Constable, Will and Desmaison, Alban and DeVito, Zachary and Ellison, Elias and Feng, Will and Gong, Jiong and Gschwind, Michael and Hirsh, Brian and Huang, Sherlock and Kalambarkar, Kshiteej and Kirsch, Laurent and Lazos, Michael and Lezcano, Mario and Liang, Yanbo and Liang, Jason and Lu, Yinghai and Luk, C. K. and Maher, Bert and Pan, Yunjie and Puhrsch, Christian and Reso, Matthias and Saroufim, Mark and Siraichi, Marcos Yukio and Suk, Helen and Zhang, Shunting and Suo, Michael and Tillet, Phil and Zhao, Xu and Wang, Eikan and Zhou, Keren and Zou, Richard and Wang, Xiaodong and Mathews, Ajit and Wen, William and Chanan, Gregory and Wu, Peng and Chintala, Soumith},
title = {{PyTorch 2: Faster Machine Learning Through Dynamic Python Bytecode Transformation and Graph Compilation}},
year = {2024},
isbn = {9798400703850},
publisher = {Association for Computing Machinery},
address = {New York, NY, USA},
url = {https://doi.org/10.1145/3620665.3640366},
doi = {10.1145/3620665.3640366},
booktitle = {{Proceedings of the 29th ACM International Conference on Architectural Support for Programming Languages and Operating Systems, Volume 2}},
pages = {929–947},
numpages = {19},
location = {La Jolla, CA, USA},
series = {ASPLOS '24},
organization={ACM}
}

@ARTICLE{Talbot2004,
  title     = "Bounds for the effective constitutive relation of a nonlinear
               composite",
  author    = "Talbot, D R S and Willis, J R",
  journal   = "Proc. Math. Phys. Eng. Sci.",
  publisher = "The Royal Society",
  volume    =  460,
  number    =  2049,
  pages     = "2705--2723",
  month     =  sep,
  year      =  2004,
  language  = "en"
}

@ARTICLE{Castaneda1992,
  title     = "Bounds and estimates for the properties of nonlinear
               heterogeneous systems",
  author    = "Castaneda, P P",
  journal   = "Philos. Trans. Phys. Sci. Eng.",
  publisher = "The Royal Society",
  volume    =  340,
  number    =  1659,
  pages     = "531--567",
  month     =  sep,
  year      =  1992,
  language  = "en"
}

\end{document}